\documentclass{article}

\PassOptionsToPackage{numbers, compress, sort}{natbib}

\def\status{main,final}
\usepackage[\status]{neurips_2025}
\usepackage{titletoc}

\usepackage[utf8]{inputenc} %
\usepackage[T1]{fontenc}    %
\usepackage{hyperref}       %
\usepackage{url}            %
\usepackage{booktabs}       %
\usepackage{amsfonts}       %
\usepackage{nicefrac}       %
\usepackage{microtype}      %
\usepackage{xcolor}         %
\usepackage{amsmath}
\usepackage{amssymb}
\usepackage{bm}
\usepackage{mathtools}

\def\rvv{{\mathbf{v}}}

\def\vzero{{\bm{0}}}
\def\vone{{\bm{1}}}

\def\vtheta{{\bm{\theta}}}

\def\vvarepsilon{{\bm{\varepsilon}}}

\def\va{{\bm{a}}}
\def\vb{{\bm{b}}}

\def\ve{{\bm{e}}}

\def\vf{{\bm{f}}}
\def\vg{{\bm{g}}}
\def\vh{{\bm{h}}}
\def\vi{{\bm{i}}}
\def\vj{{\bm{j}}}

\def\vm{{\bm{m}}}

\def\vp{{\bm{p}}}
\def\vphi{{\bm{\phi}}}
\def\vq{{\bm{q}}}

\def\vs{{\bm{s}}}

\def\vv{{\bm{v}}}

\def\vx{{\bm{x}}}

\def\mA{{\bm{A}}}

\def\mD{{\bm{D}}}

\def\mI{{\bm{I}}}

\def\mW{{\bm{W}}}

\def\msigma{{\bm{\sigma}}}

\DeclareMathAlphabet{\mathsfit}{\encodingdefault}{\sfdefault}{m}{sl}
\SetMathAlphabet{\mathsfit}{bold}{\encodingdefault}{\sfdefault}{bx}{n}
\newcommand{\tens}[1]{\bm{\mathsfit{#1}}}
\def\tA{{\tens{A}}}
\def\tB{{\tens{B}}}
\def\tC{{\tens{C}}}

\def\sN{{\mathbb{N}}}

\def\sR{{\mathbb{R}}}

\DeclareSymbolFont{bbold}{U}{bbold}{m}{n}
\DeclareSymbolFontAlphabet{\mathbbold}{bbold}

\newcommand{\E}{\mathbb{E}}

\newcommand{\R}{\mathbb{R}}

\DeclareMathOperator{\partitioning}{part}

\DeclareMathOperator{\Tr}{Tr}
\DeclareMathOperator{\diag}{diag}

\DeclareMathOperator{\rank}{rank}

\DeclarePairedDelimiterX{\KLdivx}[2]{(}{)}{%
  #1\;\delimsize\|\;#2%
}

\newcommand{\tensorprod}[1]{%
  \mathbin{\mathop{\otimes}\limits_{#1}}%
}

\usepackage{nicefrac} %
\usepackage{dsfont} %
\usepackage[%
exponent-product=\ensuremath{\cdot},%
group-minimum-digits={3}%
]{siunitx} %
\usepackage{xstring} %

\usepackage{cleveref} %
\crefname{section}{\S\!\!}{\S\!\!} %
\crefname{appendix}{\S\!\!}{\S\!\!} %

\usepackage{todonotes} %
\usepackage{comment} %
\usepackage{enumitem} %
\usepackage{xspace}
\newcommand*{\ie}{i.e.\@\xspace}
\newcommand*{\iid}{i.i.d.\@\xspace}
\newcommand*{\wrt}{w.r.t.\@\xspace}
\newcommand*{\eg}{e.g.\@\xspace}

\newcommand*{\Eg}{E.g.\@\xspace}

\usepackage{catchfile} %

\newcommand{\inputMetricOnly}[1]{%
  \unskip
  \CatchFileDef{\filecontent}{#1}{}%
  \StrBefore{\filecontent}{(}[\metricOnly]%
  {%
    \sisetup{%
      round-mode=figures,%
      round-precision=2,%
      detect-weight, %
      tight-spacing=true, %
    }%
    \ \metricOnly%
    \unskip
  }%
  \unskip%
}

\makeatletter
\newcommand{\extractfirstnum}[2]{%
  \setbox0=\hbox{%
    \begingroup
    \edef\temp{#2}%
    \expandafter\extract@firstnum\temp\relax
    \global\let\tempnum\extract@result
    \endgroup
  }%
  \pgfmathsetmacro{#1}{\tempnum}%
}
\def\extract@firstnum#1#{\extract@@firstnum}
\def\extract@@firstnum#1{%
  \def\extract@result{#1}%
}
\makeatother

\newcommand{\inputMetricNumeric}[2]{%
  \CatchFileDef{\filecontents}{#2}{}%
  \extractfirstnum{#1}{\filecontents}%
}

\newcommand{\inputMetricRatio}[2]{%
  \inputMetricNumeric{\numerator}{#1}%
  \inputMetricNumeric{\denominator}{#2}%
  \unskip
  {%
    \sisetup{%
      round-mode=figures,%
      round-precision=2,%
      detect-weight, %
      tight-spacing=true, %
    }%
    \pgfmathparse{\numerator / \denominator}%
    \ \num{\pgfmathresult}%
  }%
}

\usepackage{wrapfig} %
\usepackage{subcaption} %
\usepackage{tabularx} %
\usepackage{tikz} %
\usetikzlibrary{
  arrows.meta, %
  matrix, %
  positioning, %
  calc, %
  plotmarks, %
}
\usepackage{pgfplots} %

\definecolor{VectorBlack}{RGB}{34, 34, 34}
\definecolor{VectorGray}{RGB}{239, 238, 237}
\definecolor{VectorBlue}{RGB}{59, 69, 227}
\definecolor{VectorPink}{RGB}{253, 8, 238}
\definecolor{VectorOrange}{RGB}{250, 173, 26}
\definecolor{VectorTeal}{RGB}{82, 199, 222}

\colorlet{maincolor}{VectorBlue}
\colorlet{secondcolor}{VectorPink}
\colorlet{thirdcolor}{VectorOrange}
\colorlet{fourthcolor}{VectorTeal}
\colorlet{fifthcolor}{VectorGray}

\definecolor{tab-orange}{rgb}{1.0, 0.498, 0.055}
\definecolor{tab-blue}{rgb}{0.121, 0.466, 0.705}
\definecolor{tab-green}{rgb}{0.173, 0.627, 0.173}
\definecolor{colorbrewer-green}{RGB}{27, 158, 119}
\colorlet{tab-green}{colorbrewer-green}
\definecolor{colorbrewer-orange}{RGB}{217, 95, 2}
\colorlet{tab-orange}{colorbrewer-orange}
\definecolor{colorbrewer-blue}{RGB}{117, 112, 179}
\colorlet{tab-blue}{colorbrewer-blue}

\hypersetup{%
  colorlinks,
  citecolor = maincolor,%
  linkcolor = maincolor,%
  urlcolor = secondcolor,%
}%

\usepackage{pifont} %
\newcommand{\cmark}{\ding{51}}
\newcommand{\xmark}{\ding{55}}

\usepackage{algorithm}
\usepackage{algpseudocode}

\usepackage{multirow}
\usepackage{array} %
\usepackage{makecell} %
\usepackage[bottom]{footmisc}

\newcommand{\papertitle}{%
  Collapsing Taylor Mode Automatic Differentiation
}
\title{\papertitle}

\author{%
  Felix Dangel\thanks{Equal contribution}\\
  Vector Institute \\
  Toronto, Canada \\
  \texttt{fdangel@vectorinstitute.ai} \\
  \And
  Tim Siebert$^*$ \\
  Humboldt-Universit\"at zu Berlin and\\
  Zuse Institute Berlin\\
  Berlin, Germany \\
  \texttt{tim.siebert@hu-berlin.de} \\
  \And
  Marius Zeinhofer\\
  ETH Zurich \\
  Zurich, Switzerland, \\
  \texttt{marius.zeinhofer@sam.math.ethz.ch}
  \And
  Andrea Walther \\
  Humboldt-Universit\"at zu Berlin and \\
  Zuse Institute Berlin \\
  Berlin, Germany \\
  \texttt{andrea.walther@math.hu-berlin.de} \\
}

\definecolor{darkgreen}{rgb}{0,0.6,0}

\newcommand{\colorcTM}{tab-green}
\newcommand{\colorTM}{tab-orange}

\begin{document}

\maketitle

\begin{abstract}
  Computing partial differential equation (PDE) operators via nested backpropagation is expensive, yet popular, and severely restricts their utility for scientific machine learning.
  Recent advances, like the forward Laplacian and randomizing Taylor mode automatic differentiation (AD), propose forward schemes to address this.
  We introduce an optimization technique for Taylor mode that ``collapses''  derivatives by rewriting the computational graph, and demonstrate how to apply it to general linear PDE operators, and randomized Taylor mode.
  The modifications simply require propagating a sum up the computational graph, which could---or should---be done by a machine learning compiler, without exposing complexity to users.
  We implement our collapsing procedure and evaluate it on popular PDE operators, confirming it accelerates Taylor mode and outperforms nested backpropagation.
\end{abstract}

\section{Introduction}\label{sec:introduction}
Using neural networks to learn functions constrained by physical laws is a popular trend in scientific machine learning \cite{carleo2017solving,pfau2020ab,hermann2020deep,hu2024hutchinson,karniadakis2021physics, raissi2019physics,sun2020global}.
Typically, the Physics is encoded through partial differential equations (PDEs) that the neural net must satisfy.
The associated loss functions require evaluating differential operators \wrt the net's input, rather than weights.
Evaluating these differential operators remains a computational challenge, especially if they contain high-order derivatives.

\paragraph{Computing PDE operators.} Two important fields that build on PDE operators are variational Monte-Carlo (VMC) simulations and Physics-informed neural networks (PINNs).
VMC employs neural networks as ansatz for the Schr\"odinger equation \cite{carleo2017solving, pfau2020ab, hermann2020deep} and demands computing the net's Laplacian (the Hessian trace) for the Hamiltonian's kinetic term.
PINNs represent PDE solutions as a neural net and train it by minimizing the residuals of the governing equations \cite{raissi2019physics, karniadakis2021physics}. For instance, Kolmogorov-type equations like the Fokker-Planck and Black-Scholes equation require weighted second-order derivatives on high-dimensional spatial domains \cite{hu2024hutchinson, sun2024dynamical}.
Other PINNs for elasticity problems use the biharmonic operator \cite{vahab_physics-informed_2022, hu2024hutchinson, vikas_biharm, shi2024stochastic}, which contains fourth-order derivatives.

\paragraph{Is backpropagation all we need?}
Although nesting first-order automatic differentiation (AD) to compute high-order derivatives scales exponentially \wrt the degree in time and memory \cite[][\S3.2]{shi2024stochastic}, this approach is common practice: it is easy to implement in ML libraries, and their backpropagation is highly optimized.
A promising alternative is \emph{Taylor mode AD}~\cite[or simply \emph{Taylor mode},][\S13]{griewank2008evaluating}, introduced to the ML community in \citeyear{bettencourt2019taylor}, which scales polynomially \wrt the degree in time and memory \cite{griewank_evaluating_1999}.
However, we observe empirically that vanilla Taylor mode may not be enough to beat nesting (\cref{fig:vanilla-taylor-not-enough}): evaluating the Laplacian of a 5-layer MLP, using JAX's \emph{Taylor mode is 50\% slower than nested backpropagation} that computes, then traces, the Hessian via Hessian-vector products \cite{pearlmutter1994fast}.
This calls into question the relevance of Taylor mode for computing common PDE operators.

\paragraph{The advent of forward schemes.}
Recent works have successfully demonstrated the potential of modified forward propagation schemes, though.
For the Laplacian, \citet{li2023forward, li2024dof} developed a special forward propagation framework called the \emph{forward Laplacian}, whose JAX implementation \cite{gao2023folx} is \emph{roughly twice as fast as nested first-order AD} (\cref{fig:vanilla-taylor-not-enough}).
While the forward Laplacian does not rely on Taylor mode, recent work pointed out a connection \cite{dangel2024kroneckerfactored}; it remains unclear, though, if efficient forward schemes exist for other differential operators, and how they relate to Taylor mode.
Concurrently, \citet{shi2024stochastic} derived stochastic approximations of differential operators in high dimensions by evaluating Taylor mode along suitably sampled random directions.

Irrespective of stochastic or exact computation, at their core, these popular PDE operators are \emph{linear}: we must evaluate derivatives along multiple directions, then sum them.
Based on this linearity, we identify an optimization technique to rewrite the computational graph of standard Taylor mode that is applicable to general linear PDE operators and randomized Taylor mode:

\begin{figure*}[!t]
  \centering
  \begin{minipage}[b]{0.42\linewidth}
    \centering
    \input{figures/vanilla_taylor_not_enough.tex}

    \caption{\textbf{$\blacktriangle$ Vanilla Taylor mode is not enough to beat nested 1\textsuperscript{st}-order AD.}
      Illustrated for computing the Laplacian of a $\mathrm{tanh}$-activated $50 \!\to\! 768 \!\to\! 768 \!\to\! 512 \!\to\! 512 \!\to\! 1$
      MLP with JAX (+ \texttt{jit}) on GPU (details in \Cref{sec:jax-benchmark}).
      We show how to automatically obtain the specialized forward Laplacian through simple graph transformations that ``collapse`` vanilla Taylor mode.
    }\label{fig:vanilla-taylor-not-enough}

    \vspace{0.25ex}
    \caption{\textbf{$\blacktriangleright$ Collapsed Taylor mode directly propagates the sum of highest degree coefficients.}
      Visualized for pushing 4 $K$-jets through a $\sR^5 \!\to\! \sR^3 \!\to\! \sR$ function ($K=2$ yields the forward Laplacian).
      }\label{fig:visual-abstract}
  \end{minipage}
  \hfill
  \begin{minipage}[b]{0.57\linewidth}
    \centering
    \newcommand{\drawgridrectangle}[4]{%
  \begin{tikzpicture}[scale=#4]
    \pgfmathsetmacro{\ymax}{#1}
    \pgfmathsetmacro{\xmax}{#2}

    \fill[#3] (0,0) rectangle (\xmax,\ymax);

    \draw[white, line width=#4*3pt] (0,0) rectangle (\xmax,\ymax);

    \pgfmathsetmacro{\xsteps}{#2}
    \foreach \x in {1,...,\xsteps} {
      \draw[white, line width=#4*3pt] (\x,0) -- (\x,\ymax);
    }

    \pgfmathsetmacro{\ysteps}{#1}
    \foreach \y in {1,...,\ysteps} {
      \draw[white, line width=#4*7pt] (0,\y) -- (\xmax,\y);
    }
  \end{tikzpicture}%
}

\centering
\newsavebox{\visualAbstract}
\savebox{\visualAbstract}{
  \begin{tikzpicture}
    \matrix [%
    matrix of nodes,%
    ampersand replacement=\&,%
    nodes={anchor=center, align=center},%
    column sep=5ex,%
    row sep=1ex,%
    ] (taylor)
    {
      \drawgridrectangle{1}{5}{gray!25!white}{0.33}
      \& \drawgridrectangle{1}{3}{gray!25!white}{0.33}
      \& \drawgridrectangle{1}{1}{gray!25!white}{0.33}
      \\[-1.5ex]
      $\vx_0$ \& $\vh_0$ \& $\vg_0$
      \\
      \drawgridrectangle{4}{5}{gray!50!white}{0.33}
      \& \drawgridrectangle{4}{3}{gray!50!white}{0.33}
      \& \drawgridrectangle{4}{1}{gray!50!white}{0.33}
      \\[-1.5ex]
      $\{\vx_{1, r}\}$ \& $\{\vh_{1, r}\}$ \& $\{\vg_{1, r}\}$
      \\[0.5ex]
      \drawgridrectangle{1}{5}{white}{0.33}
      \& \drawgridrectangle{1}{3}{white}{0.33}
      \& \drawgridrectangle{1}{1}{white}{0.33}
      \\[0.5ex]
      \\
      \drawgridrectangle{4}{5}{gray!75!white}{0.33}
      \& \drawgridrectangle{4}{3}{gray!75!white}{0.33}
      \& \drawgridrectangle{4}{1}{gray!75!white}{0.33}
      \\[-1.5ex]
      $\{\vx_{K-1, r}\}$ \& $\{\vh_{K-1, r}\}$ \& $\{\vg_{K-1, r}\}$
      \\[0.5ex]
      \drawgridrectangle{4}{5}{tab-orange}{0.33}
      \&
      \drawgridrectangle{4}{3}{tab-orange}{0.33}
      \&
      \drawgridrectangle{4}{1}{tab-orange}{0.33}
      \\[-1.5ex]
      $\{\vx_{K, r}\}$ \& $\{\vh_{K,r}\}$ \& $\{\vg_{K,r}\}$
      \\
      \drawgridrectangle{1}{5}{tab-green}{0.33}
      \& \drawgridrectangle{1}{3}{tab-green}{0.33}
      \& \drawgridrectangle{1}{1}{tab-green}{0.33}
      \\[-1ex]
      \textcolor{tab-green}{$\sum_r \vx_{K,r}$}
      \& \textcolor{tab-green}{$\sum_r \vh_{K,r}$}
      \& \textcolor{tab-green}{$\sum_r \vg_{K,r}$}
      \\
    };

    \node at (taylor-5-1) {\vdots};
    \node at (taylor-5-2) {\vdots};
    \node at (taylor-5-3) {\vdots};

    \pgfmathsetmacro{\K}{5}
    \pgfmathsetmacro{\L}{2}

    \foreach \l in {1,...,\L}{
      \pgfmathsetmacro{\lother}{int(\l+1)}
      \foreach \k in {1,...,\K} {
        \pgfmathsetmacro{\row}{int(2*\k-1)}
        \foreach \kother in {\k,...,\K} {
          \pgfmathsetmacro{\rowother}{int(2*\kother-1)}
          \draw[-Stealth, line width=1pt, gray!50!white] (taylor-\row-\l.east) -- (taylor-\rowother-\lother.west);
        }
      }
    }

    \coordinate (arrowStart) at ($(taylor-1-1.north)+(0,3.5ex)$);
    \coordinate (arrowEnd) at ($(taylor-1-3.north east)+(0,3.5ex)$);
    \draw[-Stealth, line width=2pt, black] (arrowStart) to node [midway, fill=white, align=center] {\textbf{Taylor forward} \\ \textbf{propagation}} (arrowEnd);

    \node [minimum width=8ex, align=center, left=1.5ex of taylor-1-1] (zero) {0};
    \node [minimum width=8ex, align=center, left=1.5ex of taylor-3-1] {1};
    \node [minimum width=8ex, align=center, left=1.5ex of taylor-5-1] {$\vdots$};
    \node [minimum width=8ex, align=center, left=1.5ex of taylor-7-1] {$K-1$};
    \node [minimum width=8ex, align=center, left=1.5ex of taylor-9-1] {$K$};
    \node [minimum width=8ex, align=center, left=1.5ex of taylor-11-1, yshift=-8pt] {$K$ \\ \textbf{(ours)}};

    \node [align=center] (coefficientLabel) at ($(zero)+(0, 5.5ex)$) {\textbf{Derivative}\\\textbf{degree}};

    \draw[rounded corners, line width=2pt, red] (taylor-9-1.north west) rectangle (taylor-10-3.south east);
    \draw[white, fill=white] ($(taylor-10-3.south east)+(-0.3ex, 2ex)$) rectangle ($(taylor-10-3.south east)+(0.3ex, 11.25ex)$);

    \draw [-Stealth, line width=1.25pt, red]
    ($(taylor-10-3.south east)+(0, 2.75ex)$)
    -- ($(taylor-10-3.south east)+(0, 5.75ex)$);

    \draw [-Stealth, line width=1.25pt, red]
    ($(taylor-10-3.south east)+(0, 10.5ex)$)
    -- ($(taylor-10-3.south east)+(0, 7.5ex)$);
    ;

    \draw [red, line width=5pt]
    ($(taylor-10-3.south east)+(-0.7ex, 6.625ex)$)
    -- ($(taylor-10-3.south east)+(0.7ex, 6.625ex)$);
  \end{tikzpicture}
}

\resizebox{\linewidth}{!}{
  \begin{tikzpicture}
    \node {\usebox{\visualAbstract}};
  \end{tikzpicture}
}
  \end{minipage}
\end{figure*}

\begin{enumerate}[leftmargin=0.5cm]
\item \textbf{We propose optimizing standard Taylor mode by collapsing the highest Taylor coefficients,} directly \textcolor{tab-green}{\bfseries propagating their sum}, rather than \textcolor{tab-orange}{\bfseries propagating then summing} (\cref{fig:visual-abstract}).
Our approach contains the forward Laplacian as special case, is applicable to randomized Taylor mode, and also general linear PDE operators, which we show using the techniques from \citet{griewank_evaluating_1999}.

\item \textbf{We show how to collapse standard Taylor mode by simple graph rewrites based on linearity.}
  This leads to a clean separation of concepts:
  Users can build their computational graph using standard Taylor mode, then rewrite it to collapse it.
  Due to the simple nature of our proposed rewrites, they could easily be absorbed into the just-in-time (JIT) compilation of ML frameworks without introducing a new interface or exposing complexity to users.

\item \textbf{We empirically demonstrate that collapsing Taylor mode accelerates standard Taylor mode.}
We implement a Taylor mode library\footnote{Available at \url{https://github.com/f-dangel/torch-jet}.} for PyTorch \cite{paszke2019pytorch} that realizes the graph simplifications with \texttt{torch.fx} \cite{reed2022torch}.
On popular PDE operators, we empirically find that, compared to standard Taylor mode, collapsed Taylor mode achieves superior performance that is well-aligned with the theoretical expectation, while consistently outperforming nested first-order AD.
\end{enumerate}

Our work takes an important step towards the broader adoption of Taylor mode as viable alternative to nested first-order AD for computing PDE operators, while being as easy to use.

\section{Background: Introduction to Taylor Mode AD}\label{sec:background}
Taylor mode AD (or, simply, Taylor mode) computes higher-order derivatives---as needed, \eg, for PDE operators---through propagation of Taylor coefficients according to the chain rule.

\paragraph{Scalar case.}
To illustrate Taylor mode, consider the scalar function $f: \sR \to \sR$ and extend the input variable $x$ to a path $x(t)$ with $x(0) = x_0$, whose form is a univariate Taylor polynomial of degree $K$, $\smash{x(t) = \sum_{k=0}^K \frac{t^k}{k!} x_k}$ with $x_k$ the $k$-th Taylor coefficient.
If $f$ is smooth enough, we can evaluate Taylor coefficients of the transformed path $\smash{f(x(t)) = \sum_{k=0}^K \frac{t^k}{k!} f_k}$ with $\smash{f_k \coloneqq \frac{\mathrm{d}^k}{\mathrm{d}t^k} f(x(t)) |_{t=0}}$.
The chain rule provides the coefficients' propagation rules.
\Eg, for degree $K=3$ we get
\begin{align}
  \label{eq:taylor-mode-scalar}
  \begin{matrix*}[l]
    f_0 = f(x_0)\,,
    \\[0.75ex]
    f_1 = \partial f(x_0) x_1\,,
  \end{matrix*}
  \qquad
  \begin{matrix*}[l]
    f_2 = \partial^2 f(x_0)x_1^2 + \partial f(x_0) x_2\,,
    \\[0.75ex]
    f_3
    = \partial^3 f(x_0)x_1^3 + 3 \partial^2 f(x_0) x_1 x_2 + \partial f(x_0) x_3\,.
  \end{matrix*}
\end{align}
\citet{faa1857note} provided the general formula for $f_k$, and \citet{fraenkel1978formulae} extended it to the multivariate case \cite[see also][]{arbogast1800calcul,hardy2006combinatorics}.
It serves as foundation for Taylor mode to compute higher-order derivatives \citep[\eg,][\S13]{griewank2008evaluating}:
setting $x_1 = 1, x_2 = x_3 = 0$ yields $f_1 = \partial f(x_0), f_2 = \partial^2 f(x_0), f_3 = \partial^3 f(x_0)$.
We call the univariate Taylor polynomial of a function $x(t)$ of degree $K$, represented by the coefficients $(x_0, \dots, x_K)$, the \emph{$K$-jet of $x$}, following the terminology of JAX's Taylor mode \cite{bettencourt2019taylor}.

\paragraph{Notation for multivariate case.}
We consider the general case of computing higher-order derivatives, \eg, PDE operators, of a vector-to-vector function $\vf: \sR^D \to \sR^C$.
This requires additional notation to generalize \cref{eq:taylor-mode-scalar}.
Given $K$ vectors $\vv_1, \dots, \vv_K \in \sR^D$, we write their tensor product as
\begin{equation*}
  \otimes_{k=1}^K \vv_k = \vv_1 \otimes \ldots \otimes \vv_K
  \in ( \sR^D )^{\otimes K}
  \quad
  \text{with entries}
  \quad
  \left[\otimes_{k=1}^K \vv_k\right]_{d_1, \dots, d_K}
  = [\vv_1]_{d_1} \cdot \ldots \cdot [\vv_K]_{d_K}
\end{equation*}
for $d_1, \dots, d_K \in \{1, \dots, D\}$, and compactly write $\vv^{\otimes K} = \otimes_{k=1}^K \vv$.
We define the inner product of two tensors $\smash{\tA, \tB \in (\sR^{D})^{\otimes K}}$ as the Euclidean inner product of their flattened versions
\begin{align}\label{eq:derivative-tensor-scalar-product}
  \left\langle
  \tA, \tB
  \right\rangle
  \coloneqq
  \sum_{d_1}
  \sum_{d_2}
  \dots
  \sum_{d_K}
  \left[\tA\right]_{d_1, d_2, \dots, d_K}
  \left[\tB\right]_{d_1, d_2, \dots, d_K} \in \sR\,.
\end{align}
We allow broadcasting in \cref{eq:derivative-tensor-scalar-product}: if one tensor has more dimensions but matching trailing dimensions, we take the inner product for each component of the leading dimensions.
This allows to express contractions with derivative tensors of vector-valued functions, \eg,
contracting the $k$-th derivative tensor $\partial^k \vf(\vx_0) \in \sR^C \times (\sR^D)^{\otimes k}$, such that $\langle \tA, \partial^k \vf(\vx_0) \rangle \in \sR^C$.

\paragraph{Multivariate case \& composition.}
Evaluating the $K$-jet of $\vf$ at $\vx_0 \in \sR^D$ starts with the extension of $\vx_0$ to a smooth path $\vx: \sR \to \sR^D$ with $\vx(0) = \vx_0$.
Formally, the $K$-jet of $\vf$ is defined as
\begin{align*}
  J^K \vf : \sR \to \sR^C\,,
  \quad (J^K \vf)(t) := \sum_{k=0}^K \frac{t^k}{k!} \vf_k
  \quad \text{with} \quad
  \vf_k := \frac{\mathrm{d}^k}{\mathrm{d}t^k} \vf(\vx(t)) |_{t=0}
\end{align*}
and requires the $K$-jet of $\vx$, $(J^K \vx)(t) := \sum_{k=0}^K \frac{t^k}{k!} \vx_k$.
As we are interested in the coefficients, we will slightly abuse the $K$-jet as mapping $(\vx_0, \dots, \vx_K) \mapsto (\vf_0, \dots, \vf_K)$  (see \cref{fig:utp} for an illustration).

As is common for AD, propagating the coefficients is broken down into
composing $\vf$ of atomic functions with known derivatives and the chain rule.
In the simplest case, let $\vf = \vg \circ \vh: \sR^D \to \sR^I \to \sR^C$ for two elemental functions $\vg, \vh$.
Given the input $K$-jet for $\vx$, the coefficients $\vh_k = \smash{\frac{\mathrm{d}^k}{\mathrm{d}t^k}\vh(\vx(t))} |_{t=0}$ follow from the generalized Faà di Bruno formula (spelled out for some $k$s in \cref{sec:faa-di-bruno-cheatsheet})
\begin{align}
  \label{eq:faa-di-bruno}
  \vh_k
  =
  \sum_{\sigma \in \partitioning(k)}
  \nu(\sigma)
  \left<
  \partial^{|\sigma|} \vh,
  \tensorprod{s \in \sigma} \vx_s
  \right>
  \quad
  \text{with}
  \quad
  \nu(\sigma)
  =
  \frac{k!}{
    \left(
      \prod_{s \in \sigma
      }
      n_s!
    \right)
    \left(
      \prod_{s \in \sigma}
      s!
    \right)
  }\,.
\end{align}
Here, $\partitioning(k)$ is the integer partitioning of $k$ (a set of sets), $\nu$ is a multiplicity function, and $n_s$ counts occurrences of $s$ in a set $\sigma$ (\eg, $n_1(\{1,1,3\})\!=\!2$ and $n_3 \!= \!1$).
Propagating the $\vh_k$s through $\vg$ results in the $K$-jet for $\vf$.
In summary, the propagation scheme is (with $\vx_k \in \sR^D$, $\vh_k \in \sR^I$, $\vf_k \in \sR^C$)
\begin{align}\label{eq:taylor-mode-composition}
  \begin{split}
    &\begin{pmatrix*}
      \vx_0
      \\
      \vx_1
      \\
      \vx_2
      \\
      \vdots
      \\
      \vx_K
    \end{pmatrix*}
      \overset{\text{(\ref{eq:faa-di-bruno})}}{\to}
      \begin{pmatrix*}[l]
        \vh_0 \!=\!  \vh(\vx_0)
        \\
        \vh_1 \!=\!  \left<
        \partial \vh(\vx_0),
        \vx_1
        \right>
        \\
        \vh_2 \!=\! \left<
        \partial^2 \vh(\vx_0),
        \vx_1^{\otimes 2}
        \right>
        \!+\!
        \left <
        \partial \vh(\vx_0),
        \vx_2
        \right>
        \\
        \vdots
        \\
        \vh_K \!=\!
        \displaystyle \sum_{
        \mathclap{
        \sigma \in \partitioning(K)
        }
        }
        \nu(\sigma) \left<
        \partial^{|\sigma|} \vh(\vx_0),
        \tensorprod{s \in \sigma} \vx_s
        \right>\!\!\!
      \end{pmatrix*}
    \\
    &\overset{\text{(\ref{eq:faa-di-bruno})}}{\to}
      \left(\!\!\!
      \begin{array}{l}
        \vg_0 \!=\!  \vg(\vh_0)
        \\
        \vg_1 \!=\! \left<
        \partial \vg(\vh_0),
        \vh_1
        \right>
        \\
        \vg_2 \!=\! \left<
        \partial^2 \vg(\vh_0),
        \vh_1^{\otimes 2}\right>
        \!+\!
        \left< \partial \vg(\vh_0),
        \vh_2
        \right>
        \\
        \vdots
        \\
        \vg_K \!=\!
        \displaystyle\sum_{
        \mathclap{
        \sigma \in \partitioning(K)
        }
        }
        \nu(\sigma) \left<
        \partial^{|\sigma|} \vg(\vh_0),
        \tensorprod{s \in \sigma} \vh_s
        \right>
      \end{array}
      \!\!\!\!
      \right)
      \overset{\eqref{eq:taylor-mode-scalar}}{=}
      \left(\!\!\!
      \begin{array}{l}
        \vf_0 \!=\!  \vf(\vx_0)
        \\
        \vf_1 \!=\! \left<
        \partial \vf(\vx_0),
        \vx_1
        \right>
        \\
        \vf_2 \!=\! \left<
        \partial^2 \vf(\vx_0),
        \vx_1^{\otimes 2}
        \right>
        \!+\!
        \left< \partial \vf(\vx_0),
        \vx_2
        \right>
        \\
        \vdots
        \\
        \vf_K \!=\!
        \displaystyle\sum_{
        \mathclap{
        \sigma \in \partitioning(K)
        }
        }
        \nu(\sigma) \left<
        \partial^{|\sigma|} \vf(\vx_0),
        \tensorprod{s \in \sigma} \vx_s
        \right>
      \end{array}
      \!\!\!\!
      \right)
  \end{split}
\end{align}
which describes the forward propagation of a \emph{single} $K$-jet.
However, computing popular PDE operators requires propagating \emph{multiple} $K$-jets in parallel, then summing their results.
We propose to pull this accumulation inside Taylor mode's propagation scheme, thereby collapsing it.

\begin{figure}[!t]
  \centering
  \begin{minipage}[t]{0.63\linewidth}
    \centering
    \vspace{0pt}
   \begin{tikzpicture}
    \tikzset{box/.style={rectangle, rounded corners, draw=black, inner sep=3pt, very thick}}
    \node[align=center, box] (topleft) {Extent input to \\ smooth path
      $\vx(t)$};

    \node[align=center, right=2.7cm of topleft, box] (topright) {Path in output \\ space $\vf(\vx(t))$};
    \draw [-Latex] (topleft.east) to node [midway, above] {$\vf$} (topright.west);

    \node[align=center, below=0.9cm of topright, box, draw=tab-orange, fill=tab-orange!25!white] (bottomright) {%
      $K$-jet \; $\sum_{k=0}^K \frac{t^k}{k!} \vf_k$
      \\
      as $(\vf_0, \dots, \vf_K)$
    };
    \draw [-Latex] (topright.south) to node [midway, right] {$J^K$}
    (bottomright.north);

    \node[align=center, below=0.9cm of topleft, box, draw=tab-orange, fill=tab-orange!25!white] (bottomleft) {%
        $K$-jet \; $\sum_{k=0}^K \frac{t^k}{k!} \vx_k$
        \\
        as $(\vx_0, \dots, \vx_K)$
      };
    \draw [-Latex] (topleft.south) to node [midway, left] {$J^K$} (bottomleft.north);

    \draw [-Latex, \colorTM, align=center, very thick] (bottomleft.east) to node [midway, above, tab-orange] {\color{\colorTM}Taylor mode} (bottomright.west);
  \end{tikzpicture}
  \end{minipage}
  \hfill
  \begin{minipage}[t]{0.35\linewidth}
    \caption{\textbf{Taylor mode propagates Taylor coefficients of a path in input space.}
    This results in the function-transformed path's Taylor coefficients.
    The Taylor expansion of degree $K$ is called a $K$-jet; hence Taylor mode propagates the input $K$-jet to the output $K$-jet.}
  \label{fig:utp}
  \end{minipage}
\end{figure}

\section{Collapsing Taylor Mode AD}\label{sec:methodology}
We now describe how to collapse the Taylor mode AD computation of popular linear PDE operators and their stochastic approximations proposed in \cite{shi2024stochastic}, and provide a general recipe for computing and collapsing general linear differential operators by interpolation, using earlier work from \citet{griewank_evaluating_1999}.
At its core, our procedure
uses the linearity of the highest Taylor coefficient's propagation rule.
It allows to collapse coefficients along multiple directions and directly \textcolor{tab-green}{\bfseries propagate their sum}, rather than \textcolor{tab-orange}{\bfseries propagating then summing},
yielding substantial reductions in computational cost.

\subsection{Exploiting Linearity to Collapse Taylor Mode AD}

To derive our proposed method, we start with a sum of $K$th-order directional derivatives of the function $\vf$ along $R$ directions $\{\vv_r\}_{r=1}^R$, which is a common building block for all our PDE operators:
\begin{align}\label{eq:sum-k-directional}
  {\color{tab-orange}
  \sum_{r=1}^R
  }
  \left<
  \partial^{K} \vf(\vx_0),
  \vv_r^{\otimes K}
  \right> \, \in \sR^C
\end{align}
(\eg, the exact Laplacian uses $K = 2$, $R = \dim(\vx_0) = D$, and the unit vectors $\vv_r = \ve_r\in \R^D$ as directions; see \cref{sec:second-order-operators} below).
Instead of nesting $K$ calls to 1\textsuperscript{st}-order AD, we can use $K$-jets to calculate each summand of \cref{eq:sum-k-directional} with Taylor mode.
In total, we need $R$ $K$-jets, and have to set the $r$-th jet's coefficients to $\vx_{0, r} = \vx_0$, $\vx_{1, r} = \vv_r$ and $\vx_{2, r} = \ldots = \vx_{K, r} = \vzero$ (\cref{eq:sum-taylor-mode-naive} applies this to \cref{eq:taylor-mode-composition}).

Standard Taylor mode propagates $1 + KR$ vectors through every node of the computational graph (the $0$th component is shared across all jets, see \cref{fig:visual-abstract}).
This gives the output jets $\smash{\{\{\vf_{k,r}\}_{k=1}^K\}_{r=1}^{R}}$, from which we only select the highest-degree coefficients $\smash{\{\vf_{K,r}\}_{r=1}^R}$, then sum them to obtain \cref{eq:sum-k-directional}.

The approach we propose here exploits that the $K$-th derivative of $\vg \circ \vh$ is $\partial \vg$ times the $K$-th derivative of $\vh$ plus other lower-order terms in $\vh$. Therefore, $\vg \circ \vh$ is linear in the $K$-th derivative of $\vh$. Mathematically speaking, there is a special element in the set of integer partitions $\partitioning(K)$, namely the trivial partition $\tilde{\sigma}= \{K\}$, which contributes the term $ \nu(\tilde{\sigma}) \left< \partial \vg(\vh_0), \vh_{K,r} \right>$ to \cref{eq:faa-di-bruno}.
This is the only term that uses the input jet's highest coefficient $\vh_{K,r}$, and its dependency is \emph{linear}.
Separating it in the highest coefficient's forward propagation, we get (using $\nu(\tilde{\sigma}) = 1$)
\begin{align}
  \nonumber
  \sum_{r=1}^R\vf_{K, r}
  =
  \sum_{r=1}^R\vg_{K,r}
  &=
  {
  \color{tab-orange}
  \sum_{r=1}^R
  }
  \sum_{
    \sigma \in \partitioning(K) \setminus \{\tilde{\sigma}\}
  }\!\!\!\!
  \nu(\sigma) \left<\!
    \partial^{|\sigma|} \vg(\vh_0),
    \tensorprod{s \in \sigma} \vh_{s, r}\!
  \right>
  \!+\!
  {
  \color{tab-orange}
  \sum_{r=1}^R
  }
  \left<
    \partial \vg(\vh_0),
    \vh_{K, r}
  \right>
  \shortintertext{and since the propagation rule is linear \wrt $\vh_{K,r}$, we can pull the summation inside:}
  {
  \color{tab-green}
  \sum_{r=1}^R
  }
  \vg_{K,r}
  &=
  {\color{tab-orange}
  \sum_{r=1}^R
  }
  \sum_{
    \sigma \in \partitioning(K) \setminus \{\tilde{\sigma}\}
  }\!\!\!\!
  \nu(\sigma) \left<\!
    \partial^{|\sigma|} \vg(\vh_0),
    \tensorprod{s \in \sigma} \vh_{s, r}\!
  \right>
  \!+\!
  \left<\!
    \partial \vg(\vh_0),
    {
    \color{tab-green}\sum_{r=1}^R}
    \vh_{K, r}
  \right>
  \,.\label{eq:faa-di-bruno-expanded}
\end{align}
This is the key insight of our work:
\emph{The summed highest-degree output coefficients depend on the summed highest-degree input coefficients} (as well as all lower-degree coefficients).
The reason is \emph{linearity} in Fa\`a di Bruno's formula.
Hence, to compute the sum $\sum_r \vg_{K, r}$ we can directly propagate the sum $\sum_r \vh_{K,r}$, collapsing coefficients over all directions.
We call this \emph{collapsed Taylor mode AD}.

Collapsed Taylor mode propagates only $1 + (K - 1)R + 1$ vectors through every node in the computational graph (see \cref{fig:visual-abstract} and \cref{eq:sum-taylor-mode-efficient} which applies this to \cref{eq:faa-di-bruno}).
These savings of $R-1$ coefficients are significant improvements over standard Taylor mode, as we show below.
In the following, we discuss how to collapse the Taylor mode computation of various PDE operators.

\subsection{Linear Second-order Operators}\label{sec:second-order-operators}

\paragraph{Laplacian.} The Laplace operator plays a central role in Physics and engineering, including electrostatics, fluid dynamics, heat conduction, and quantum mechanics \cite{foulkes2001quantum, pfau2020ab}.
It contains the Hessian trace of each element of a function, \ie, for $\vf: \sR^D \to \sR^C$, it is
\begin{subequations}
  \begin{align}\label{eq:laplacian}
    \underbrace{
    \Delta \vf(\vx_0)
    }_{\in \sR^C}
    :=
    \left<
    \partial^2 \vf(\vx_0), \mI_D
    \right>
    \quad
    \begin{cases}
      \displaystyle
      = \sum_{d=1}^D \left<
      \partial^2 \vf(\vx_0),
      \ve_d^{\otimes 2}
      \right>
      &\text{(exact)}
      \\
      \displaystyle
      \overset{\text{\cite{shi2024stochastic}}}{\approx}
      \frac{1}{S} \sum_{s=1}^S
      \left<
      \partial^2 \vf(\vx_0),
      \vv_s^{\otimes 2}
      \right>
      &\text{(stochastic)}
    \end{cases}
  \end{align}
  with the $d$-th standard basis vector $\ve_d$ used for exact computation, and $S$ random vectors $\vv_s$ drawn \iid from a distribution with unit variance (\eg Rademacher or standard Gaussian) for stochastic estimation.
  By pattern-matching \cref{eq:laplacian} with \cref{eq:sum-k-directional} we conclude that $K=2$, and the following choices for computing the Laplacian with standard Taylor mode:
  \begin{align}
    \begin{array}{rlrlrlll}
      \{(\vx_{0,d} & \!\!\!\!= \vx_0, &\vx_{1, d} & \!\!\!\!= \ve_d, & \vx_{2,d} &\!\!\!\!= \vzero)\}_{d=1}^D
                                                                   \quad
      &{\color{tab-orange} 1 + D + D\, \text{vectors}}
        \quad
      &\text{(exact)}
      \\[1ex]
      \{(\vx_{0,s} & \!\!\!\!= \vx_0, &\vx_{1, s} & \!\!\!\!= \vv_s, &\vx_{2,s} & \!\!\!\!= \vzero)\}_{s=1}^S
                                                                  \quad
      &{\color{tab-orange} 1 + S + S\, \text{vectors}}
        \quad
      &\text{(stochastic)}
    \end{array}.
  \end{align}
\end{subequations}
Collapsing standard Taylor mode yields {\color{tab-green}$1 + D + 1$} (exact) and {\color{tab-green}{$1 + S + 1$}} (stochastic) vectors.
In fact, the collapsed Taylor mode for the exact Laplacian is the forward Laplacian from \citet{li2023forward} (see \cref{eq:laplacian-efficient} for detailed presentation of the forward propagation).
Note how we can seamlessly also collapse the stochastic approximation over the sampled directions, which is currently not done.

\paragraph{Weighted Laplacian.}
A natural generalization of the Laplacian involves contracting with a positive semi-definite matrix $\mD = \msigma \msigma^\top \in \mathbb R^{D\times D}$ rather than the identity. $\mD$ may represent the diffusion tensor in Kolmogorov-type PDEs like the Fokker-Planck equation \cite{hu2024hutchinson}, and $\msigma$ can depend on $\vx_0$ \cite{fa2011solution}.
The weighted Laplacian contains the weighted Hessian's trace $\Tr(\msigma \msigma^{\top} \partial^2 [\vf]_c)$ for each output element $c$ of $\vf$.
If $\rank(\mD) = R$ and therefore $\msigma = (\vs_1, \dots, \vs_R) \in \sR^{D \times R}$, it is
\begin{subequations}
  \begin{align}\label{eq:weighted-laplacian}
    \underbrace{
    \Delta_\mD \vf(\vx_0)
    }_{\in \sR^C}
    :=
    \left<
    \partial^2 \vf(\vx_0), \mD
    \right>
    \quad
    \begin{cases}
      \displaystyle
      = \sum_{r=1}^R
      \left< \partial^2 \vf(\vx_0), \vs_r^{\otimes 2} \right>
      & \text{(exact)}
      \\
      \displaystyle
      \overset{\text{\cite{hu2024hutchinson}}}{\approx}
      \frac{1}{S}
      \sum_{s=1}^S
      \left< \partial^2 \vf(\vx_0), (\msigma \vv_s)^{\otimes 2} \right>
      & \text{(stochastic)}
    \end{cases}
    .
  \end{align}
  Computing it requires evaluating the following 2-jets with standard Taylor mode:
  \begin{align}
    \begin{array}{rlrlrlll}
      \{(\vx_{0,r} & \!\!\!\!= \vx_0, &\vx_{1, r} & \!\!\!\!= \vs_r, & \vx_{2,r} &\!\!\!\!= \vzero)\}_{r=1}^R
                                                                   \quad
      &{\color{tab-orange} 1 + R + R\, \text{vectors}}
        \quad
      &\text{(exact)}
      \\[1ex]
      \{(\vx_{0,s} &\!\!\!\!= \vx_0, &\vx_{1, s} & \!\!\!\!= \msigma \vv_s, &\vx_{2,s} & \!\!\!\!= \vzero)\}_{s=1}^S
                                                                          \quad
      &{\color{tab-orange} 1 + S + S\, \text{vectors}}
        \quad
      &\text{(stochastic)}
    \end{array}.
  \end{align}
\end{subequations}
Our collapsed Taylor mode uses {\color{tab-green}$1 + R + 1$} (exact) and {\color{tab-green}{$1 + S + 1$}} (stochastic) vectors.
This yields the modified forward Laplacian from \citet{li2024dof}; collapsing the stochastic variant speeds up the Hutchinson trace estimator from \citet{hu2024hutchinson}.
For indefinite $\mD$, we can simply apply this scheme to the positive and negative eigen-spaces (however, such weightings are not used in practise).

\subsection{Collapsed Taylor Mode for Arbitrary Mixed Partial Derivatives}
So far, we discussed operators that result from contracting the second-order derivative tensor with a coefficient matrix ($\mI$ or $\mD$) that can conveniently be written as sum of vector outer products.
For orders higher than two, the coefficient tensor can in general \emph{not} easily be decomposed as such.
Hence, we extend our framework to also cover differential operators containing mixed-partial derivatives by evaluating a suitable family of jets using the interpolation result of \citet{griewank_evaluating_1999}.
As illustrative example, we will use the biharmonic operator with a 4-dimensional coefficient tensor:
\begin{align}
  \label{eq:biharm}
  \underbrace{
  \Delta^2 \vf(\vx_0)
  }_{\in \sR^C}
  \quad
  \begin{cases}
    \displaystyle
    =
    \sum_{d_1=1}^D \sum_{d_2=1}^D
    \left<
    \partial^4 \vf(\vx_0),
    \ve_{d_1}^{\otimes 2} \otimes \ve_{d_2}^{\otimes 2}
    \right>
    & \text{(exact)}
    \\
    \displaystyle
    \overset{\text{\cite{shi2024stochastic}}}{\approx}
    \frac{1}{3S}
    \sum_{s=1}^S
    \left< \partial^4 \vf(\vx_0), \vv_s^{\otimes 4} \right>
    & \text{(stochastic)}
  \end{cases}\,.
\end{align}
We can directly collapse the stochastic version: draw $S$ standard normal vectors $\vv_1, \dots, \vv_S$ and propagate the coefficients $\{(\vx_{0,s}= \vx_0, \vx_{1,s} = \vv_s, \vx_{2,s} = \vx_{3,s} = \vx_{4,s} = \vzero)\}_{s=1}^S$.
With standard Taylor mode, this uses ${\color{tab-orange} 1 + 4S}$ vectors; collapsed Taylor mode uses ${\color{tab-green} 1 + 3S + 1}$ vectors.
For the exact biharmonic operator, however, we need to develop an approach to compute mixed partials.

\paragraph{General approach.}
Assume we want to compute a linear differential operator of degree $K$.
We can do so by contracting the $K$-th order derivative tensor $\partial^K \vf(\vx_0)$ with a coefficient tensor $\tC \in (\sR^D)^{\otimes K}$.
We can always express this tensor in a tensor product basis, such that
\begin{align}\label{eq:sums-k-directional}
  \left<
  \partial^K \vf(\vx_0), \tC
  \right>
  =
  \sum_{d_1=1}^{D_1} \dots \sum_{d_I=1}^{D_I}\left<
  \partial^{K} \vf(\vx_0),
  \vv_{d_1}^{\otimes i_1}
  \otimes \ldots \otimes
  \vv_{d_I}^{\otimes i_I}
  \right>
  \,
  \in \sR^C\,,
\end{align}
where the multi-index entries $\vi = (i_1, \dots, i_I)$ sum to $K$ and $D_j \leq D$.
For the exact biharmonic operator (\cref{eq:biharm}), we identify $K = 4, I = 2, \vi = (2, 2), D_1 = D_2 = D, \vv_{d_1} = \ve_{d_1}$, and $\vv_{d_2} = \ve_{d_2}$.
From the Fa\'a di Bruno formula, we know that we can only compute terms of the form $\langle \partial^{K} \vf(\vx_0), \vv^{\otimes K}\rangle$ with a $K$-jet.
The challenge in \cref{eq:sums-k-directional} is that it includes terms where \emph{not} all directions coincide (\eg, for the biharmonic we have $I=2$ different directions).

Fortunately, \citet{griewank_evaluating_1999} derived an approach to reconstruct such mixed-direction terms by linearly combining a \emph{family} of $K$-jets that is determined by all vectors $\vj \in \sN^I$ whose entries sum to $K$, see \cref{fig:ttc_biharm_coeffs} for an illustration for the biharmonic (5 members).
The $K$-jets along these directions are then combined with coefficients $\gamma_{\vi, \vj} \in \sR$, whose definition we provide in \cref{sec:appendix_ttc}.
In summary, we get
\begin{equation}
  \label{eq:ttc_general}
  \!\!\!\!\!
  \left<
    \partial^{K} \vf(\vx_0),
    \vv_{d_1}^{\otimes i_1}
    \otimes \ldots \otimes
    \vv_{d_I}^{\otimes i_I}
  \right>
  = \sum_{\vj \in \sN^I, \lVert \vj \rVert_1 = K}
  \frac{\gamma_{\vi, \vj}}{K!}
  \left<
    \partial^{K}\vf(\vx_0),
    \left(\sum_{i=1}^I \vv_{d_i} [\vj]_i\right)^{\otimes K}
  \right>\,.
\end{equation}
This construction allows us to rewrite \cref{eq:sums-k-directional} as
\begin{align*}
  \sum_{d_1=1}^{D_1} \dots \sum_{d_I=1}^{D_I}
  \sum_{\vj \in \mathbb{N}^I, \lVert \vj \rVert_1 = K}
  \frac{\gamma_{\vi, \vj}}{K!}
  \left<
  \partial^{K}\vf(\vx_0),
  \left(
  \sum_{i = 1}^I \vv_{d_i} [\vj]_i
  \right)^{\otimes K}
  \right>\,,
\end{align*}
and---since the coefficients $\gamma_{\vi,\vj}$ only depend on the problem structure ($K$, $I$ and $\vi$) and \emph{not} on the function $\vf$ and the directions $\vv_{d_i}$ \cite{griewank_evaluating_1999}---we can pull out the inner sum to obtain the final expression
\begin{equation}\label{eq:ttc-general}
  \sum_{\vj \in \mathbb{N}^I, \lVert \vj \rVert_1 = K}
  \frac{\gamma_{\vi, \vj}}{K!}
  {\color{tab-orange}
  \sum_{d_1=1}^{D_1} \dots \sum_{d_I=1}^{D_I}
  }
  \left<
    \partial^{K}\vf(\vx_0),
    \left(
      \sum_{i = 1}^I \vv_{d_i} [\vj]_i
    \right)^{\otimes K}
  \right>\,.
\end{equation}

We can evaluate \cref{eq:ttc-general} with standard Taylor mode: For each $\vj$, compute $\smash{\prod_{i=1}^{I}}D_i$ $K$-jets with coefficients $\vx_0, \vx_1 = \sum_i \vv_{d_i}[\vj]_i, \vx_2= \ldots =\vx_K = \vzero$.
The sums from the tensor basis expansion can be collapsed with our proposed optimization, removing $\smash{\prod_{i=1}^{I}}D_i$ vectors from the propagation for each $\vj$.
After repeating for each member $\vj$ of the interpolation family, we form the linear combination using the $\gamma_{{\vi, \vj}}$s, which yields the desired differential operator.
We can often exploit symmetries in the $\gamma_{\vi, \vj}$s and basis vectors to further reduce the number of $K$-jets (see \cref{sec:appendix-biharmonic-details} for a complete example).

\paragraph{Applied to the biharmonic operator.}
Let us now illustrate the key steps of applying \cref{eq:ttc-general} to the exact biharmonic operator \cref{eq:biharm} (full procedure in \cref{sec:appendix-biharmonic-details}).
\Cref{fig:ttc_biharm_coeffs} illustrates the 5 multi-indices $\vj$ characterizing the $4$-jets we need to interpolate $\langle \partial^4 \vf(\vx_0), \ve_{d_1}^{\otimes 2} \otimes \ve_{d_2}^{\otimes 2} \rangle$, and their coefficients $\gamma_{\vi, \vj}$.
Their definition, see \cref{eq:ttc_coeff}, shows the equality of $\gamma_{\vi, \vj}$ for $\vj = (4,0)$ and $\vj = (0, 4)$, as well as $\vj = (3, 1)$ and $\vj = (1, 3)$.
Exploiting those symmetries reduces the number of interpolation terms from 5 to 3 (\cref{eq:ttc_for_biharm_2}), corresponding to $D + D^2 + D^2$ $4$-jets.
Removing doubly-computed terms brings down the number of $4$-jets to $D + D(D-1) + \nicefrac{1}{2}D(D-1)$ (\cref{eq:ttc_for_biharm_final}).
Translated to vectors, standard Taylor mode propagates $1 + 4D + 4D(D-1) + \nicefrac{4}{2}D(D-1) = {\color{tab-orange} 6D^2 - 2D + 1}$ vectors.
After collapsing, we get $1 + 3D + 1 + 3 D(D-1) + 1 + \nicefrac{3}{2} D(D-1)  + 1 = {\color{tab-green}\nicefrac{9}{2}D^2 - \nicefrac{3}{2} D + 4}$ vectors.
\begin{wrapfigure}[17]{r}{0.415\textwidth}
  \centering
  \vspace*{-0.5ex}
  \begin{tikzpicture}
  \tikzset{transparentfill/.style={fill=white, fill opacity=0.7, text opacity=1.0}}
  \begin{axis}[
    axis lines=middle,
    axis line style={-Stealth, thick},
    xmin=0, xmax=5.25,
    ymin=0, ymax=5.25,
    xtick={0,1,2,3,4},
    ytick={0,1,2,3,4},
    grid=major,
    enlargelimits=false,
    tick style={black},
    tick align=outside,
    tick label style={font=\small},
    width=0.4\textwidth,
    height=0.4\textwidth,
    xlabel style={at={(axis description cs:0.5,-0.15)},anchor=north, font=\small},
    ylabel style={at={(axis description cs:-0.15,0.5)},anchor=south, rotate=90, font=\small},
    clip=false, %
    ]

    \foreach \x in {0,1,2,3,4} {
      \pgfmathsetmacro{\y}{4 - \x}
      \addplot[
      only marks,
      mark=*,
      mark size=2.5pt,
      tab-orange,
      ] coordinates {(\x,\y)};
    }

    \draw[tab-orange, ultra thick, dotted] (axis cs:-0.3, 4.3) to (axis cs:4.3, -0.3);

    \node[anchor=north, yshift=-2pt] at (axis cs:5.25,0) {$[\vj]_1$};
    \node[anchor=east, xshift=-2pt] at (axis cs:0,5.25) {$[\vj]_2$};

    \node[anchor=south west, transparentfill, xshift=2pt, yshift=2pt] at (axis cs:0,4) {$\gamma_{{\vi, (4,0)}} = \nicefrac{13}{192} $};
    \node[anchor=south west, transparentfill, xshift=2pt, yshift=2pt] at (axis cs:1,3) {$\gamma_{{\vi, (1,3)}} = -\nicefrac{1}{3} $};
    \node[anchor=south west, transparentfill, xshift=2pt, yshift=2pt] at (axis cs:2,2) {$\gamma_{{\vi, (2,2)}} = \nicefrac{5}{8} $};
    \node[anchor=south west, transparentfill, xshift=2pt, yshift=2pt] at (axis cs:3,1) {$\gamma_{{\vi, (3,1)}}$};
    \node[anchor=south west, transparentfill, xshift=2pt, yshift=2pt] at (axis cs:4,0) {$\gamma_{{\vi, (4,0)}}$};
  \end{axis}

  \node[anchor=north east, transparentfill] at (current bounding box.north east) {$\vi = (2, 2)$};
\end{tikzpicture}
  \vspace{-1ex}
  \caption{\textbf{Illustration of \cref{eq:ttc-general} for the biharmonic operator}, \ie, the 5 values of $\vj$ with $\lVert \vj \rVert_1 = 4$ and their coefficients $\gamma_{\vi, \vj}$ to interpolate the desired mixed partials.}
  \label{fig:ttc_biharm_coeffs}
\end{wrapfigure}
This demonstrates the relevance of collapsing: it achieves a 25\,\% reduction in the quadratic coefficient.

\paragraph{Summary \& relation to other approaches for computing mixed partials.}
The scheme we propose based on \citet{griewank_evaluating_1999}'s interpolation result allows to calculate \emph{general} linear differential operators beyond Laplacians, and is amenable to collapsing.
Admittedly, \cref{eq:ttc-general} seems daunting at first glance.
However, it (i) offers a one-fits-all recipe to construct schemes for general linear PDE operators, and (ii) does not use jets of order $K'>K$ to compute $K$-th order derivatives.
It is possible to derive more ``pedagogical'' approaches, which however require hand-crafted interpolation rules case by case, and propagation of higher-order jets which is costly (see \cref{sec:appendix_ttc_other_methods} for a pedagogical example using less efficient $6$-jets to compute the biharmonic operator, or \cite[\S{}F]{shi2024stochastic} for other operators).

\section{Implementation \& Experiments}\label{sec:experiments}
\begin{figure*}[!t]
  \small
  \centering
  \newsavebox{\benchmarkLegend}
  \savebox{\benchmarkLegend}{
    \begin{tikzpicture}[font=\small]
      \matrix [%
      matrix of nodes,%
      ampersand replacement=\&,%
      nodes={anchor=west, align=left, inner sep=1pt},%
      column sep=1ex,%
      row sep=0ex,%
      ] (legend)
      {
        \draw[tab-blue] plot[mark=*] coordinates {(0,0)};
        \& Nested 1\textsuperscript{st}-order\phantom{y}\!\!\!\!
        \\
        \draw[tab-orange] plot[mark=triangle*, rotate=270] coordinates {(0,0)};
        \& Standard Taylor
        \\
        \draw[tab-green] plot[mark=triangle*, rotate=90] coordinates {(0,0)};
        \& Collapsed (ours)\phantom{y}\!\!\!\!
        \\[2ex]
        \node[anchor=center]{\tikz\draw[thick] (0, 0) to ++(2.5ex, 0);};
        \& Differentiable\phantom{y}
        \\
        \node[anchor=center, opacity=0.5]{\tikz\draw[thick, dashed] (0, 0) to ++(2.5ex, 0);};
        \& Non-diff.\phantom{y}
        \\
      };

      \draw[gray, rounded corners] (current bounding box.north west) rectangle (current bounding box.south east);
    \end{tikzpicture}
  }
  \newcolumntype{C}{ >{\centering\arraybackslash} m{0.18\textwidth} }
  \newcolumntype{D}{ >{\centering\arraybackslash} m{0.27\textwidth} }
  \newcolumntype{E}{ >{\centering\arraybackslash} m{0.22\textwidth} }
  \begin{tabular}{CDEE}
    & \makecell{\textbf{Laplacian} \\ $(D=50)$}
    & \makecell{\textbf{Weighted Laplacian} \\ $(D=50)$}
    & \makecell{\textbf{Biharmonic} \\ $(D=5)$}
    \\[1ex]
    \makecell{\textbf{Exact} \\ \\ \\ \\ \\ \\ }
    & \includegraphics{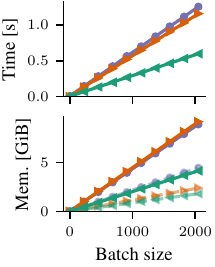}
    & \includegraphics[trim={0.475cm 0 0 0},clip]{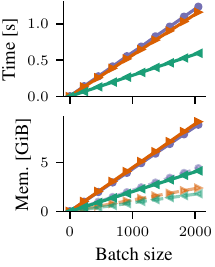}
    & \includegraphics[trim={0.475cm 0 0 0},clip]{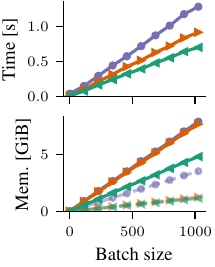}
    \\[-6ex]
  \scalebox{0.85}{
    \begin{tikzpicture}
      \node {\usebox{\benchmarkLegend}};
    \end{tikzpicture}
  }
    & ($N=2048$)
    & ($N=2048$)
    & ($N=256$)
    \\[-6ex]
    \makecell{\\ \textbf{Stochastic}}
    & \includegraphics{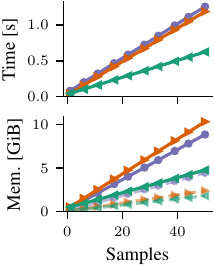}
    & \includegraphics[trim={0.45cm 0 0 0},clip]{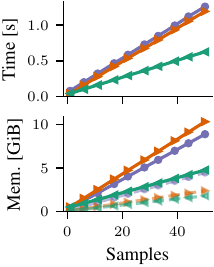}
    & \includegraphics[trim={0.45cm 0 0 0},clip]{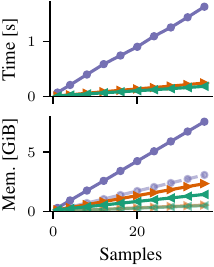}
  \end{tabular}

  \caption{\textbf{\textcolor{tab-green}{Collapsed Taylor mode} accelerates \textcolor{tab-orange}{standard Taylor mode} and outperforms \textcolor{tab-blue}{nested 1\textsuperscript{st}-order AD}.}
    Exact computation varies the batch size, stochastic computation fixes a batch size and varies the samples such that $S < D$ (Laplacians), and $2 + 3S < \nicefrac{9}{2} D^2 - \nicefrac{3}{2}D + 4$ (biharmonic operator); we could compute exactly otherwise.
    Opaque markers are non-differentiable computations.
  }
  \label{fig:benchmark}
\end{figure*}

Here, we describe our implementation of the Taylor mode collapsing process and empirically validate its performance improvements on the previously discussed operators.

\paragraph{Design decisions \& limitations.}
JAX~\cite{bradbury2018jax} already offers an---albeit experimental---Taylor mode implementation~\cite{bettencourt2019taylor}.
However, we found it challenging to capture the computation graph and modify it using JAX's public interface.
In contrast, PyTorch \cite{paszke2019pytorch} provides \texttt{torch.fx} \cite{reed2022torch}, which offers a user-friendly interface to capture and transform computational graphs purely in Python.
Hence, we re-implemented Taylor mode in PyTorch, taking heavy inspiration from the JAX implementation.

This deliberate choice imposes certain limitations.
First, as of now, our Taylor mode in PyTorch supports only a small number of primitives, because the Taylor arithmetic in \cref{eq:faa-di-bruno} needs to be implemented case by case (this of course also applies to JAX's Taylor mode, which has broader operator coverage).
Second, while our Taylor mode implementation is competitive with JAX's, we did not fully optimize it (\eg, we do \emph{not} use in-place operations, and we do \emph{not} implement the efficient schemes from \citet[][\S13]{griewank2008evaluating}, but stick to Fa\`a di Bruno (\cref{eq:faa-di-bruno})).
Given our implementation's superiority compared to nested first-order AD that we demonstrate below, these are promising future efforts that will further improve performance, and we believe that making Taylor mode available to the PyTorch community is also an important step towards establishing its use.

\paragraph{Usage (overview in \cref{sec:appendix-visual-tour}).}
Our implementation takes a PyTorch function (\eg, a neural net) and first captures its computational graph using \texttt{torch.fx}'s symbolic tracing mechanism.
Then, it replaces each operation with its Taylor arithmetic, which yields the computational graph of the function's $K$-jet.
Users can then write a function to compute their differential operator with this vanilla Taylor mode.
Collapsing is achieved using a function \texttt{simplify}, which traces the computation again, rewrites the graph, and propagates the summation of highest coefficients up to its leafs.
This requires one backward traversal through the graph (\cref{sec:graph-simplifications} presents a detailed example).
The simplified graph produces the same result, but propagates summed coefficients, \ie, uses collapsed Taylor mode.

\paragraph{Experimental setup.}
We empirically validate our proposed collapsing approach in PyTorch.
We compare \textcolor{tab-orange}{standard Taylor mode} with \textcolor{tab-green}{collapsed Taylor mode} and \textcolor{tab-blue}{nested 1\textsuperscript{st}-order AD} on an Nvidia RTX 6000 GPU with 24 GiB memory.
To implement the (weighted) Laplacian and its stochastic counterpart, we use vector-Hessian-vector products (VHVPs) in forward-over-reverse order, as recommended \cite{griewank2008evaluating,dagreou2024how}.
For the biharmonic operator, we simply nest two VHVPs.
For the weighted Laplacian's coefficient matrix, we choose a full-rank diagonal matrix (\cref{sec:rank-deficient-weighted-laplacian} shows results for rank-deficient weightings).
To avoid confounding factors, all implementations are executed without compilation (our JAX experiments with the Laplacian in \cref{sec:jax-benchmark} confirm that \texttt{jit} does not affect the relative performance).
As common for PINNs \cite[\eg,][]{shi2024stochastic,dangel2024kroneckerfactored}, we use a 5-layer MLP $\smash{f_\vtheta}: D \to 768 \to 768 \to 512 \to 512 \to 1$ with $\tanh$ activations and trainable parameters $\vtheta$, and compute the PDE operators on batches of size $N$.
We measure three performance metrics:
\textbf{(1) runtime} reports the smallest execution time of 50 repetitions.
\textbf{(2) Peak memory (non-differentiable)} measures the maximum allocated GPU memory when computing the PDE operator's value (\eg, used in VMC \cite{pfau2020ab}) inside a \texttt{torch.no\_grad} context.
\textbf{(3) Peak memory (differentiable)} is the maximum memory usage when computing the PDE operator inside a \texttt{torch.enable\_grad} context, which allows backpropagation to $\vtheta$ (required for training PINNs, or alternative VMC works \cite{webber2022rayleigh, toulouse2007optimization}). This demands saving intermediates, which uses more memory but does not affect runtime.
As memory allocation does not fluctuate much, we measure it in a single run.

\begin{table}[!t]
  \centering
  \caption{\textbf{Benchmark from \cref{fig:benchmark} in numbers.}
    We fit linear functions and report their slopes, \ie, how much runtime and memory increase when incrementing the batch size or random samples.
    We show two significant digits and bold values are best according to parenthesized values.}
  \label{tab:benchmark}
  \vspace{1.5ex}
  \def\datapathLaplacianExact{jet/exp/exp01_benchmark_laplacian/performance/architecture_tanh_mlp_768_768_512_512_1_device_cuda_dim_50_name_laplacian_vary_batch_size}
  \def\datapathLaplacianStochastic{jet/exp/exp01_benchmark_laplacian/performance/architecture_tanh_mlp_768_768_512_512_1_batch_size_2048_device_cuda_dim_50_distribution_normal_name_laplacian_vary_num_samples}
  \def\datapathWeightedLaplacianExact{jet/exp/exp01_benchmark_laplacian/performance/architecture_tanh_mlp_768_768_512_512_1_device_cuda_dim_50_name_weighted_laplacian_vary_batch_size_rank_ratio_1_0}
  \def\datapathWeightedLaplacianStochastic{jet/exp/exp01_benchmark_laplacian/performance/architecture_tanh_mlp_768_768_512_512_1_batch_size_2048_device_cuda_dim_50_distribution_normal_name_weighted_laplacian_vary_num_samples_rank_ratio_1_0}
  \def\datapathBilaplacianExact{jet/exp/exp01_benchmark_laplacian/performance/architecture_tanh_mlp_768_768_512_512_1_device_cuda_dim_5_name_bilaplacian_vary_batch_size}
  \def\datapathBilaplacianStochastic{jet/exp/exp01_benchmark_laplacian/performance/architecture_tanh_mlp_768_768_512_512_1_batch_size_256_device_cuda_dim_5_distribution_normal_name_bilaplacian_vary_num_samples}
  \sisetup{%
    round-mode=figures,%
    round-precision=2,%
    detect-weight, %
    tight-spacing=true, %
  }
  \begin{small}
    \begin{tabular}{ccc|cccc}
      \toprule
      \textbf{Mode}
      & \makecell{\textbf{Per-datum or } \\ \textbf{-sample cost}}
      & \textbf{Implementation}
      & \textbf{Laplacian}
      & \makecell{\textbf{Weighted} \\ \textbf{Laplacian}}
      & \textbf{Biharmonic}
      \\
      \midrule
      \multirow{9}{*}{\textbf{Exact}}
      & \multirow{3}{*}{Time [ms]}
      & \textcolor{tab-blue}{Nested 1\textsuperscript{st}-order}
      & \num{0.5884870457016232} (\num{1.0}x)
      & \num{0.5805526928424408} (\num{1.0}x)
      & \num{1.2431203687661039} (\num{1.0}x)
      \\
      &
      & \textcolor{tab-orange}{Standard Taylor}
      & \num{0.5486171149456046} (\num{0.9322501131550249}x)
      & \num{0.5506234432925975} (\num{0.9484469714483506}x)
      & \num{0.8982907495333762} (\num{0.7226096298502465}x)
      \\
      &
      & \textcolor{tab-green}{Collapsed (ours)}
      & \textbf{\num{0.28438100215488904} (\num{0.4832408873432991}x)}
      & \textbf{\num{0.2853385981869548} (\num{0.4914947457911358}x)}
      & \textbf{\num{0.6881202414621329} (\num{0.5535427290481509}x)}
      \\ \cmidrule{2-6}
      & \multirow{3}{*}{\makecell{Mem.\,[MiB] \\ (differentiable)}}
      & \textcolor{tab-blue}{Nested 1\textsuperscript{st}-order}
      & \num{3.5749680525604055} (\num{1.0}x)
      & \num{3.403211467109172} (\num{1.0}x)
      & \num{7.855936095166244} (\num{1.0}x)
      \\
x      &
      & \textcolor{tab-orange}{Standard Taylor}
      & \num{3.147110397590361} (\num{0.880318467555644}x)
      & \num{3.147110397590361} (\num{0.9247472359581717}x)
      & \num{7.681819604881255} (\num{0.9778363153447592}x)
      \\
      &
      & \textcolor{tab-green}{Collapsed (ours)}
      & \textbf{\num{1.5325974347024895} (\num{0.4287024141669847}x)}
      & \textbf{\num{1.4711803076429506} (\num{0.4322917696597419}x)}
      & \textbf{\num{4.820656216648105} (\num{0.6136323104275574}x)}
      \\ \cmidrule{2-6}
      & \multirow{3}{*}{\makecell{Mem.\,[MiB] \\ (non-diff.)}}
      & \textcolor{tab-blue}{Nested 1\textsuperscript{st}-order}
      & \num{1.2195471026153386} (\num{1.0}x)
      & \num{1.141825801600289} (\num{1.0}x)
      & \num{3.5169997755341087} (\num{1.0}x)
      \\
      &
      & \textcolor{tab-orange}{Standard Taylor}
      & \num{0.8458782974738871} (\num{0.6936003502118838}x)
      & \num{0.8390930911717096} (\num{0.734869618461683}x)
      & \num{1.2494003296969105} (\num{0.3552460646680509}x)
      \\
      &
      & \textcolor{tab-green}{Collapsed (ours)}
      & \textbf{\num{0.6001617975249248} (\num{0.4921185874968405}x)}
      & \textbf{\num{0.6001617975249248} (\num{0.5256159010277991}x)}
      & \textbf{\num{1.144122743620398} (\num{0.3253121457611256}x)}
      \\
      \midrule
      \multirow{9}{*}{\textbf{Stochastic}}
      & \multirow{3}{*}{Time [ms]}
      & \textcolor{tab-blue}{Nested 1\textsuperscript{st}-order}
      & \num{23.10768183294421} (\num{1.0}x)
      & \num{23.21994389563674} (\num{1.0}x)
      & \num{43.81230709599838} (\num{1.0}x)
      \\
      &
      & \textcolor{tab-orange}{Standard Taylor}
      & \num{22.649913821633568} (\num{0.9801897907968419}x)
      & \num{22.687909113919662} (\num{0.9770871633407757}x)
      & \num{6.633426018548653} (\num{0.15140554009205603}x)
      \\
      &
      & \textcolor{tab-green}{Collapsed (ours)}
      & \textbf{\num{11.654805055586671} (\num{0.5043692889595971}x)}
      & \textbf{\num{11.720202787318662} (\num{0.5047472483135933}x)}
      & \textbf{\num{4.93249574117899} (\num{0.11258242416614261}x)}
      \\ \cmidrule{2-6}
      & \multirow{3}{*}{\makecell{Mem.\,[MiB]\\(differentiable)}}
      & \textcolor{tab-blue}{Nested 1\textsuperscript{st}-order}
      & \num{148.90348678329067} (\num{1.0}x)
      & \num{148.90348678329067} (\num{1.0}x)
      & \num{212.26228637837823} (\num{1.0}x)
      \\
      &
      & \textcolor{tab-orange}{Standard Taylor}
      & \num{140.8007539281551} (\num{0.945583995175828}x)
      & \num{140.8007539281551} (\num{0.945583995175828}x)
      & \num{64.48353862502677} (\num{0.3037917838597035}x)
      \\
      &
      & \textcolor{tab-green}{Collapsed (ours)}
      & \textbf{\num{73.18030804517024} (\num{0.49146134604406205}x)}
      & \textbf{\num{73.18030804517024} (\num{0.49146134604406205}x)}
      & \textbf{\num{37.94950079711943} (\num{0.17878588535257148}x)}
      \\ \cmidrule{2-6}
      & \multirow{3}{*}{\makecell{Mem.\,[MiB]\\(non-diff.)}}
      & \textcolor{tab-blue}{Nested 1\textsuperscript{st}-order}
      & \num{50.28844883985535} (\num{1.0}x)
      & \num{50.28844883985535} (\num{1.0}x)
      & \num{86.44269130866216} (\num{1.0}x)
      \\
      &
      & \textcolor{tab-orange}{Standard Taylor}
      & \num{33.453246966144256} (\num{0.6652272587026266}x)
      & \num{33.45324696614425} (\num{0.6652272587026264}x)
      & \num{14.953454358984649} (\num{0.17298691344060696}x)
      \\
      &
      & \textcolor{tab-green}{Collapsed (ours)}
      & \textbf{\num{16.780587668478763} (\num{0.33368672241048647}x)}
      & \textbf{\num{16.780587668478763} (\num{0.33368672241048647}x)}
      & \textbf{\num{14.94800141000759} (\num{0.17292383177465573}x)}
      \\
      \bottomrule
    \end{tabular}
  \end{small}
\end{table}

\paragraph{Results.}
\Cref{fig:benchmark} visualizes the growth in computational resources \wrt the batch size (exact) and random samples (stochastic) for fixed dimensions $D$.
Runtime and memory increase linearly in both, as expected.
We quantify the results by fitting linear functions and reporting their slopes (\ie, time and memory added per datum/sample) in \cref{tab:benchmark}.
We make the following observations:
\begin{itemize}[leftmargin=0.5cm]
\item \textbf{Collapsed Taylor mode accelerates standard Taylor mode.}
  The measured performance differences correspond well with the theoretical estimate from counting the number of forward-propagated vectors.
  \Eg, for the exact Laplacian, adding one datum introduces {\color{tab-green}$2 + D$} versus {\color{tab-orange}$1 + 2D$} new vectors.
  For $D=50$, their ratio is $\nicefrac{\color{tab-green}(2 + D)}{\color{tab-orange}(1 + 2D)} \approx 0.51$.
Empirically, we measure that adding one datum adds {\color{tab-orange}\inputMetricOnly{jet/exp/exp01_benchmark_laplacian/performance/architecture_tanh_mlp_768_768_512_512_1_device_cuda_dim_50_name_laplacian_vary_batch_size/jet_naive_best.txt}\,ms} to standard, and {\color{tab-green}\inputMetricOnly{jet/exp/exp01_benchmark_laplacian/performance/architecture_tanh_mlp_768_768_512_512_1_device_cuda_dim_50_name_laplacian_vary_batch_size/jet_simplified_best.txt}\,ms} to collapsed, Taylor mode (\cref{tab:benchmark}); the ratio of $\approx \!\!\!\inputMetricRatio{jet/exp/exp01_benchmark_laplacian/performance/architecture_tanh_mlp_768_768_512_512_1_device_cuda_dim_50_name_laplacian_vary_batch_size/jet_simplified_best.txt}{jet/exp/exp01_benchmark_laplacian/performance/architecture_tanh_mlp_768_768_512_512_1_device_cuda_dim_50_name_laplacian_vary_batch_size/jet_naive_best.txt}$ is close.
  Similar arguments hold for peak memory of differentiable computation, stochastic approximation, and the other PDE operators (see \cref{tab:benchmark-ratios} for all comparisons).

\item \textbf{Collapsed Taylor mode outperforms nested 1\textsuperscript{st}-order AD.}
  For the exact and stochastic (weighted) Laplacians, collapsed Taylor mode is roughly twice as fast (consistent with the JAX results in \cref{fig:vanilla-taylor-not-enough}) while using only 40-50\% memory.
  For the biharmonic operator, we also observe speed-ups;
  in the stochastic case up to 9x in time, and 5x in memory (differentiable).
\end{itemize}

\paragraph{Comparison with JAX.} We also conducted experiments with JAX (+ \texttt{jit}) to rule out artifacts from choosing PyTorch, implementation mistakes in our Taylor mode library, or unexpected simplifications from the JIT compiler.
We find that the choice of the ML framework does not affect the results.
\Eg, when computing the exact Laplacian with nested first-order AD, PyTorch consumes \inputMetricOnly{jet/exp/exp01_benchmark_laplacian/performance/architecture_tanh_mlp_768_768_512_512_1_device_cuda_dim_50_name_laplacian_vary_batch_size/hessian_trace_best.txt}\,ms per datum (\cref{tab:benchmark}), while JAX uses \inputMetricOnly{jet/exp/exp04_jax_benchmark/performance/architecture_tanh_mlp_768_768_512_512_1_device_cuda_dim_50_name_jax_laplacian_vary_batch_size/hessian_trace_best.txt}\,ms (\cref{fig:vanilla-taylor-not-enough,tab:jax-benchmark}).
We find the same trend when comparing our collapsed Taylor mode and JAX's forward Laplacian.
Interestingly, we noticed that JAX's Taylor mode was consistently slower than our PyTorch implementation, despite using \texttt{jit}.
We hypothesize that this could stem from algorithmic differences in the Taylor mode implementations and conclude from these results that (both ours, as well as the existing JAX) Taylor mode still has potential for improvements that may further increase the margin to nested first-order.

\section{Conclusion}\label{sec:conclusion}
Computing differential operators is a critical component in scientific machine learning, particularly for Physics-informed neural networks and variational Monte-Carlo.
Our work introduces collapsed Taylor mode, a simple yet effective optimization based on linearity in Fa\`a di Bruno's formula, that propagates the sum of highest-order Taylor coefficients, rather than propagating then summing.
It contains recent advances in forward-mode schemes, recovering the forward Laplacian \cite{li2023forward}, while being applicable to stochastic Taylor mode \cite{shi2024stochastic,hu2024hutchinson}.
We demonstrated that collapsed Taylor mode is useful to compute general linear differential operators, leveraging \citet{griewank_evaluating_1999}'s interpolation formula.
Empirically, we confirmed speed-ups and memory savings for computing (randomized) Laplacians and biharmonic operators after collapsing Taylor mode, in accordance with our theoretical analysis, and confirmed its superiority to nesting first-order automatic differentiation.
As the optimizations are achieved through simple graph rewrites based on linearity, we believe they could be integrated into existing just-in-time compilers without requiring a new interface or burdening users.

Our work takes an important step towards making Taylor mode a practical alternative to nested first-order differentiation in scientific machine learning, while maintaining ease of use.
Future work could focus on integrating these optimizations directly into ML compilers, broadening operator coverage of our PyTorch implementation, and exploring additional graph optimizations for AD.

\begin{ack}
  Resources used in preparing this research were provided, in part, by the Province of Ontario, the Government of Canada through CIFAR, and companies sponsoring the Vector Institute.
  The research was funded partly by the DFG under Germany’s Excellence Strategy -- The Berlin Mathematics Research Center MATH+ (EXC-2046/1, project ID:390685689). M.Z. acknowledges support from an ETH Postdoctoral Fellowship for the project ``Reliable, Efficient, and Scalable Methods for Scientific
Machine Learning''.
\end{ack}

\bibliography{references}

\begin{thebibliography}{32}
\providecommand{\natexlab}[1]{#1}
\providecommand{\url}[1]{\texttt{#1}}
\expandafter\ifx\csname urlstyle\endcsname\relax
  \providecommand{\doi}[1]{doi: #1}\else
  \providecommand{\doi}{doi: \begingroup \urlstyle{rm}\Url}\fi

\bibitem[Arbogast(1800)]{arbogast1800calcul}
Arbogast, L.
\newblock \emph{Du calcul des d{\'e}rivations}.
\newblock 1800.

\bibitem[Bettencourt et~al.(2019)Bettencourt, Johnson, and
  Duvenaud]{bettencourt2019taylor}
Bettencourt, J., Johnson, M.~J., and Duvenaud, D.
\newblock Taylor-mode automatic differentiation for higher-order derivatives in
  {JAX}.
\newblock In \emph{Advances in Neural Information Processing Systems (NeurIPS);
  Workhop on Program Transformations for ML}, 2019.

\bibitem[Bradbury et~al.(2018)Bradbury, Frostig, Hawkins, Johnson, Leary,
  Maclaurin, and Wanderman-Milne]{bradbury2018jax}
Bradbury, J., Frostig, R., Hawkins, P., Johnson, M.~J., Leary, C., Maclaurin,
  D., and Wanderman-Milne, S.
\newblock {JAX}: composable transformations of {P}ython+{N}um{P}y programs,
  2018.

\bibitem[Carleo \& Troyer(2017)Carleo and Troyer]{carleo2017solving}
Carleo, G. and Troyer, M.
\newblock Solving the quantum many-body problem with artificial neural
  networks.
\newblock \emph{Science}, 355\penalty0 (6325):\penalty0 602--606, 2017.

\bibitem[Dagréou et~al.(2024)Dagréou, Ablin, Vaiter, and
  Moreau]{dagreou2024how}
Dagréou, M., Ablin, P., Vaiter, S., and Moreau, T.
\newblock How to compute {H}essian-vector products?
\newblock In \emph{International Conference on Learning Representations (ICLR)
  Blogposts}, 2024.

\bibitem[Dangel et~al.(2024)Dangel, Müller, and
  Zeinhofer]{dangel2024kroneckerfactored}
Dangel, F., Müller, J., and Zeinhofer, M.
\newblock Kronecker-factored approximate curvature for physics-informed neural
  networks.
\newblock In \emph{Advances in Neural Information Processing Systems
  (NeurIPS)}, 2024.

\bibitem[Dwivedi \& Srinivasan(2020)Dwivedi and Srinivasan]{vikas_biharm}
Dwivedi, V. and Srinivasan, B.
\newblock Solution of biharmonic equation in complicated geometries with
  physics informed extreme learning machine.
\newblock \emph{Journal of Computing and Information Science in Engineering},
  20\penalty0 (6), 05 2020.
\newblock ISSN 1530-9827.

\bibitem[Fa(2011)]{fa2011solution}
Fa, K.~S.
\newblock Solution of {F}okker-{P}lanck equation for a broad class of drift and
  diffusion coefficients.
\newblock \emph{Phys. Rev. E}, 2011.

\bibitem[Fa{\`a} Di~Bruno(1857)]{faa1857note}
Fa{\`a} Di~Bruno, F.
\newblock Note sur une nouvelle formule de calcul diff{\'e}rentiel.
\newblock \emph{Quarterly J. Pure Appl. Math}, 1857.

\bibitem[Foulkes et~al.(2001)Foulkes, Mitas, Needs, and
  Rajagopal]{foulkes2001quantum}
Foulkes, W.~M., Mitas, L., Needs, R., and Rajagopal, G.
\newblock Quantum {M}onte {C}arlo simulations of solids.
\newblock \emph{Reviews of Modern Physics}, 73\penalty0 (1):\penalty0 33, 2001.

\bibitem[Fraenkel(1978)]{fraenkel1978formulae}
Fraenkel, L.
\newblock Formulae for high derivatives of composite functions.
\newblock In \emph{Mathematical Proceedings of the Cambridge Philosophical
  Society}, 1978.

\bibitem[Gao et~al.(2023)Gao, Köhler, and Foster]{gao2023folx}
Gao, N., Köhler, J., and Foster, A.
\newblock folx - forward {L}aplacian for {JAX}, 2023.
\newblock URL \url{http://github.com/microsoft/folx}.

\bibitem[Griewank \& Walther(2008)Griewank and Walther]{griewank2008evaluating}
Griewank, A. and Walther, A.
\newblock \emph{Evaluating derivatives: principles and techniques of
  algorithmic differentiation}.
\newblock SIAM, 2008.

\bibitem[Griewank et~al.(1999)Griewank, Utke, and
  Walther]{griewank_evaluating_1999}
Griewank, A., Utke, J., and Walther, A.
\newblock Evaluating higher derivative tensors by forward propagation of
  univariate {Taylor} series.
\newblock \emph{Mathematics of Computation}, 69, 1999.

\bibitem[Hardy(2006)]{hardy2006combinatorics}
Hardy, M.
\newblock Combinatorics of partial derivatives, 2006.
\newblock arXiv.

\bibitem[Hermann et~al.(2020)Hermann, Sch{\"a}tzle, and
  No{\'e}]{hermann2020deep}
Hermann, J., Sch{\"a}tzle, Z., and No{\'e}, F.
\newblock Deep-neural-network solution of the electronic {S}chr{\"o}dinger
  equation.
\newblock \emph{Nature Chemistry}, 12\penalty0 (10):\penalty0 891--897, 2020.

\bibitem[Hu et~al.(2024)Hu, Shi, Karniadakis, and Kawaguchi]{hu2024hutchinson}
Hu, Z., Shi, Z., Karniadakis, G.~E., and Kawaguchi, K.
\newblock Hutchinson trace estimation for high-dimensional and high-order
  physics-informed neural networks.
\newblock \emph{Computer Methods in Applied Mechanics and Engineering}, 2024.

\bibitem[Hutchinson(1989)]{hutchinson1989stochastic}
Hutchinson, M.
\newblock A stochastic estimator of the trace of the influence matrix for
  laplacian smoothing splines.
\newblock \emph{Communication in Statistics---Simulation and Computation},
  1989.

\bibitem[Karniadakis et~al.(2021)Karniadakis, Kevrekidis, Lu, Perdikaris, Wang,
  and Yang]{karniadakis2021physics}
Karniadakis, G.~E., Kevrekidis, I.~G., Lu, L., Perdikaris, P., Wang, S., and
  Yang, L.
\newblock Physics-informed machine learning.
\newblock \emph{Nature Reviews Physics}, 3\penalty0 (6):\penalty0 422--440,
  2021.

\bibitem[Li et~al.(2024{\natexlab{a}})Li, Wang, Ye, He, and Wang]{li2024dof}
Li, R., Wang, C., Ye, H., He, D., and Wang, L.
\newblock {DOF}: Accelerating high-order differential operators with forward
  propagation.
\newblock In \emph{International Conference on Learning Representations (ICLR),
  Workshop on AI4DifferentialEquations In Science}, 2024{\natexlab{a}}.

\bibitem[Li et~al.(2024{\natexlab{b}})Li, Ye, Jiang, Wen, Wang, Li, Li, He,
  Chen, Ren, et~al.]{li2023forward}
Li, R., Ye, H., Jiang, D., Wen, X., Wang, C., Li, Z., Li, X., He, D., Chen, J.,
  Ren, W., et~al.
\newblock A computational framework for neural network-based variational
  {M}onte {C}arlo with forward {L}aplacian.
\newblock \emph{Nature Machine Intelligence}, 2024{\natexlab{b}}.

\bibitem[Paszke et~al.(2019)Paszke, Gross, Massa, Lerer, Bradbury, Chanan,
  Killeen, Lin, Gimelshein, Antiga, Desmaison, Kopf, Yang, DeVito, Raison,
  Tejani, Chilamkurthy, Steiner, Fang, Bai, and Chintala]{paszke2019pytorch}
Paszke, A., Gross, S., Massa, F., Lerer, A., Bradbury, J., Chanan, G., Killeen,
  T., Lin, Z., Gimelshein, N., Antiga, L., Desmaison, A., Kopf, A., Yang, E.,
  DeVito, Z., Raison, M., Tejani, A., Chilamkurthy, S., Steiner, B., Fang, L.,
  Bai, J., and Chintala, S.
\newblock {PyTorch}: An imperative style, high-performance deep learning
  library.
\newblock In \emph{Advances in Neural Information Processing Systems
  (NeurIPS)}. 2019.

\bibitem[Pearlmutter(1994)]{pearlmutter1994fast}
Pearlmutter, B.~A.
\newblock Fast exact multiplication by the {H}essian.
\newblock \emph{Neural Computation}, 1994.

\bibitem[Pfau et~al.(2020)Pfau, Spencer, Matthews, and Foulkes]{pfau2020ab}
Pfau, D., Spencer, J.~S., Matthews, A.~G., and Foulkes, W. M.~C.
\newblock Ab initio solution of the many-electron {S}chr{\"o}dinger equation
  with deep neural networks.
\newblock \emph{Physical Review Research}, 2020.

\bibitem[Raissi et~al.(2019)Raissi, Perdikaris, and
  Karniadakis]{raissi2019physics}
Raissi, M., Perdikaris, P., and Karniadakis, G.~E.
\newblock Physics-informed neural networks: A deep learning framework for
  solving forward and inverse problems involving nonlinear partial differential
  equations.
\newblock \emph{Journal of Computational physics}, 378:\penalty0 686--707,
  2019.

\bibitem[Reed et~al.(2022)Reed, DeVito, He, Ussery, and Ansel]{reed2022torch}
Reed, J., DeVito, Z., He, H., Ussery, A., and Ansel, J.
\newblock torch.fx: Practical program capture and transformation for deep
  learning in python.
\newblock \emph{Proceedings of Machine Learning and Systems (MLSys)}, 2022.

\bibitem[Shi et~al.(2024)Shi, Hu, Lin, and Kawaguchi]{shi2024stochastic}
Shi, Z., Hu, Z., Lin, M., and Kawaguchi, K.
\newblock Stochastic {T}aylor derivative estimator: Efficient amortization for
  arbitrary differential operators.
\newblock In \emph{Advances in Neural Information Processing Systems
  (NeurIPS)}, 2024.

\bibitem[Sun et~al.(2024)Sun, Berner, Richter, Zeinhofer, M{\"u}ller,
  Azizzadenesheli, and Anandkumar]{sun2024dynamical}
Sun, J., Berner, J., Richter, L., Zeinhofer, M., M{\"u}ller, J.,
  Azizzadenesheli, K., and Anandkumar, A.
\newblock Dynamical measure transport and neural {PDE} solvers for sampling.
\newblock \emph{arXiv preprint arXiv:2407.07873}, 2024.

\bibitem[Sun et~al.(2020)Sun, Li, Liang, Ding, and Srikant]{sun2020global}
Sun, R., Li, D., Liang, S., Ding, T., and Srikant, R.
\newblock The global landscape of neural networks: An overview, 2020.

\bibitem[Toulouse \& Umrigar(2007)Toulouse and
  Umrigar]{toulouse2007optimization}
Toulouse, J. and Umrigar, C.~J.
\newblock Optimization of quantum {M}onte {C}arlo wave functions by energy
  minimization.
\newblock \emph{The Journal of Chemical Physics (JCP)}, 2007.

\bibitem[Vahab et~al.(2022)Vahab, Haghighat, Khaleghi, and
  Khalili]{vahab_physics-informed_2022}
Vahab, M., Haghighat, E., Khaleghi, M., and Khalili, N.
\newblock A physics-informed neural network approach to solution and
  identification of biharmonic equations of elasticity.
\newblock \emph{Journal of Engineering Mechanics}, 148\penalty0 (2), 2022.

\bibitem[Webber \& Lindsey(2022)Webber and Lindsey]{webber2022rayleigh}
Webber, R.~J. and Lindsey, M.
\newblock Rayleigh-{G}auss-{N}ewton optimization with enhanced sampling for
  variational {M}onte {C}arlo.
\newblock \emph{Physical Review Research}, 2022.

\end{thebibliography}
\bibliographystyle{icml2024.bst}

\clearpage

\section*{NeurIPS Paper Checklist}

\begin{enumerate}

\item {\bf Claims}
    \item[] Question: Do the main claims made in the abstract and introduction accurately reflect the paper's contributions and scope?
    \item[] Answer: \answerYes{} %
    \item[] Justification:
    {We propose an acceleration technique for Taylor mode automatic differentiation (AD), describe the scope of its applicability and how it works, then empirically verify the claimed acceleration capabilities. This is clearly communicated in the abstract.}

\item {\bf Limitations}
    \item[] Question: Does the paper discuss the limitations of the work performed by the authors?
    \item[] Answer: \answerYes{} %
    \item[] Justification:
    {We specifically incorporate a paragraph about limitations of the implementation in \cref{sec:experiments} and state the assumptions (linear differential operator) under which the proposed optimization technique is applicable.}

\item {\bf Theory assumptions and proofs}
    \item[] Question: For each theoretical result, does the paper provide the full set of assumptions and a complete (and correct) proof?
    \item[] Answer: \answerNA{} %
    \item[] Justification:
    {The paper does not present new theoretical results.
    All theoretical results used in this work (specifically, \cite{griewank_evaluating_1999}) are cited, and we include self-contained introductions that are relevant to our approach in the appendix (specifically, \cref{sec:appendix_ttc}).
    }

    \item {\bf Experimental result reproducibility}
    \item[] Question: Does the paper fully disclose all the information needed to reproduce the main experimental results of the paper to the extent that it affects the main claims and/or conclusions of the paper (regardless of whether the code and data are provided or not)?
    \item[] Answer: \answerYes{} %
    \item[] Justification:
    {
    We describe the experimental procedure, including hardware and hyper-parameter details, in \cref{sec:experiments}, together with a more detailed description in \cref{sec:jax-benchmark,sec:pytorch-benchmark}, where we also partially reproduce our results using a different ML library.
    We will open-source the code used to produce our results to further facilitate their reproducibility.
    }

\item {\bf Open access to data and code}
    \item[] Question: Does the paper provide open access to the data and code, with sufficient instructions to faithfully reproduce the main experimental results, as described in supplemental material?
    \item[] Answer: \answerYes{} %
    \item[] Justification:
    {We will open-source the code for our PyTorch Taylor mode library (which is not the main contribution of our paper), as well as the code used to generate our results, including instructions how to reproduce them.
      See the link in the main text.}

\item {\bf Experimental setting/details}
    \item[] Question: Does the paper specify all the training and test details (e.g., data splits, hyperparameters, how they were chosen, type of optimizer, etc.) necessary to understand the results?
    \item[] Answer: \answerYes{} %
    \item[] Justification:
    {Our experiments consist of performance measurements.
    We clearly specify the protocol, used hardware, architectural details, and other hyper-parameters in \cref{sec:experiments}, with more details in \cref{sec:jax-benchmark,sec:pytorch-benchmark}.
    }

\item {\bf Experiment statistical significance}
    \item[] Question: Does the paper report error bars suitably and correctly defined or other appropriate information about the statistical significance of the experiments?
    \item[] Answer: \answerYes{} %
    \item[] Justification:
    {
    While we do not report error bars, we choose the experimental procedure to minimize noise: To measure runtime, we report the smallest number from repeating a run 50 times.
    Reporting mean and standard deviations is less conclusive because runtimes on hardware can vary due to interference from other processes or warm-up effects during the first execution.
    Reporting the minimum time as best approximation to the actual runtime
    is recommended, \eg in \url{https://realpython.com/python-profiling/}.
    }

\item {\bf Experiments compute resources}
    \item[] Question: For each experiment, does the paper provide sufficient information on the computer resources (type of compute workers, memory, time of execution) needed to reproduce the experiments?
    \item[] Answer: \answerYes{} %
    \item[] Justification:
    {We mention the exact type of GPU used in \cref{sec:experiments}.
    Since we evaluate runtime and memory performances, we used a single GPU.
    The total compute time can be estimated by multiplying our reported timings with the number of repetitions, then summing over all measurement points.
    It totals less than 24 GPU hours.
    }

\item {\bf Code of ethics}
    \item[] Question: Does the research conducted in the paper conform, in every respect, with the NeurIPS Code of Ethics \url{https://neurips.cc/public/EthicsGuidelines}?
    \item[] Answer: \answerYes{} %
    \item[] Justification:
    {
    After carefully reading the Code of Ethics, we believe our research conforms with it.
    }

\item {\bf Broader impacts}
    \item[] Question: Does the paper discuss both potential positive societal impacts and negative societal impacts of the work performed?
    \item[] Answer: \answerNA{} %
    \item[] Justification:
    {
    Our work introduces algorithmic improvements to Taylor mode automatic differentiation.
    We think they positively impact the research landscape of scientific machine learning, by reducing computational resources.
    We do not foresee any direct negative societal impacts of our work.
    }

\item {\bf Safeguards}
    \item[] Question: Does the paper describe safeguards that have been put in place for responsible release of data or models that have a high risk for misuse (e.g., pretrained language models, image generators, or scraped datasets)?
    \item[] Answer: \answerNA{} %
    \item[] Justification:
    {
    Our work neither uses data sets, nor trains models.
    }

\item {\bf Licenses for existing assets}
    \item[] Question: Are the creators or original owners of assets (e.g., code, data, models), used in the paper, properly credited and are the license and terms of use explicitly mentioned and properly respected?
    \item[] Answer: \answerYes{} %
    \item[] Justification:
    {
    We give credit to code packages we use through citations, specifically JAX's Taylor mode~\cite{bettencourt2019taylor} and forward Laplacian~\cite{gao2023folx}, as well as \texttt{torch.fx}~\cite{reed2022torch}.
    }

\item {\bf New assets}
    \item[] Question: Are new assets introduced in the paper well documented and is the documentation provided alongside the assets?
    \item[] Answer: \answerYes{} %
    \item[] Justification:
    {
    Our PyTorch implementation of Taylor mode is fully-documented and tested.
    We are planning to open-source it soon.
    }

\item {\bf Crowdsourcing and research with human subjects}
    \item[] Question: For crowdsourcing experiments and research with human subjects, does the paper include the full text of instructions given to participants and screenshots, if applicable, as well as details about compensation (if any)?
    \item[] Answer: \answerNA{} %
    \item[] Justification:
    {
    The paper does not involve crowdsourcing nor research with human subjects.
    }

\item {\bf Institutional review board (IRB) approvals or equivalent for research with human subjects}
    \item[] Question: Does the paper describe potential risks incurred by study participants, whether such risks were disclosed to the subjects, and whether Institutional Review Board (IRB) approvals (or an equivalent approval/review based on the requirements of your country or institution) were obtained?
    \item[] Answer: \answerNA{} %
    \item[] Justification:
    {
    The paper does not involve crowdsourcing nor research with human subjects.
    }

\item {\bf Declaration of LLM usage}
    \item[] Question: Does the paper describe the usage of LLMs if it is an important, original, or non-standard component of the core methods in this research? Note that if the LLM is used only for writing, editing, or formatting purposes and does not impact the core methodology, scientific rigorousness, or originality of the research, declaration is not required.
    \item[] Answer: \answerNA{} %
    \item[] Justification:
    {
    Our work does not rely on LLM usage in any way.
    }

\end{enumerate}

\clearpage
\appendix

\renewcommand\theequation{\thesection\arabic{equation}}
\renewcommand\thefigure{\thesection\arabic{figure}}
\renewcommand\thetable{\thesection\arabic{table}}

\makeatletter
\vbox{%
  \hsize\textwidth
  \linewidth\hsize
  \vskip 0.1in
  \@toptitlebar
  \centering
  {\LARGE\bf \@title (Supplementary Material)\par}
  \@bottomtitlebar
  \vskip 0.3in \@minus 0.1in
}
\makeatother

\startcontents[sections]
\printcontents[sections]{l}{1}{\setcounter{tocdepth}{2}}

\clearpage
\section{Fa\`a Di Bruno Formula Cheat Sheet}\label{sec:faa-di-bruno-cheatsheet}
To give some intuition on the Fa\`a di Bruno formula, we illustrate \cref{eq:taylor-mode-composition} for higher orders here:
\vspace{2ex}
\begin{tiny}
  \begin{align*}
    \vx_{0}
    &\to&
          \vh_{0}
          =
          \vh(\vx_{0})
    &\to&
          \vg_{0}
          =
          \vg(\vh_{0})
          =
    &\vf_{0}
      =
      \vf(\vx_{0})
    \\
    \vx_{1}
    &\to&
          \vh_{1}
          =
          \langle \partial \vh, \vx_{1} \rangle
    &\to&
          \vg_{1}
          =
          \langle \partial \vg, \vh_{1} \rangle
          =
    &\vf_{1}
      =
      \langle \partial \vf, \vx_{1} \rangle
    \\
    \vx_{2}
    &\to&
          \vh_{2}
          =
          \begin{matrix}
            \langle \partial^{2} \vh, \vx_{1}^{\otimes2} \rangle
            \\
            +
            \langle \partial \vh, \vx_{2} \rangle
          \end{matrix}
    &\to&
          \vg_{2}
          =
          \begin{matrix}
            \langle \partial^{2} \vg, \vh_{1}^{\otimes2} \rangle
            \\
            +
            \langle \partial \vg, \vh_{2} \rangle
          \end{matrix}
          =
    &\vf_{2}
      =
      \begin{matrix}
        \langle \partial^{2} \vf, \vx_{1}^{\otimes2} \rangle
        \\
        +
        \langle \partial \vf, \vx_{2} \rangle
      \end{matrix}
    \\
    \vx_{3}
    &\to&
          \vh_{3}
          =
          \begin{matrix}
            \langle \partial^{3} \vh, \vx_{1}^{\otimes3} \rangle
            \\
            +
            3 \langle \partial^{2} \vh, \vx_{1}\otimes \vx_{2} \rangle
            \\
            +
            \langle \partial \vh, \vx_{3} \rangle
          \end{matrix}
    &\to&
          \vg_{3}
          =
          \begin{matrix}
            \langle \partial^{3} \vg, \vh_{1}^{\otimes3} \rangle
            \\
            +
            3 \langle \partial^{2} \vg, \vh_{1}\otimes \vh_{2} \rangle
            \\
            +
            \langle \partial \vg, \vh_{3} \rangle
          \end{matrix}
          =
    &\vf_{3}
      =
      \begin{matrix}
        \langle \partial^{3} \vf, \vx_{1}^{\otimes3} \rangle
        \\
        +
        3 \langle \partial^{2} \vf, \vx_{1}\otimes \vx_{2} \rangle
        \\
        +
        \langle \partial \vf, \vx_{3} \rangle
      \end{matrix}
    \\
    \vx_{4}
    &\to&
          \vh_{4}
          =
          \begin{matrix}
            \langle \partial^{4} \vh, \vx_{1}^{\otimes4} \rangle
            \\
            +
            6 \langle \partial^{3} \vh, \vx_{1}^{\otimes2}\otimes \vx_{2} \rangle
            \\
            +
            4 \langle \partial^{2} \vh, \vx_{1}\otimes \vx_{3} \rangle
            \\
            +
            3 \langle \partial^{2} \vh, \vx_{2}^{\otimes2} \rangle
            \\
            +
            \langle \partial \vh, \vx_{4} \rangle
          \end{matrix}
    &\to&
          \vg_{4}
          =
          \begin{matrix}
            \langle \partial^{4} \vg, \vh_{1}^{\otimes4} \rangle
            \\
            +
            6 \langle \partial^{3} \vg, \vh_{1}^{\otimes2}\otimes \vh_{2} \rangle
            \\
            +
            4 \langle \partial^{2} \vg, \vh_{1}\otimes \vh_{3} \rangle
            \\
            +
            3 \langle \partial^{2} \vg, \vh_{2}^{\otimes2} \rangle
            \\
            +
            \langle \partial \vg, \vh_{4} \rangle
          \end{matrix}
          =
    &\vf_{4}
      =
      \begin{matrix}
        \langle \partial^{4} \vf, \vx_{1}^{\otimes4} \rangle
        \\
        +
        6 \langle \partial^{3} \vf, \vx_{1}^{\otimes2}\otimes \vx_{2} \rangle
        \\
        +
        4 \langle \partial^{2} \vf, \vx_{1}\otimes \vx_{3} \rangle
        \\
        +
        3 \langle \partial^{2} \vf, \vx_{2}^{\otimes2} \rangle
        \\
        +
        \langle \partial \vf, \vx_{4} \rangle
      \end{matrix}
    \\
    \vx_{5}
    &\to&
          \vh_{5}
          =
          \begin{matrix}
            \langle \partial^{5} \vh, \vx_{1}^{\otimes5} \rangle
            \\
            +
            10 \langle \partial^{4} \vh, \vx_{1}^{\otimes3}\otimes \vx_{2} \rangle
            \\
            +
            10 \langle \partial^{3} \vh, \vx_{1}^{\otimes2}\otimes \vx_{3} \rangle
            \\
            +
            15 \langle \partial^{3} \vh, \vx_{1}\otimes \vx_{2}^{\otimes2} \rangle
            \\
            +
            5 \langle \partial^{2} \vh, \vx_{1}\otimes \vx_{4} \rangle
            \\
            +
            10 \langle \partial^{2} \vh, \vx_{2}\otimes \vx_{3} \rangle
            \\
            +
            \langle \partial \vh, \vx_{5} \rangle
          \end{matrix}
    &\to&
          \vg_{5}
          =
          \begin{matrix}
            \langle \partial^{5} \vg, \vh_{1}^{\otimes5} \rangle
            \\
            +
            10 \langle \partial^{4} \vg, \vh_{1}^{\otimes3}\otimes \vh_{2} \rangle
            \\
            +
            10 \langle \partial^{3} \vg, \vh_{1}^{\otimes2}\otimes \vh_{3} \rangle
            \\
            +
            15 \langle \partial^{3} \vg, \vh_{1}\otimes \vh_{2}^{\otimes2} \rangle
            \\
            +
            5 \langle \partial^{2} \vg, \vh_{1}\otimes \vh_{4} \rangle
            \\
            +
            10 \langle \partial^{2} \vg, \vh_{2}\otimes \vh_{3} \rangle
            \\
            +
            \langle \partial \vg, \vh_{5} \rangle
          \end{matrix}
          =
    &\vf_{5}
      =
      \begin{matrix}
        \langle \partial^{5} \vf, \vx_{1}^{\otimes5} \rangle
        \\
        +
        10 \langle \partial^{4} \vf, \vx_{1}^{\otimes3}\otimes \vx_{2} \rangle
        \\
        +
        10 \langle \partial^{3} \vf, \vx_{1}^{\otimes2}\otimes \vx_{3} \rangle
        \\
        +
        15 \langle \partial^{3} \vf, \vx_{1}\otimes \vx_{2}^{\otimes2} \rangle
        \\
        +
        5 \langle \partial^{2} \vf, \vx_{1}\otimes \vx_{4} \rangle
        \\
        +
        10 \langle \partial^{2} \vf, \vx_{2}\otimes \vx_{3} \rangle
        \\
        +
        \langle \partial \vf, \vx_{5} \rangle
      \end{matrix}
    \\
    \vx_{6}
    &\to&
          \vh_{6}
          =
          \begin{matrix}
            \langle \partial^{6} \vh, \vx_{1}^{\otimes6} \rangle
            \\
            +
            15 \langle \partial^{5} \vh, \vx_{1}^{\otimes4}\otimes \vx_{2} \rangle
            \\
            +
            20 \langle \partial^{4} \vh, \vx_{1}^{\otimes3}\otimes \vx_{3} \rangle
            \\
            +
            45 \langle \partial^{4} \vh, \vx_{1}^{\otimes2}\otimes \vx_{2}^{\otimes2} \rangle
            \\
            +
            15 \langle \partial^{3} \vh, \vx_{1}^{\otimes2}\otimes \vx_{4} \rangle
            \\
            +
            60 \langle \partial^{3} \vh, \vx_{1}\otimes \vx_{2}\otimes \vx_{3} \rangle
            \\
            +
            15 \langle \partial^{3} \vh, \vx_{2}^{\otimes3} \rangle
            \\
            +
            6 \langle \partial^{2} \vh, \vx_{1}\otimes \vx_{5} \rangle
            \\
            +
            15 \langle \partial^{2} \vh, \vx_{2}\otimes \vx_{4} \rangle
            \\
            +
            10 \langle \partial^{2} \vh, \vx_{3}^{\otimes2} \rangle
            \\
            +
            \langle \partial \vh, \vx_{6} \rangle
          \end{matrix}
    &\to&
          \vg_{6}
          =
          \begin{matrix}
            \langle \partial^{6} \vg, \vh_{1}^{\otimes6} \rangle
            \\
            +
            15 \langle \partial^{5} \vg, \vh_{1}^{\otimes4}\otimes \vh_{2} \rangle
            \\
            +
            20 \langle \partial^{4} \vg, \vh_{1}^{\otimes3}\otimes \vh_{3} \rangle
            \\
            +
            45 \langle \partial^{4} \vg, \vh_{1}^{\otimes2}\otimes \vh_{2}^{\otimes2} \rangle
            \\
            +
            15 \langle \partial^{3} \vg, \vh_{1}^{\otimes2}\otimes \vh_{4} \rangle
            \\
            +
            60 \langle \partial^{3} \vg, \vh_{1}\otimes \vh_{2}\otimes \vh_{3} \rangle
            \\
            +
            15 \langle \partial^{3} \vg, \vh_{2}^{\otimes3} \rangle
            \\
            +
            6 \langle \partial^{2} \vg, \vh_{1}\otimes \vh_{5} \rangle
            \\
            +
            15 \langle \partial^{2} \vg, \vh_{2}\otimes \vh_{4} \rangle
            \\
            +
            10 \langle \partial^{2} \vg, \vh_{3}^{\otimes2} \rangle
            \\
            +
            \langle \partial \vg, \vh_{6} \rangle
          \end{matrix}
          =
    &\vf_{6}
      =
      \begin{matrix}
        \langle \partial^{6} \vf, \vx_{1}^{\otimes6} \rangle
        \\
        +
        15 \langle \partial^{5} \vf, \vx_{1}^{\otimes4}\otimes \vx_{2} \rangle
        \\
        +
        20 \langle \partial^{4} \vf, \vx_{1}^{\otimes3}\otimes \vx_{3} \rangle
        \\
        +
        45 \langle \partial^{4} \vf, \vx_{1}^{\otimes2}\otimes \vx_{2}^{\otimes2} \rangle
        \\
        +
        15 \langle \partial^{3} \vf, \vx_{1}^{\otimes2}\otimes \vx_{4} \rangle
        \\
        +
        60 \langle \partial^{3} \vf, \vx_{1}\otimes \vx_{2}\otimes \vx_{3} \rangle
        \\
        +
        15 \langle \partial^{3} \vf, \vx_{2}^{\otimes3} \rangle
        \\
        +
        6 \langle \partial^{2} \vf, \vx_{1}\otimes \vx_{5} \rangle
        \\
        +
        15 \langle \partial^{2} \vf, \vx_{2}\otimes \vx_{4} \rangle
        \\
        +
        10 \langle \partial^{2} \vf, \vx_{3}^{\otimes2} \rangle
        \\
        +
        \langle \partial \vf, \vx_{6} \rangle
      \end{matrix}
    \\
    \vx_{7}
    &\to&
          \vh_{7}
          =
          \begin{matrix}
            \langle \partial^{7} \vh, \vx_{1}^{\otimes7} \rangle
            \\
            +
            21 \langle \partial^{6} \vh, \vx_{1}^{\otimes5}\otimes \vx_{2} \rangle
            \\
            +
            35 \langle \partial^{5} \vh, \vx_{1}^{\otimes4}\otimes \vx_{3} \rangle
            \\
            +
            105 \langle \partial^{5} \vh, \vx_{1}^{\otimes3}\otimes \vx_{2}^{\otimes2} \rangle
            \\
            +
            35 \langle \partial^{4} \vh, \vx_{1}^{\otimes3}\otimes \vx_{4} \rangle
            \\
            +
            210 \langle \partial^{4} \vh, \vx_{1}^{\otimes2}\otimes \vx_{2}\otimes \vx_{3} \rangle
            \\
            +
            105 \langle \partial^{4} \vh, \vx_{1}\otimes \vx_{2}^{\otimes3} \rangle
            \\
            +
            21 \langle \partial^{3} \vh, \vx_{1}^{\otimes2}\otimes \vx_{5} \rangle
            \\
            +
            105 \langle \partial^{3} \vh, \vx_{1}\otimes \vx_{2}\otimes \vx_{4} \rangle
            \\
            +
            70 \langle \partial^{3} \vh, \vx_{1}\otimes \vx_{3}^{\otimes2} \rangle
            \\
            +
            105 \langle \partial^{3} \vh, \vx_{2}^{\otimes2}\otimes \vx_{3} \rangle
            \\
            +
            7 \langle \partial^{2} \vh, \vx_{1}\otimes \vx_{6} \rangle
            \\
            +
            21 \langle \partial^{2} \vh, \vx_{2}\otimes \vx_{5} \rangle
            \\
            +
            35 \langle \partial^{2} \vh, \vx_{3}\otimes \vx_{4} \rangle
            \\
            +
            \langle \partial \vh, \vx_{7} \rangle
          \end{matrix}
    &\to&
          \vg_{7}
          =
          \begin{matrix}
            \langle \partial^{7} \vg, \vh_{1}^{\otimes7} \rangle
            \\
            +
            21 \langle \partial^{6} \vg, \vh_{1}^{\otimes5}\otimes \vh_{2} \rangle
            \\
            +
            35 \langle \partial^{5} \vg, \vh_{1}^{\otimes4}\otimes \vh_{3} \rangle
            \\
            +
            105 \langle \partial^{5} \vg, \vh_{1}^{\otimes3}\otimes \vh_{2}^{\otimes2} \rangle
            \\
            +
            35 \langle \partial^{4} \vg, \vh_{1}^{\otimes3}\otimes \vh_{4} \rangle
            \\
            +
            210 \langle \partial^{4} \vg, \vh_{1}^{\otimes2}\otimes \vh_{2}\otimes \vh_{3} \rangle
            \\
            +
            105 \langle \partial^{4} \vg, \vh_{1}\otimes \vh_{2}^{\otimes3} \rangle
            \\
            +
            21 \langle \partial^{3} \vg, \vh_{1}^{\otimes2}\otimes \vh_{5} \rangle
            \\
            +
            105 \langle \partial^{3} \vg, \vh_{1}\otimes \vh_{2}\otimes \vh_{4} \rangle
            \\
            +
            70 \langle \partial^{3} \vg, \vh_{1}\otimes \vh_{3}^{\otimes2} \rangle
            \\
            +
            105 \langle \partial^{3} \vg, \vh_{2}^{\otimes2}\otimes \vh_{3} \rangle
            \\
            +
            7 \langle \partial^{2} \vg, \vh_{1}\otimes \vh_{6} \rangle
            \\
            +
            21 \langle \partial^{2} \vg, \vh_{2}\otimes \vh_{5} \rangle
            \\
            +
            35 \langle \partial^{2} \vg, \vh_{3}\otimes \vh_{4} \rangle
            \\
            +
            \langle \partial \vg, \vh_{7} \rangle
          \end{matrix}
          =
    &\vf_{7}
      =
      \begin{matrix}
        \langle \partial^{7} \vf, \vx_{1}^{\otimes7} \rangle
        \\
        +
        21 \langle \partial^{6} \vf, \vx_{1}^{\otimes5}\otimes \vx_{2} \rangle
        \\
        +
        35 \langle \partial^{5} \vf, \vx_{1}^{\otimes4}\otimes \vx_{3} \rangle
        \\
        +
        105 \langle \partial^{5} \vf, \vx_{1}^{\otimes3}\otimes \vx_{2}^{\otimes2} \rangle
        \\
        +
        35 \langle \partial^{4} \vf, \vx_{1}^{\otimes3}\otimes \vx_{4} \rangle
        \\
        +
        210 \langle \partial^{4} \vf, \vx_{1}^{\otimes2}\otimes \vx_{2}\otimes \vx_{3} \rangle
        \\
        +
        105 \langle \partial^{4} \vf, \vx_{1}\otimes \vx_{2}^{\otimes3} \rangle
        \\
        +
        21 \langle \partial^{3} \vf, \vx_{1}^{\otimes2}\otimes \vx_{5} \rangle
        \\
        +
        105 \langle \partial^{3} \vf, \vx_{1}\otimes \vx_{2}\otimes \vx_{4} \rangle
        \\
        +
        70 \langle \partial^{3} \vf, \vx_{1}\otimes \vx_{3}^{\otimes2} \rangle
        \\
        +
        105 \langle \partial^{3} \vf, \vx_{2}^{\otimes2}\otimes \vx_{3} \rangle
        \\
        +
        7 \langle \partial^{2} \vf, \vx_{1}\otimes \vx_{6} \rangle
        \\
        +
        21 \langle \partial^{2} \vf, \vx_{2}\otimes \vx_{5} \rangle
        \\
        +
        35 \langle \partial^{2} \vf, \vx_{3}\otimes \vx_{4} \rangle
        \\
        +
        \langle \partial \vf, \vx_{7} \rangle
      \end{matrix}
    \\
    \vx_{8}
    &\to&
          \vh_{8}
          =
          \begin{matrix}
            \langle \partial^{8} \vh, \vx_{1}^{\otimes8} \rangle
            \\
            +
            28 \langle \partial^{7} \vh, \vx_{1}^{\otimes6}\otimes \vx_{2} \rangle
            \\
            +
            56 \langle \partial^{6} \vh, \vx_{1}^{\otimes5}\otimes \vx_{3} \rangle
            \\
            +
            210 \langle \partial^{6} \vh, \vx_{1}^{\otimes4}\otimes \vx_{2}^{\otimes2} \rangle
            \\
            +
            70 \langle \partial^{5} \vh, \vx_{1}^{\otimes4}\otimes \vx_{4} \rangle
            \\
            +
            560 \langle \partial^{5} \vh, \vx_{1}^{\otimes3}\otimes \vx_{2}\otimes \vx_{3} \rangle
            \\
            +
            420 \langle \partial^{5} \vh, \vx_{1}^{\otimes2}\otimes \vx_{2}^{\otimes3} \rangle
            \\
            +
            56 \langle \partial^{4} \vh, \vx_{1}^{\otimes3}\otimes \vx_{5} \rangle
            \\
            +
            420 \langle \partial^{4} \vh, \vx_{1}^{\otimes2}\otimes \vx_{2}\otimes \vx_{4} \rangle
            \\
            +
            280 \langle \partial^{4} \vh, \vx_{1}^{\otimes2}\otimes \vx_{3}^{\otimes2} \rangle
            \\
            +
            840 \langle \partial^{4} \vh, \vx_{1}\otimes \vx_{2}^{\otimes2}\otimes \vx_{3} \rangle
            \\
            +
            105 \langle \partial^{4} \vh, \vx_{2}^{\otimes4} \rangle
            \\
            +
            28 \langle \partial^{3} \vh, \vx_{1}^{\otimes2}\otimes \vx_{6} \rangle
            \\
            +
            168 \langle \partial^{3} \vh, \vx_{1}\otimes \vx_{2}\otimes \vx_{5} \rangle
            \\
            +
            280 \langle \partial^{3} \vh, \vx_{1}\otimes \vx_{3}\otimes \vx_{4} \rangle
            \\
            +
            210 \langle \partial^{3} \vh, \vx_{2}^{\otimes2}\otimes \vx_{4} \rangle
            \\
            +
            280 \langle \partial^{3} \vh, \vx_{2}\otimes \vx_{3}^{\otimes2} \rangle
            \\
            +
            8 \langle \partial^{2} \vh, \vx_{1}\otimes \vx_{7} \rangle
            \\
            +
            28 \langle \partial^{2} \vh, \vx_{2}\otimes \vx_{6} \rangle
            \\
            +
            56 \langle \partial^{2} \vh, \vx_{3}\otimes \vx_{5} \rangle
            \\
            +
            35 \langle \partial^{2} \vh, \vx_{4}^{\otimes2} \rangle
            \\
            +
            \langle \partial \vh, \vx_{8} \rangle
          \end{matrix}
    &\to&
          \vg_{8}
          =
          \begin{matrix}
            \langle \partial^{8} \vg, \vh_{1}^{\otimes8} \rangle
            \\
            +
            28 \langle \partial^{7} \vg, \vh_{1}^{\otimes6}\otimes \vh_{2} \rangle
            \\
            +
            56 \langle \partial^{6} \vg, \vh_{1}^{\otimes5}\otimes \vh_{3} \rangle
            \\
            +
            210 \langle \partial^{6} \vg, \vh_{1}^{\otimes4}\otimes \vh_{2}^{\otimes2} \rangle
            \\
            +
            70 \langle \partial^{5} \vg, \vh_{1}^{\otimes4}\otimes \vh_{4} \rangle
            \\
            +
            560 \langle \partial^{5} \vg, \vh_{1}^{\otimes3}\otimes \vh_{2}\otimes \vh_{3} \rangle
            \\
            +
            420 \langle \partial^{5} \vg, \vh_{1}^{\otimes2}\otimes \vh_{2}^{\otimes3} \rangle
            \\
            +
            56 \langle \partial^{4} \vg, \vh_{1}^{\otimes3}\otimes \vh_{5} \rangle
            \\
            +
            420 \langle \partial^{4} \vg, \vh_{1}^{\otimes2}\otimes \vh_{2}\otimes \vh_{4} \rangle
            \\
            +
            280 \langle \partial^{4} \vg, \vh_{1}^{\otimes2}\otimes \vh_{3}^{\otimes2} \rangle
            \\
            +
            840 \langle \partial^{4} \vg, \vh_{1}\otimes \vh_{2}^{\otimes2}\otimes \vh_{3} \rangle
            \\
            +
            105 \langle \partial^{4} \vg, \vh_{2}^{\otimes4} \rangle
            \\
            +
            28 \langle \partial^{3} \vg, \vh_{1}^{\otimes2}\otimes \vh_{6} \rangle
            \\
            +
            168 \langle \partial^{3} \vg, \vh_{1}\otimes \vh_{2}\otimes \vh_{5} \rangle
            \\
            +
            280 \langle \partial^{3} \vg, \vh_{1}\otimes \vh_{3}\otimes \vh_{4} \rangle
            \\
            +
            210 \langle \partial^{3} \vg, \vh_{2}^{\otimes2}\otimes \vh_{4} \rangle
            \\
            +
            280 \langle \partial^{3} \vg, \vh_{2}\otimes \vh_{3}^{\otimes2} \rangle
            \\
            +
            8 \langle \partial^{2} \vg, \vh_{1}\otimes \vh_{7} \rangle
            \\
            +
            28 \langle \partial^{2} \vg, \vh_{2}\otimes \vh_{6} \rangle
            \\
            +
            56 \langle \partial^{2} \vg, \vh_{3}\otimes \vh_{5} \rangle
            \\
            +
            35 \langle \partial^{2} \vg, \vh_{4}^{\otimes2} \rangle
            \\
            +
            \langle \partial \vg, \vh_{8} \rangle
          \end{matrix}
          =
    &\vf_{8}
      =
      \begin{matrix}
        \langle \partial^{8} \vf, \vx_{1}^{\otimes8} \rangle
        \\
        +
        28 \langle \partial^{7} \vf, \vx_{1}^{\otimes6}\otimes \vx_{2} \rangle
        \\
        +
        56 \langle \partial^{6} \vf, \vx_{1}^{\otimes5}\otimes \vx_{3} \rangle
        \\
        +
        210 \langle \partial^{6} \vf, \vx_{1}^{\otimes4}\otimes \vx_{2}^{\otimes2} \rangle
        \\
        +
        70 \langle \partial^{5} \vf, \vx_{1}^{\otimes4}\otimes \vx_{4} \rangle
        \\
        +
        560 \langle \partial^{5} \vf, \vx_{1}^{\otimes3}\otimes \vx_{2}\otimes \vx_{3} \rangle
        \\
        +
        420 \langle \partial^{5} \vf, \vx_{1}^{\otimes2}\otimes \vx_{2}^{\otimes3} \rangle
        \\
        +
        56 \langle \partial^{4} \vf, \vx_{1}^{\otimes3}\otimes \vx_{5} \rangle
        \\
        +
        420 \langle \partial^{4} \vf, \vx_{1}^{\otimes2}\otimes \vx_{2}\otimes \vx_{4} \rangle
        \\
        +
        280 \langle \partial^{4} \vf, \vx_{1}^{\otimes2}\otimes \vx_{3}^{\otimes2} \rangle
        \\
        +
        840 \langle \partial^{4} \vf, \vx_{1}\otimes \vx_{2}^{\otimes2}\otimes \vx_{3} \rangle
        \\
        +
        105 \langle \partial^{4} \vf, \vx_{2}^{\otimes4} \rangle
        \\
        +
        28 \langle \partial^{3} \vf, \vx_{1}^{\otimes2}\otimes \vx_{6} \rangle
        \\
        +
        168 \langle \partial^{3} \vf, \vx_{1}\otimes \vx_{2}\otimes \vx_{5} \rangle
        \\
        +
        280 \langle \partial^{3} \vf, \vx_{1}\otimes \vx_{3}\otimes \vx_{4} \rangle
        \\
        +
        210 \langle \partial^{3} \vf, \vx_{2}^{\otimes2}\otimes \vx_{4} \rangle
        \\
        +
        280 \langle \partial^{3} \vf, \vx_{2}\otimes \vx_{3}^{\otimes2} \rangle
        \\
        +
        8 \langle \partial^{2} \vf, \vx_{1}\otimes \vx_{7} \rangle
        \\
        +
        28 \langle \partial^{2} \vf, \vx_{2}\otimes \vx_{6} \rangle
        \\
        +
        56 \langle \partial^{2} \vf, \vx_{3}\otimes \vx_{5} \rangle
        \\
        +
        35 \langle \partial^{2} \vf, \vx_{4}^{\otimes2} \rangle
        \\
        +
        \langle \partial \vf, \vx_{8} \rangle
      \end{matrix}
  \end{align*}
\end{tiny}

\clearpage
\section{Visual Tour: From Function to Collapsed Taylor Mode}\label{sec:appendix-visual-tour}
\input{figures/interface_overview}

\clearpage
\section{Graph Simplifications}\label{sec:graph-simplifications}
In this section, we illustrate the two graph simplifications that are required to collapse Taylor mode.

We will consider collapsing the 2-jet of $f = \sin$ as an example.
Recall the propagation scheme \cref{eq:sum-taylor-mode-naive} and assume that the Taylor coefficients are given by $\{\vx_{0,r} = \vx_0\}$, $\{\vx_{1,r}\}$, and $\{\vx_{2,r}\}$ where $r$ indexes the directions along which we evaluate the sum:
\begin{align*}
  \begin{matrix}
    \vx_0
    \\
    \{\vx_{1,r}\}
    \\
    \{\vx_{2,r}\}
  \end{matrix}
  &\overset{\text{replicate $\vx_0$}}{\to}
    \begin{Bmatrix}
      \vx_{0,r} = \vx_0
      \\
      \vx_{1,r}
      \\
      \vx_{2,r}
    \end{Bmatrix}
    \overset{\eqref{eq:sum-taylor-mode-naive}}{\to}
    \begin{Bmatrix}
      \vf_{0,r} = \sin(\vx_0)
      \\
      \vf_{1,r} = \cos(\vx_0) \odot \vx_{1,r}
      \\
      \vf_{2,r} = -\sin(\vx_0) \odot \vx_{1,r} \odot \vx_{1, r} + \cos(\vx_0) \odot \vx_{2,r}
    \end{Bmatrix}
  \\
  &\overset{\text{sum highest component}}{\to}
    \begin{matrix}
      \begin{Bmatrix}
        \vf_{0,r}
        \\
        \vf_{1,r}
      \end{Bmatrix}
      \\
      \sum_r \vf_{2,r}
    \end{matrix}
\end{align*}
Here, $\sin$ applies element-wise and $\odot$ denotes element-wise multiplication.
The computational graph for this procedure is displayed in the following diagram, with input and output nodes highlighted in dark and light gray. The suffix \texttt{\_r} means that all $R$ corresponding tensors are stacked along their leading axis.
$\texttt{replicate}$ is a function that replicates a tensor $R$ times along a new leading axis, which is in PyTorch usually for free and without additional memory overhead (using \texttt{torch.expand}). All other functions refer to those of the PyTorch API:

\begin{figure}[!h]
  \centering
  \scalebox{0.66}{{
  \tikzset{state/.style={rectangle, rounded corners, align=left, font=\ttfamily, draw=black, anchor=west, align=left, inner sep=5pt}}
  \tikzset{leaf/.style={fill=gray!50!white}}
  \tikzset{output/.style={fill=gray!25!white}}
  \begin{tikzpicture}
    \matrix [%
    ampersand replacement=\&,%
    column sep=5ex,%
    row sep=4ex,%
    ]{
      \node[state, leaf] (x0) {x0};
      \& \node[state] (x0-d) {x0\_r = replicate(x0)};
      \&
      \& \node[state, output] (f0-d) {f0\_r = sin(x0\_r)};
      \\
      \& \node[state] (df-d) {df\_r = cos(x0\_r)};
      \& \node[state] (d2f-d) {d2f\_r = -f0\_r};
      \\
      \node[state, leaf] (x1-d) {x1\_r};
      \&
      \& \node[state] (temp1-d) {temp1\_r = einsum(``...,...,...\\->...'', d2f\_r, x1\_r, x1\_r)};
      \& \node[state, output] (f1-d) {f1\_r =
        einsum(``...,...\\->...'', df\_r, x1\_r)};
      \\
      \node[state, leaf] (x2-d) {x2\_r};
      \& \node[state] (temp2-d) {temp2\_r = einsum(``...,...\\->...'', df\_r, x2\_r)};
      \& \node[state] (f2-d) {f2\_r = temp1\_r + temp2\_r};
      \& \node[state, output] (f2) {f2 = sum(f2\_r, dim=0)};
      \\
    };

    \tikzset{arrow/.style={-Stealth}}
    \path
    (x0) edge [arrow] (x0-d)
    (x0-d) edge [arrow] (f0-d)
    (x0-d) edge [arrow] (df-d)
    (f0-d) edge [arrow] (d2f-d)
    (df-d) edge [arrow, out=-8, in=169] (f1-d)
    (x1-d) edge [arrow, out=10, in=167] (f1-d)
    (df-d) edge [arrow] (temp2-d)
    (d2f-d) edge [arrow] (temp1-d)
    (x2-d) edge [arrow] (temp2-d)
    (x1-d) edge [arrow] (temp1-d)
    (temp1-d) edge [arrow] (f2-d)
    (temp2-d) edge [arrow] (f2-d)
    (f2-d) edge [arrow] (f2)
    ;
  \end{tikzpicture}
}}
\end{figure}

Our simplification proceeds in two steps.
First, propagate \texttt{replicate} nodes down the graph to remove repeated computations on the same tensors. This is done in a forward traversal through the graph.
Second, in a single backward traversal through the graph, we propagate the \texttt{sum} node up.
After applying both steps, the graph looks as follows:

\begin{figure}[!h]
  \centering
  \scalebox{0.66}{
    {
  \tikzset{state/.style={rectangle, rounded corners, align=left, font=\ttfamily, draw=black, anchor=west, align=left, inner sep=5pt}}
  \tikzset{leaf/.style={fill=gray!50!white}}
  \tikzset{output/.style={fill=gray!25!white}}
  \begin{tikzpicture}
    \matrix [%
    ampersand replacement=\&,%
    column sep=5ex,%
    row sep=4ex,%
    ]{
      \node[state, leaf] (x0) {x0};
      \& \node[state] (x0-d) {f0 = sin(x0)};
      \&
      \& \node[state, output] (f0-d) {f0\_r = replicate(f0)};
      \\
      \& \node[state] (df-d) {df = cos(x0)};
      \& \node[state] (d2f-d) {d2f = -f0};
      \\
      \node[state, leaf] (x1-d) {x1\_r};
      \&
      \& \node[state] (temp1-d) {temp1 = einsum(``...,r...,r...\\->...'', d2f, x1\_r, x1\_r)};
      \& \node[state, output] (f1-d) {f1\_r =
        einsum(``r...,...\\->r...'', df, x1\_r)};
      \\
      \node[state, leaf] (x2-d) {x2\_r};
      \& \node[state] (temp2-d) {x2 = sum(x2\_r, dim=0)};
      \& \node[state] (f2-d) {temp2 = einsum(``...,...\\->...'', df, x2)};
      \& \node[state, output] (f2) {f2 = temp1 + temp2};
      \\
    };

    \tikzset{arrow/.style={-Stealth}}
    \path
    (x0) edge [arrow] (x0-d)
    (x0-d) edge [arrow] (f0-d)
    (x0) edge [arrow] (df-d)
    (x0-d) edge [arrow] (d2f-d)
    (df-d) edge [arrow, out=-8, in=169] (f1-d)
    (x1-d) edge [arrow, out=10, in=167] (f1-d)
    (df-d) edge [arrow, out=-60, in=150] (f2-d)
    (d2f-d) edge [arrow] (temp1-d)
    (x2-d) edge [arrow] (temp2-d)
    (x1-d) edge [arrow] (temp1-d)
    (temp1-d) edge [arrow] (f2)
    (temp2-d) edge [arrow] (f2-d)
    (f2-d) edge [arrow] (f2)
    ;
  \end{tikzpicture}
}
  }
\end{figure}

Two important properties of the new graph are (i) the \texttt{replicate} node moved to an output node, hence the corresponding redundant computation was successfully removed (ii) the highest component \texttt{x2\_r} is immediately summed then propagated, \ie, we collapsed Taylor mode and avoid the separate propagation for all \texttt{x2\_r}.

We will now illustrate the two simplification steps in full detail.
The first stage starts from the original graph and pushes forward the replicate node, as illustrated step-by-step in \cref{fig:push-replicate-simplification}.
The second stage starts from the graph produced by the replicate-push procedure, and propagates the final sum node up the graph, illustrated by \cref{fig:pull-sum-simplification}.
This yields the final computation graph shown above.

\begin{figure}[!h]
  \scalebox{0.70}{
    {
  \tikzset{state/.style={rectangle, rounded corners, align=left, font=\ttfamily, draw=black, anchor=west, align=left, inner sep=5pt}}
  \tikzset{leaf/.style={fill=gray!50!white}}
  \tikzset{output/.style={fill=gray!25!white}}
  \tikzset{replicate/.style={fill=blue!25!white}}
  \begin{tikzpicture}
    \matrix [%
    ampersand replacement=\&,%
    column sep=5ex,%
    row sep=4ex,%
    ]{
      \node[state, leaf] (x0) {x0};
      \& \node[state, replicate] (x0-d) {x0\_r = replicate(x0)};
      \&
      \& \node[state, output] (f0-d) {f0\_r = sin(x0\_r)};
      \\
      \& \node[state] (df-d) {df\_r = cos(x0\_r)};
      \& \node[state] (d2f-d) {d2f\_r = -f0\_r};
      \\
      \node[state, leaf] (x1-d) {x1\_r};
      \&
      \& \node[state] (temp1-d) {temp1\_r = einsum(``...,...,...\\->...'', d2f\_r, x1\_r, x1\_r)};
      \& \node[state, output] (f1-d) {f1\_r =
        einsum(``...,...\\->...'', df\_r, x1\_r)};
      \\
      \node[state, leaf] (x2-d) {x2\_r};
      \& \node[state] (temp2-d) {temp2\_r = einsum(``...,...\\->...'', df\_r, x2\_r)};
      \& \node[state] (f2-d) {f2\_r = temp1\_r + temp2\_r};
      \& \node[state, output] (f2) {f2 = sum(f2\_r, dim=0)};
      \\
    };

    \tikzset{arrow/.style={-Stealth}}
    \path
    (x0) edge [arrow] (x0-d)
    (x0-d) edge [arrow] (f0-d)
    (x0-d) edge [arrow] (df-d)
    (f0-d) edge [arrow] (d2f-d)
    (df-d) edge [arrow, out=-8, in=169] (f1-d)
    (x1-d) edge [arrow, out=10, in=167] (f1-d)
    (df-d) edge [arrow] (temp2-d)
    (d2f-d) edge [arrow] (temp1-d)
    (x2-d) edge [arrow] (temp2-d)
    (x1-d) edge [arrow] (temp1-d)
    (temp1-d) edge [arrow] (f2-d)
    (temp2-d) edge [arrow] (f2-d)
    (f2-d) edge [arrow] (f2)
    ;
  \end{tikzpicture}
}
  }
  \vspace{-0.6ex}
  \noindent\rule{\textwidth}{1pt}
  \vspace{-0.6ex}

  \scalebox{0.70}{
    {
  \tikzset{state/.style={rectangle, rounded corners, align=left, font=\ttfamily, draw=black, anchor=west, align=left, inner sep=5pt}}
  \tikzset{leaf/.style={fill=gray!50!white}}
  \tikzset{output/.style={fill=gray!25!white}}
  \tikzset{replicate/.style={fill=blue!25!white}}
  \begin{tikzpicture}
    \matrix [%
    ampersand replacement=\&,%
    column sep=5ex,%
    row sep=4ex,%
    ]{
      \node[state, leaf] (x0) {x0};
      \& \node[state] (x0-d) {f0 = sin(x0)};
      \&
      \& \node[state, replicate] (f0-d) {f0\_r = replicate(f0)};
      \\
      \& \node[state, replicate] (x0-r) {x0\_r = replicate(x0)};
      \\
      \& \node[state] (df-d) {df\_r = cos(x0\_r)};
      \& \node[state] (d2f-d) {d2f\_r = -f0\_r};
      \\
      \node[state, leaf] (x1-d) {x1\_r};
      \&
      \& \node[state] (temp1-d) {temp1\_r = einsum(``...,...,...\\->...'', d2f\_r, x1\_r, x1\_r)};
      \& \node[state, output] (f1-d) {f1\_r =
        einsum(``...,...\\->...'', df\_r, x1\_r)};
      \\
      \node[state, leaf] (x2-d) {x2\_r};
      \& \node[state] (temp2-d) {temp2\_r = einsum(``...,...\\->...'', df\_r, x2\_r)};
      \& \node[state] (f2-d) {f2\_r = temp1\_r + temp2\_r};
      \& \node[state, output] (f2) {f2 = sum(f2\_r, dim=0)};
      \\
    };

    \tikzset{arrow/.style={-Stealth}}
    \path
    (x0) edge [arrow] (x0-d)
    (x0) edge [arrow] (x0-r)
    (x0-d) edge [arrow] (f0-d)
    (x0-r) edge [arrow] (df-d)
    (f0-d) edge [arrow] (d2f-d)
    (df-d) edge [arrow, out=-8, in=169] (f1-d)
    (x1-d) edge [arrow, out=10, in=167] (f1-d)
    (df-d) edge [arrow] (temp2-d)
    (d2f-d) edge [arrow] (temp1-d)
    (x2-d) edge [arrow] (temp2-d)
    (x1-d) edge [arrow] (temp1-d)
    (temp1-d) edge [arrow] (f2-d)
    (temp2-d) edge [arrow] (f2-d)
    (f2-d) edge [arrow] (f2)
    ;
  \end{tikzpicture}
}
  }

  \vspace{-0.6ex}
  \noindent\rule{\textwidth}{1pt}
  \vspace{-0.6ex}

  \scalebox{0.70}{
    {
  \tikzset{state/.style={rectangle, rounded corners, align=left, font=\ttfamily, draw=black, anchor=west, align=left, inner sep=5pt}}
  \tikzset{leaf/.style={fill=gray!50!white}}
  \tikzset{output/.style={fill=gray!25!white}}
  \tikzset{replicate/.style={fill=blue!25!white}}
  \begin{tikzpicture}
    \matrix [%
    ampersand replacement=\&,%
    column sep=5ex,%
    row sep=4ex,%
    ]{
      \node[state, leaf] (x0) {x0};
      \& \node[state] (x0-d) {f0 = sin(x0)};
      \&
      \& \node[state, replicate] (f0-d) {f0\_r = replicate(f0)};
      \\
      \& \node[state] (df-d) {df = cos(x0)};
      \& \node[state, replicate] (d2f-d) {d2f\_r = replicate(-f0)};
      \\
      \& \node[state, replicate] (df-r) {df\_r = replicate(df)};
      \\
      \node[state, leaf] (x1-d) {x1\_r};
      \&
      \& \node[state] (temp1-d) {temp1\_r = einsum(``...,...,...\\->...'', d2f\_r, x1\_r, x1\_r)};
      \& \node[state, output] (f1-d) {f1\_r =
        einsum(``...,...\\->...'', df\_r, x1\_r)};
      \\
      \node[state, leaf] (x2-d) {x2\_r};
      \& \node[state] (temp2-d) {temp2\_r = einsum(``...,...\\->...'', df\_r, x2\_r)};
      \& \node[state] (f2-d) {f2\_r = temp1\_r + temp2\_r};
      \& \node[state, output] (f2) {f2 = sum(f2\_r, dim=0)};
      \\
    };

    \tikzset{arrow/.style={-Stealth}}
    \path
    (x0) edge [arrow] (x0-d)
    (x0) edge [arrow] (x0-r)
    (x0-d) edge [arrow] (f0-d)
    (df-d) edge [arrow] (df-r)
    (x0-d) edge [arrow] (d2f-d)
    (df-r) edge [arrow, out=-8, in=169] (f1-d)
    (x1-d) edge [arrow, out=10, in=167] (f1-d)
    (df-r) edge [arrow] (temp2-d)
    (d2f-d) edge [arrow] (temp1-d)
    (x2-d) edge [arrow] (temp2-d)
    (x1-d) edge [arrow] (temp1-d)
    (temp1-d) edge [arrow] (f2-d)
    (temp2-d) edge [arrow] (f2-d)
    (f2-d) edge [arrow] (f2)
    ;
  \end{tikzpicture}
}
  }

  \vspace{-0.6ex}
  \noindent\rule{\textwidth}{1pt}
  \vspace{-0.6ex}
  \scalebox{0.70}{
    {
  \tikzset{state/.style={rectangle, rounded corners, align=left, font=\ttfamily, draw=black, anchor=west, align=left, inner sep=5pt}}
  \tikzset{leaf/.style={fill=gray!50!white}}
  \tikzset{output/.style={fill=gray!25!white}}
  \tikzset{replicate/.style={fill=blue!25!white}}
  \begin{tikzpicture}
    \matrix [%
    ampersand replacement=\&,%
    column sep=5ex,%
    row sep=4ex,%
    ]{
      \node[state, leaf] (x0) {x0};
      \& \node[state] (x0-d) {f0 = sin(x0)};
      \&
      \& \node[state, replicate] (f0-d) {f0\_r = replicate(f0)};
      \\
      \& \node[state] (df-d) {df = cos(x0)};
      \& \node[state] (d2f-d) {d2f= -f0};
      \\
      \node[state, leaf] (x1-d) {x1\_r};
      \&
      \& \node[state, replicate] (temp1-d) {temp1\_r = einsum(``...,r...,r...\\->r...'', d2f, x1\_r, x1\_r)};
      \& \node[state, output] (f1-d) {f1\_r =
        einsum(``...,r...\\->r...'', df, x1\_r)};
      \\
      \node[state, leaf] (x2-d) {x2\_r};
      \& \node[state, replicate] (temp2-d) {temp2\_r = einsum(``...,r...\\->r...'', df, x2\_r)};
      \& \node[state] (f2-d) {f2\_r = temp1\_r + temp2\_r};
      \& \node[state, output] (f2) {f2 = sum(f2\_r, dim=0)};
      \\
    };

    \tikzset{arrow/.style={-Stealth}}
    \path
    (x0) edge [arrow] (x0-d)
    (x0) edge [arrow] (df-d)
    (x0-d) edge [arrow] (f0-d)
    (df-d) edge [arrow] (temp2-d)
    (x0-d) edge [arrow] (d2f-d)
    (df-d) edge [arrow, out=-8, in=169] (f1-d)
    (x1-d) edge [arrow, out=10, in=167] (f1-d)
    (d2f-d) edge [arrow] (temp1-d)
    (x2-d) edge [arrow] (temp2-d)
    (x1-d) edge [arrow] (temp1-d)
    (temp1-d) edge [arrow] (f2-d)
    (temp2-d) edge [arrow] (f2-d)
    (f2-d) edge [arrow] (f2)
    ;
  \end{tikzpicture}
}
      }

  \caption{\textbf{Step-by-step illustration of pushing \texttt{replicate} nodes down a computation graph.}}
  \label{fig:push-replicate-simplification}
\end{figure}

\begin{figure}[!h]
  \centering

  \scalebox{0.70}{
    {
  \tikzset{state/.style={rectangle, rounded corners, align=left, font=\ttfamily, draw=black, anchor=west, align=left, inner sep=5pt}}
  \tikzset{leaf/.style={fill=gray!50!white}}
  \tikzset{output/.style={fill=gray!25!white}}
  \tikzset{sum/.style={fill=green!25!white}}
  \begin{tikzpicture}
    \matrix [%
    ampersand replacement=\&,%
    column sep=5ex,%
    row sep=4ex,%
    ]{
      \node[state, leaf] (x0) {x0};
      \& \node[state] (x0-d) {f0 = sin(x0)};
      \&
      \& \node[state, output] (f0-d) {f0\_r = replicate(f0)};
      \\
      \& \node[state] (df-d) {df = cos(x0)};
      \& \node[state] (d2f-d) {d2f= -f0};
      \\
      \node[state, leaf] (x1-d) {x1\_r};
      \&
      \& \node[state] (temp1-d) {temp1\_r = einsum(``...,r...,r...\\->r...'', d2f, x1\_r, x1\_r)};
      \& \node[state, output] (f1-d) {f1\_r =
        einsum(``...,r...\\->r...'', df, x1\_r)};
      \\
      \node[state, leaf] (x2-d) {x2\_r};
      \& \node[state] (temp2-d) {temp2\_r = einsum(``...,r...\\->r...'', df, x2\_r)};
      \& \node[state] (f2-d) {f2\_r = temp1\_r + temp2\_r};
      \& \node[state, sum] (f2) {f2 = sum(f2\_r, dim=0)};
      \\
    };

    \tikzset{arrow/.style={-Stealth}}
    \path
    (x0) edge [arrow] (x0-d)
    (x0) edge [arrow] (df-d)
    (x0-d) edge [arrow] (f0-d)
    (df-d) edge [arrow] (temp2-d)
    (x0-d) edge [arrow] (d2f-d)
    (df-d) edge [arrow, out=-8, in=169] (f1-d)
    (x1-d) edge [arrow, out=10, in=167] (f1-d)
    (d2f-d) edge [arrow] (temp1-d)
    (x2-d) edge [arrow] (temp2-d)
    (x1-d) edge [arrow] (temp1-d)
    (temp1-d) edge [arrow] (f2-d)
    (temp2-d) edge [arrow] (f2-d)
    (f2-d) edge [arrow] (f2)
    ;
  \end{tikzpicture}
}
  }

  \vspace{-0.6ex}
  \noindent\rule{\textwidth}{1pt}
  \vspace{-0.6ex}

  \scalebox{0.70}{
    {
  \tikzset{state/.style={rectangle, rounded corners, align=left, font=\ttfamily, draw=black, anchor=west, align=left, inner sep=5pt}}
  \tikzset{leaf/.style={fill=gray!50!white}}
  \tikzset{output/.style={fill=gray!25!white}}
  \tikzset{sum/.style={fill=green!25!white}}
  \begin{tikzpicture}
    \matrix [%
    ampersand replacement=\&,%
    column sep=5ex,%
    row sep=4ex,%
    ]{
      \node[state, leaf] (x0) {x0};
      \& \node[state] (x0-d) {f0 = sin(x0)};
      \&
      \& \node[state, output] (f0-d) {f0\_r = replicate(f0)};
      \\
      \& \node[state] (df-d) {df = cos(x0)};
      \& \node[state] (d2f-d) {d2f= -f0};
      \\
      \node[state, leaf] (x1-d) {x1\_r};
      \&
      \& \node[state] (temp1-d) {temp1\_r = einsum(``...,r...,r...\\->r...'', d2f, x1\_r, x1\_r)};
      \& \node[state, output] (f1-d) {f1\_r =
        einsum(``...,r...\\->r...'', df, x1\_r)};
      \\
      \&
      \& \node[state, sum] (temp1) {temp1 = sum(temp1\_r, dim=0)};
      \\
      \node[state, leaf] (x2-d) {x2\_r};
      \& \node[state] (temp2-d) {temp2\_r = einsum(``...,r...\\->r...'', df, x2\_r)};
      \& \node[state, sum] (temp2) {temp2 = sum(temp2\_r, dim=0)};
      \& \node[state, output] (f2) {f2 = temp1 + temp2};
      \\
    };

    \tikzset{arrow/.style={-Stealth}}
    \path
    (x0) edge [arrow] (x0-d)
    (x0) edge [arrow] (df-d)
    (x0-d) edge [arrow] (f0-d)
    (df-d) edge [arrow] (temp2-d)
    (x0-d) edge [arrow] (d2f-d)
    (df-d) edge [arrow, out=-8, in=169] (f1-d)
    (x1-d) edge [arrow, out=10, in=167] (f1-d)
    (d2f-d) edge [arrow] (temp1-d)
    (x2-d) edge [arrow] (temp2-d)
    (x1-d) edge [arrow] (temp1-d)
    (temp1-d) edge [arrow] (temp1)
    (temp1) edge [arrow] (f2)
    (temp2-d) edge [arrow] (temp2)
    (temp2) edge [arrow] (f2)
    ;
  \end{tikzpicture}
}
  }

  \vspace{-0.6ex}
  \noindent\rule{\textwidth}{1pt}
  \vspace{-0.6ex}

  \scalebox{0.70}{
    {
  \tikzset{state/.style={rectangle, rounded corners, align=left, font=\ttfamily, draw=black, anchor=west, align=left, inner sep=5pt}}
  \tikzset{leaf/.style={fill=gray!50!white}}
  \tikzset{output/.style={fill=gray!25!white}}
  \tikzset{sum/.style={fill=green!25!white}}
  \begin{tikzpicture}
    \matrix [%
    ampersand replacement=\&,%
    column sep=5ex,%
    row sep=4ex,%
    ]{
      \node[state, leaf] (x0) {x0};
      \& \node[state] (x0-d) {f0 = sin(x0)};
      \&
      \& \node[state, output] (f0-d) {f0\_r = replicate(f0)};
      \\
      \& \node[state] (df-d) {df = cos(x0)};
      \& \node[state] (d2f-d) {d2f= -f0};
      \\
      \node[state, leaf] (x1-d) {x1\_r};
      \&
      \& \node[state, sum] (temp1-d) {temp1 = einsum(``...,r...,r...\\->...'', d2f, x1\_r, x1\_r)};
      \& \node[state, output] (f1-d) {f1\_r =
        einsum(``...,r...\\->r...'', df, x1\_r)};
      \\
      \node[state, leaf] (x2-d) {x2\_r};
      \& \node[state, sum] (temp2-d) {temp2 = einsum(``...,r...\\->...'', df, x2\_r)};
      \&
      \& \node[state, output] (f2) {f2 = temp1 + temp2};
      \\
    };

    \tikzset{arrow/.style={-Stealth}}
    \path
    (x0) edge [arrow] (x0-d)
    (x0) edge [arrow] (df-d)
    (x0-d) edge [arrow] (f0-d)
    (df-d) edge [arrow] (temp2-d)
    (x0-d) edge [arrow] (d2f-d)
    (df-d) edge [arrow, out=-8, in=169] (f1-d)
    (x1-d) edge [arrow, out=10, in=167] (f1-d)
    (d2f-d) edge [arrow] (temp1-d)
    (x2-d) edge [arrow] (temp2-d)
    (x1-d) edge [arrow] (temp1-d)
    (temp1-d) edge [arrow] (f2)
    (temp2-d) edge [arrow] (f2)
    ;
  \end{tikzpicture}
}
  }

  \vspace{-0.6ex}
  \noindent\rule{\textwidth}{1pt}
  \vspace{-0.6ex}

  \scalebox{0.70}{
    {
  \tikzset{state/.style={rectangle, rounded corners, align=left, font=\ttfamily, draw=black, anchor=west, align=left, inner sep=5pt}}
  \tikzset{leaf/.style={fill=gray!50!white}}
  \tikzset{output/.style={fill=gray!25!white}}
  \tikzset{sum/.style={fill=green!25!white}}
  \begin{tikzpicture}
    \matrix [%
    ampersand replacement=\&,%
    column sep=5ex,%
    row sep=4ex,%
    ]{
      \node[state, leaf] (x0) {x0};
      \& \node[state] (x0-d) {f0 = sin(x0)};
      \&
      \& \node[state, output] (f0-d) {f0\_r = replicate(f0)};
      \\
      \& \node[state] (df-d) {df = cos(x0)};
      \& \node[state] (d2f-d) {d2f= -f0};
      \\
      \node[state, leaf] (x1-d) {x1\_r};
      \&
      \& \node[state, sum] (temp1-d) {temp1 = einsum(``...,r...,r...\\->...'', d2f, x1\_r, x1\_r)};
      \& \node[state, output] (f1-d) {f1\_r =
        einsum(``...,r...\\->r...'', df, x1\_r)};
      \\
      \node[state, leaf] (x2-d) {x2\_r};
      \& \node[state, sum] (x2) {x2 = sum(x2\_r, dim=0)};
      \& \node[state] (temp2-d) {temp2 = einsum(``...,...\\->...'', df, x2)};
      \& \node[state, output] (f2) {f2 = temp1 + temp2};
      \\
    };

    \tikzset{arrow/.style={-Stealth}}
    \path
    (x0) edge [arrow] (x0-d)
    (x0) edge [arrow] (df-d)
    (x0-d) edge [arrow] (f0-d)
    (df-d) edge [arrow] (temp2-d)
    (x0-d) edge [arrow] (d2f-d)
    (df-d) edge [arrow, out=-8, in=169] (f1-d)
    (x1-d) edge [arrow, out=10, in=167] (f1-d)
    (d2f-d) edge [arrow] (temp1-d)
    (x2-d) edge [arrow] (x2)
    (x2) edge [arrow] (temp2-d)
    (x1-d) edge [arrow] (temp1-d)
    (temp1-d) edge [arrow] (f2)
    (temp2-d) edge [arrow] (f2)
    ;
  \end{tikzpicture}
}
  }

  \caption{\textbf{Step-by-step illustration of propagating \texttt{sum} nodes up a computation graph.}}
  \label{fig:pull-sum-simplification}
\end{figure}

\clearpage
\section{Exploiting Linearity to Collapse Taylor Mode}
 Here, we illustrate the idea behind propagating $R$ $K$-jets through $\vf = \vg \circ \vh$ with input jets $\smash{\left(J^K\vx\right)_r(t)} = \smash{\sum_{k=0}^K \frac{t^k}{k!} \vx_{k,r}}$.
The Taylor mode scheme results from inserting \cref{eq:sum-k-directional} into \cref{eq:taylor-mode-composition}:
\begin{align}\label{eq:sum-taylor-mode-naive}
  \begin{split}
      \begin{pmatrix*}
        \vx_0
        \\
       \{\vx_{1,r}\}
        \\
        \{\vx_{2,r}\}
        \\
        \vdots
        \\
        \{\vx_{K, r}\}
      \end{pmatrix*}
      &\overset{\text{(\ref{eq:faa-di-bruno})}}{\to}
        \begin{pmatrix*}[l]
          \vh_0 =  \vh(\vx_0)
          \\
          \{\vh_{1,r}\} =
          \left\{
          \left<
          \partial \vh(\vx_0),
          \vx_{1, r}
          \right>
          \right\}
          \\
          \{\vh_{2, r}\} =
          \left\{
          \left<
          \partial^2 \vh(\vx_0),
          \vx_{1, r} \otimes \vx_{1, r}
          \right>
          +
          \left<
          \partial \vh(\vx_0),
          \vx_{2, r}
          \right>
          \right\}
          \\
          \vdots
          \\
          \{\vh_{K, r}\} =
          \left\{
          \displaystyle \sum_{
          \sigma \in \partitioning(K)
          }
          \nu(\sigma) \left<
          \partial^{|\sigma|} \vh(\vx_0),
          \tensorprod{s \in \sigma} \vx_{s, r}
          \right>
          \right\}
        \end{pmatrix*}
        \\
        &\overset{\text{(\ref{eq:faa-di-bruno})}}{\to}
        \begin{pmatrix*}[l]
          \vg_0 =  \vg(\vh_0)
          \\
          \{\vg_{1,r}\} =
          \left\{
          \left<
          \partial \vg(\vh_0),
          \vh_{1, r}
          \right>
          \right\}
          \\
          \{\vg_{2,r}\} =
          \left\{
          \left<
          \partial^2 \vg(\vh_0),
          \vh_{1, r} \otimes \vh_{1, r}\right>
          +
          \left< \partial \vg(\vh_0),
          \vh_{2, r}
          \right>
          \right\}
          \\
          \vdots
          \\
          \{\vg_{K,r}\} =
          \left\{
          \displaystyle\sum_{
          \sigma \in \partitioning(K)
          }
          \nu(\sigma) \left<
          \partial^{|\sigma|} \vg(\vh_0),
          \tensorprod{s \in \sigma} \vh_{s, r}
          \right>
          \right\}
        \end{pmatrix*}
        \\
        &\overset{\text{(\ref{eq:faa-di-bruno})}}{=}
        \begin{pmatrix*}[l]
          \vf_0 =  \vf(\vx_0)
          \\
          \{\vf_{1,r}\} = \left\{
          \left<
          \partial \vf(\vx_0),
          \vx_{1, r}
          \right>
          \right\}
          \\
          \{\vf_{2,r}\} = \left\{
          \left<
          \partial^2 \vf(\vx_0),
          \vx_{1,r} \otimes \vx_{1, r}
          \right>
          +
          \left< \partial \vf(\vx_0),
          \vx_{2, r}
          \right>
          \right\}
          \\
          \vdots
          \\
          \{\vf_{K, r}\} =
          \left\{
          \displaystyle\sum_{
          \sigma \in \partitioning(K)
          }
          \nu(\sigma) \left<
          \partial^{|\sigma|} \vf(\vx_0),
          \tensorprod{s \in \sigma} \vx_{s, r}
          \right>
          \right\}
        \end{pmatrix*}
        \\
        &\overset{\text{slice}}{\to} \{ \vg_{K,r} \}
        \\
        &\overset{\text{sum}}{\to} \sum_{r=1}^R \vg_{K,r}
          \overset{\{\vx_{1,r} = \vv_r, \vx_{2,r} =  \ldots = \vx_{K, r} = \vzero \}}{=}
          \sum_{r=1}^R \left<
          \partial^K \vf(\vx_0),
          \otimes_{k=1}^K \vv_r
          \right>
    \end{split}
\end{align}
Leveraging linearity in certain terms (in green) of the highest coefficient, as explained in \cref{eq:faa-di-bruno-expanded}, instead leads to
\begin{align}\label{eq:sum-taylor-mode-efficient}
      \begin{pmatrix*}
        \vx_0
        \\
       \{\vx_{1,r}\}
        \\
        \{\vx_{2,r}\}
        \\
        \vdots
        \\
        \textcolor{\colorcTM}{\displaystyle\sum_{r = 1}^R\vx_{K, r}}
      \end{pmatrix*}
  &\overset{\text{(\ref{eq:faa-di-bruno})}}{\to}        \begin{pmatrix*}[l]
          \vh_0 &=  \vh(\vx_0)
          \\
          \{\vh_{1,r}\} &=
          \left\{
          \left<
          \partial \vh(\vx_0),
          \vx_{1, r}
          \right>
          \right\}
          \\
          \{\vh_{2, r}\} &=
          \left\{
          \left<
          \partial^2 \vh(\vx_0),
          \vx_{1, r} \otimes \vx_{1, r}
          \right>
          +
          \left<
          \partial \vh(\vx_0),
          \vx_{2, r}
          \right>
          \right\}
          \\
          \vdots
          \\
          \textcolor{\colorcTM}{ \displaystyle \sum_{r = 1}^R \vh_{K, r}} &=
          \displaystyle \sum_{r=1}^R \sum_{
          \sigma \in \partitioning(K) \setminus \{\tilde{\sigma}\}
          }
          \nu(\sigma) \left<
          \partial^{|\sigma|} \vh(\vx_0),
          \tensorprod{s \in \sigma} \vx_{s, r}
          \right>
          \\
          &+
          \left<
          \partial \vh(\vx_0),
          \textcolor{\colorcTM}{
          \displaystyle\sum_{r=1}^R \vx_{K, r}
          }
          \right>
        \end{pmatrix*}      \nonumber \\
       &\overset{\text{(\ref{eq:faa-di-bruno})}}{\to}
        \begin{pmatrix*}[l]
          \vg_0 &=  \vg(\vh_0)
          \\
          \{\vg_{1,r}\} &=
          \left\{
          \left<
          \partial \vg(\vh_0),
          \vh_{1, r}
          \right>
          \right\}
          \\
          \{\vg_{2,r}\} &=
          \left\{
          \left<
          \partial^2 \vg(\vh_0),
          \vh_{1, r} \otimes \vh_{1, r}\right>
          +
          \left< \partial \vg(\vh_0),
          \vh_{2, r}
          \right>
          \right\}
          \\
          \vdots
          \\
          \textcolor{\colorcTM}{\displaystyle\sum_{r=1}^R\vg_{K,r}} &=
          \displaystyle \sum_{r=1}^R \sum_{
          \sigma \in \partitioning(K) \setminus \{\tilde{\sigma}\}
          }
          \nu(\sigma) \left<
          \partial^{|\sigma|} \vg(\vh_0),
          \tensorprod{s \in \sigma} \vh_{s, r}
          \right>
          \\
          &+
          \left<
          \partial \vg(\vh_0),
          \textcolor{\colorcTM}{\displaystyle \sum_{r=1}^R\vh_{K, r}}
          \right>
        \end{pmatrix*}
         \end{align}
\begin{align*}%
        &\overset{\text{(\ref{eq:faa-di-bruno})}}{=}
        \begin{pmatrix*}[l]
          \vf_0 &=  \vf(\vx_0)
          \\
          \{\vf_{1,r}\} &= \left\{
          \left<
          \partial \vf(\vx_0),
          \vx_{1, r}
          \right>
          \right\}
          \\
          \{\vf_{2,r}\} &= \left\{
          \left<
          \partial^2 \vf(\vx_0),
          \vx_{1, r} \otimes \vx_{1, r}
          \right>
          +
          \left< \partial \vf(\vx_0),
          \vx_{2, r}
          \right>
          \right\}
          \\
          \vdots
          \\
          \textcolor{\colorcTM}{\displaystyle \sum_{r=1}^R \vf_{K, r}} &=
          \displaystyle \sum_{r=1}^R\sum_{
          \sigma \in \partitioning(K) \setminus \{ \tilde{\sigma} \}
          }
          \nu(\sigma) \left<
          \partial^{|\sigma|} \vf(\vx_0),
          \tensorprod{s \in \sigma} \vx_{s, r}
          \right>
          \\
          &+
          \left<
          \partial \vf(\vx_0),
          \textcolor{\colorcTM}{\displaystyle \sum_{r=1}^R\vx_{K, r}}
          \right>
        \end{pmatrix*}
        \\
        &\overset{\text{slice}}{\to} \textcolor{\colorcTM}{\sum_{r=1}^R \vg_{K,r}}
          \overset{\{\vx_{1,r} = \vv_r, \vx_{2,r} =  \ldots = \vx_{K, r} = \vzero \}}{=}
          \sum_{r=1}^R \left<
          \partial^K \vf(\vx_0),
          \otimes_{k=1}^K \vv_r
          \right>
\end{align*}

\subsection{Second-order Operators: Laplacian}
Here, we show details about the propagation schemes of standard Taylor mode and collapsed Taylor mode for the computation of the Laplacian of $\vf$. We consider the decomposition $\vf = \vg \circ \vh$.

\paragraph{Standard Taylor mode.}
Using standard Taylor mode (\cref{eq:sum-taylor-mode-naive}) to compute the Laplacian yields
{
\setlength{\abovedisplayskip}{0pt}
\setlength{\belowdisplayskip}{0pt}
\begin{align}\label{eq:laplacian-naive}
  \begin{split}
    \begin{pmatrix*}
      \vx_0
      \\
      \{\vx_{1,d} \}
      \\
      \{\vx_{2,d} \}
    \end{pmatrix*}
    &\overset{\text{(\ref{eq:faa-di-bruno})}}{\to}
      \begin{pmatrix*}[l]
        \vh_0 =  \vh(\vx_0)
        \\
        \{\vh_{1,d}\} = \{
        \left<
        \partial \vh(\vx_0),
        \vx_{1,d}
        \right>
        \}
        \\
        \{\vh_{2,d}\} =
        \{
        \left<
        \partial^2 \vh(\vx_0),
        \vx_{1,d} \otimes \vx_{1,d}
        \right>
        +
        \left<
        \partial \vh(\vx_0),
        \vx_{2,d}
        \right>
        \}
      \end{pmatrix*}
    \\
    &\overset{\text{(\ref{eq:faa-di-bruno})}}{\to}
      \begin{pmatrix*}[l]
        \vg_0 =  \vg(\vh_0)
        \\
        \{\vg_{1,d}\} = \{
        \left<\partial \vg(\vh_0),
        \vh_{1,d}
        \right>
        \}
        \\
        \{\vg_{2,d}\} = \{
        \left<
        \partial^2 \vg(\vh_0),
        \vh_{1,d} \otimes \vh_{1,d}
        \right>
        +
        \left<
        \partial \vg(\vh_0),
        \vh_{2,d}
        \right>
        \}
      \end{pmatrix*}
      \\
      &\overset{\text{(\ref{eq:faa-di-bruno})}}{=}
      \begin{pmatrix*}[l]
        \vf_0 =  \vf(\vx_0)
        \\
        \{\vf_{1,d} \} = \{
        \left<
        \partial \vf(\vx_0),
        \vx_{1,d}
        \right>
        \}
        \\
        \{ \vf_{2,d} \} = \{
        \left<
        \partial^2 \vf(\vx_0),
        \vx_{1,d} \otimes \vx_{1,d}
        \right>
        +
        \left<
        \partial \vf(\vx_0),
        \vx_{2,d}
        \right>
        \}
      \end{pmatrix*}
    \\
    &\overset{\text{slice}}{\to} \{ \vg_{2,d} \}
    \\
    &\overset{\text{sum}}{\to} \sum_{d=1}^D \{ \vg_{2,d} \}
      \overset{\{\vx_{1,d} = \ve_d, \vx_{2,d} = \vzero\}}{=} \Delta \vf(\vx_0).
  \end{split}
\end{align}
}

\paragraph{Collapsed Taylor Mode AD}
Using our proposed collapsed Taylor mode, we get
{
\setlength{\abovedisplayskip}{2pt}
\setlength{\belowdisplayskip}{2pt}
\begin{align}\label{eq:laplacian-efficient}
  \begin{split}
    \begin{pmatrix*}
      \vx_0
      \\
      \{\vx_{1,d} \}
      \\
      \textcolor{\colorcTM}{\displaystyle\sum_{d=1}^D \vx_{2,d}}
    \end{pmatrix*}
    &\overset{\text{(\ref{eq:taylor-mode-scalar})}}{\to}
      \begin{pmatrix*}[l]
        \vh_0 =  \vh(\vx_0)
        \\
        \{\vh_{1,d}\} = \{
        \left<
        \partial \vh(\vx_0),
        \vx_{1,d}
        \right>
        \}
        \\
        \textcolor{\colorcTM}{\displaystyle\sum_{d=1}^D \vh_{2,d}} = \displaystyle\sum_{d=1}^D
        \left< \partial^2 \vh(\vx_0),
        \vx_{1,d} \otimes \vx_{1,d}
        \right>
        +
        \left<
        \partial \vh(\vx_0),
        \textcolor{\colorcTM}{\displaystyle\sum_{d=1}^D\vx_{2,d}}
        \right>
      \end{pmatrix*}
    \\
    &\overset{\text{(\ref{eq:taylor-mode-scalar})}}{\to}
      \begin{pmatrix*}[l]
        \vg_0 =  \vg(\vh_0)
        \\
        \{\vg_{1,d}\} = \{
        \left<
        \partial \vg(\vh_0),
        \vh_{1,d}
        \right>
        \}
        \\
        \textcolor{\colorcTM}{\displaystyle\sum_{d=1}^D\vg_{2,d}}
        =
        \displaystyle\sum_{d=1}^D
        \left<
        \partial^2 \vg(\vh_0),
        \vh_{1,d} \otimes \vh_{1,d}
        \right>
        +
        \left<
        \partial \vg(\vh_0),
        \textcolor{\colorcTM}{\displaystyle\sum_{d=1}^D\vh_{2,d}}
        \right>
      \end{pmatrix*}
      \\
      &\overset{\text{(\ref{eq:taylor-mode-scalar})}}{=}
      \begin{pmatrix*}[l]
        \vf_0 =  \vf(\vx_0)
        \\
        \{\vf_{1,d} \} = \{
        \left<
        \partial \vf(\vx_0),
        \vx_{1,d}
        \right>
        \}
        \\
        \textcolor{\colorcTM}{\displaystyle\sum_{d=1}^D \vf_{2,d}}
        =
        \displaystyle\sum_{d=1}^D
        \left<
        \partial^2 \vf(\vx_0),
        \vx_{1,d} \otimes \vx_{1,d}
        \right>
        +
        \left<
        \partial \vf(\vx_0),
        \textcolor{\colorcTM}{\displaystyle\sum_{i=1}^D\vx_{2,d}}
        \right>
      \end{pmatrix*}
    \\
    &\overset{\text{slice}}{\to} \textcolor{\colorcTM}{\sum_{d=1}^D \{ \vg_{2,d} \}}
      \overset{\{(\vx_{1,d} = \ve_d, \vx_{2,d} = \vzero)\}}{=}
      \Delta \vf(\vx_0)
  \end{split}
\end{align}
}

\section{(Collapsed) Taylor Mode for Arbitrary Mixed Partial Derivatives}\label{sec:appendix_ttc}
Here, we introduce the notation of \cref{eq:ttc_general}.
The right side of the formula sums over the set of all $\vj \in \sN^I$ such that $\lVert \vj \rVert_1 := \sum_i [\vj]_i = K$.
For example, if $I=2$ and $\lVert \vj \rVert_1 = 4$, this set is $\left\{ (4,0), (0, 4), (3, 1), (1, 3), (2, 2)\right\}$.

The coefficient $\gamma_{\vi, \vj}$ is defined as
\begin{equation}
  \label{eq:ttc_coeff}
  \gamma_{\vi, \vj} := \sum_{0 < \vm \leq \vi} (-1)^{\lVert \vi - \vm \rVert_1}
  \left(
    \begin{matrix}
      \vi \\
      \vm
    \end{matrix}
  \right)
  \left(
    \begin{matrix}
      \lVert \vi \rVert_1 \frac{\vm}{\lVert \vm \rVert_1} \\
      \vj
    \end{matrix}
  \right)
  \left(
    \frac{\lVert \vm \rVert_1}{\lVert \vi \rVert_1}
  \right)^{\lVert \vi \rVert_1}\,.
\end{equation}
The summation ranges over the set $\left\{ \vm \in \sN^I \mid [\vm]_1 \leq [\vi]_1, \dots, [\vm]_I \leq [\vi]_I, \lVert \vm \rVert_1 > 0 \right\}$.
Furthermore, we utilize the generalized binomial coefficient
\begin{equation*}
  \left(
    \begin{matrix}
      a \\
      b
    \end{matrix}
  \right) := \prod_{l=0}^{b-1} \frac{a - l}{b - l}
\end{equation*}
to allow the computation for all $a \in \sR$ and $b \in \sN$, which is defined to be $1$ if $b=0$.
The generalized binomial coefficient of vectors is the component-wise product of generalized binomial coefficients:
\begin{equation*}
  \left(
    \begin{matrix}
      \va \\
      \vb
    \end{matrix}
  \right)
  :=
  \prod_{i=1}^I
  \left(
    \begin{matrix}
      [\va]_i \\
      [\vb]_i
    \end{matrix}
  \right)\,.
\end{equation*}
This notation also includes cases where the vector $\va$ has components of $\sR$.

\paragraph{Example computation.}
Let us compute the coefficient $\gamma_{(2, 2),(3, 1)}$, used by the biharmonic operator:
\begin{align*}
  \gamma_{(2, 2),(3, 1)} = \,\,\,\,\,\mathclap{\sum_{
  \substack{
  \vm \in \sN, ||\vm||_1 > 0
  \\
  [\vm]_1 \leq 2, [\vm]_2 \leq 2
  }
  }
  }\,\,\,\,\,\,\,
  (-1)^{ 2 - [\vm]_1 + 2 - [\vm]_2 }
  \left(
  \begin{matrix}
    2 \\
    [\vm]_1
  \end{matrix}
  \right)
  \left(
  \begin{matrix}
    2 \\
    [\vm]_2
  \end{matrix}
  \right)
  \left(
  \begin{matrix}
    4 \frac{[\vm]_1}{\lVert \vm \rVert_1} \\
    3
  \end{matrix}
  \right)
  \left(
  \begin{matrix}
    4 \frac{[\vm]_2}{\lVert \vm \rVert_1} \\
    1
  \end{matrix}
  \right)
  \left(
  \frac{\lVert \vm \rVert_1}{4}
  \right)^4\,.
\end{align*}
We have $\vm \in \{(1, 0), (2, 0), (1, 1), (2, 1), (2, 2), (1, 2), (0, 1), (0, 2)\}$, which results in the terms
\begin{align*}
  =\phantom{+}&(-1)^{ 2 - 1 + 2 - 0}
                \left(
                \begin{matrix}
                  2 \\
                  1
                \end{matrix}
                \right)
                \left(
                \begin{matrix}
                  2 \\
                  0
                \end{matrix}
                \right)
                \left(
                \begin{matrix}
                  4 \frac{1}{1} \\
                  3
                \end{matrix}
                \right)
                \left(
                \begin{matrix}
                  4 \frac{0}{1} \\
                  1
                \end{matrix}
                \right)
                \left(
                \frac{1}{4}
                \right)^4
  \\
  +&
     (-1)^{ 2 - 2 + 2 - 0 }
     \left(
     \begin{matrix}
       2 \\
       2
     \end{matrix}
     \right)
     \left(
     \begin{matrix}
       2 \\
       0
     \end{matrix}
     \right)
     \left(
     \begin{matrix}
       4 \frac{2}{2} \\
       3
     \end{matrix}
     \right)
     \left(
     \begin{matrix}
       4 \frac{0}{2} \\
       1
     \end{matrix}
     \right)
     \left(
     \frac{2}{4}
     \right)^4
  \\
  +&
     (-1)^{ 2 - 1 + 2 - 1 }
     \left(
     \begin{matrix}
       2 \\
       1
     \end{matrix}
     \right)
     \left(
     \begin{matrix}
       2 \\
       1
     \end{matrix}
     \right)
     \left(
     \begin{matrix}
       4 \frac{1}{2} \\
       3
     \end{matrix}
     \right)
     \left(
     \begin{matrix}
       4 \frac{1}{2} \\
       1
     \end{matrix}
     \right)
     \left(
     \frac{2}{4}
     \right)^4
  \\
  +&
     (-1)^{ 2 - 2 + 2 - 1 }
     \left(
     \begin{matrix}
       2 \\
       2
     \end{matrix}
     \right)
     \left(
     \begin{matrix}
       2 \\
       1
     \end{matrix}
     \right)
     \left(
     \begin{matrix}
       4 \frac{2}{3} \\
       3
     \end{matrix}
     \right)
     \left(
     \begin{matrix}
       4 \frac{1}{3} \\
       1
     \end{matrix}
     \right)
     \left(
     \frac{3}{4}
     \right)^4
  \\
  +&
     (-1)^{ 2 - 2 + 2 - 2}
     \left(
     \begin{matrix}
       2 \\
       2
     \end{matrix}
     \right)
     \left(
     \begin{matrix}
       2 \\
       2
     \end{matrix}
     \right)
     \left(
     \begin{matrix}
       4 \frac{2}{4} \\
       3
     \end{matrix}
     \right)
     \left(
     \begin{matrix}
       4 \frac{2}{4} \\
       1
     \end{matrix}
     \right)
     \left(
     \frac{4}{4}
     \right)^4
  \\
  +&
     (-1)^{ 2 - 1 + 2 - 2}
     \left(
     \begin{matrix}
       2 \\
       1
     \end{matrix}
     \right)
     \left(
     \begin{matrix}
       2 \\
       2
     \end{matrix}
     \right)
     \left(
     \begin{matrix}
       4 \frac{1}{3} \\
       3
     \end{matrix}
     \right)
     \left(
     \begin{matrix}
       4 \frac{2}{3} \\
       1
     \end{matrix}
     \right)
     \left(
     \frac{3}{4}
     \right)^4
  \\
  +&
     (-1)^{ 2 - 0 + 2 - 1 }
     \left(
     \begin{matrix}
       2 \\
       0
     \end{matrix}
     \right)
     \left(
     \begin{matrix}
       2 \\
       1
     \end{matrix}
     \right)
     \left(
     \begin{matrix}
       4 \frac{0}{1} \\
       3
     \end{matrix}
     \right)
     \left(
     \begin{matrix}
       4 \frac{1}{1} \\
       1
     \end{matrix}
     \right)
     \left(
     \frac{1}{4}
     \right)^4
  \\
  +&
     (-1)^{ 2 - 0 + 2 - 2 }
     \left(
     \begin{matrix}
       2 \\
       0
     \end{matrix}
     \right)
     \left(
     \begin{matrix}
       2 \\
       2
     \end{matrix}
     \right)
     \left(
     \begin{matrix}
       4 \frac{0}{2} \\
       3
     \end{matrix}
     \right)
     \left(
     \begin{matrix}
       4 \frac{2}{2} \\
       1
     \end{matrix}
     \right)
     \left(
     \frac{2}{4}
     \right)^4\,.
\end{align*}
The next step is to evaluate the binomial coefficients:
\begin{align*}
  =\phantom{+}&(-1) \cdot 2 \cdot 1 \cdot 4 \cdot 0 \cdot
                \left(
                \frac{1}{4}
                \right)^4
  \\
  +&
     1 \cdot 1 \cdot 4 \cdot 0 \cdot
     \left(
     \frac{2}{4}
     \right)^4
  \\
  +&
     2 \cdot 2 \cdot 0 \cdot 2 \cdot
     \left(
     \frac{2}{4}
     \right)^4
  \\
  -&
     1 \cdot 2 \cdot \frac{8}{9}\frac{5}{6}\frac{2}{3} \cdot \frac{4}{3} \cdot
     \left(
     \frac{3}{4}
     \right)^4
  \\
  +&
     1 \cdot 1 \cdot 0 \cdot 2 \cdot
     \left(
     \frac{4}{4}
     \right)^4
  \\
  -& 2 \cdot 1
     \cdot \frac{4}{9}\frac{1}{6} \frac{-2}{3} \cdot \frac{8}{3} \cdot
     \left(
     \frac{3}{4}
     \right)^4
  \\
  -& 1 \cdot 2 \cdot 0 \cdot 4 \cdot
     \left(
     \frac{1}{4}
     \right)^4
  \\
  +&
     1 \cdot 1 \cdot 1 \cdot 0 \cdot 4 \cdot
     \left(
     \frac{2}{4}
     \right)^4\,.
\end{align*}
After simplification, the final result is
\begin{align*}
  \gamma_{(2, 2),(3, 1)} = &(-1) \cdot 2 \cdot \frac{8}{9}\frac{5}{6}\frac{2}{3} \cdot \frac{4}{3} \cdot
                             \left(
                             \frac{3}{4}
                             \right)^4
                             - 2 \left(\frac{4}{9} \cdot \frac{1}{6} \cdot \frac{-2}{3}\right) \cdot \frac{8}{3} \cdot
                             \left(
                             \frac{3}{4}
                             \right)^4
  \\
                           &=  \frac{-640}{486}\frac{81}{256} + \frac{128}{486}\frac{81}{256} = \frac{-5}{12}+\frac{1}{12} = -\frac{1}{3}.
\end{align*}

\subsection{Applied to the Biharmonic Operator}\label{sec:appendix-biharmonic-details}

To compute \cref{eq:biharm} with \cref{eq:ttc_general}, we first select $K = 4, I = 2, D_1 = D_2 = D, \vi = (2, 2), \vv_{d_1} = \ve_{d_1}$ and $\ve_{d_2} = \ve_{d_2}$. Then we insert these parameters into the general equation \cref{eq:ttc_general} and get
\begin{equation} \label{eq:ttc_for_biharm}
  \begin{aligned}
    \Delta^2 \vf(\vx_0)
    &=
      \sum_{\vj \in \sN^2, \lVert \vj \rVert_1 = 4}
      \gamma_{(2, 2), \vj}
      \frac{1}{4!}
      \sum_{d_1=1}^D \sum_{d_2=1}^D
      \left<
      \partial^4 \vf(\vx_0),
      \left(\ve_{d_1} [\vj]_1 + \ve_{d_2} [\vj]_2\right)^{\otimes 4}
      \right>
    \\
    &=
      \frac{1}{24}
      \Big(
      \gamma_{(2, 2), (4, 0)}
      \sum_{d_1=1}^D \sum_{d_2=1}^D
      \left<
      \partial^4 \vf(\vx_0),
      \left(4 \ve_{d_1}\right)^{\otimes 4}
      \right>
    \\
    &+
      \gamma_{(2, 2), (0, 4)}
      \sum_{d_1=1}^D \sum_{d_2=1}^D
      \left<
      \partial^4 \vf(\vx_0),
      \left(4 \ve_{d_2} \right)^{\otimes 4}
      \right>
    \\
    &+
      \gamma_{(2, 2), (3, 1)}
      \sum_{d_1=1}^D \sum_{d_2=1}^D
      \left<
      \partial^4 \vf(\vx_0),
      \left(3 \ve_{d_1} + \ve_{d_2}\right)^{\otimes 4}
      \right>
    \\
    &+
      \gamma_{(2, 2), (1, 3)}
      \sum_{d_1=1}^D \sum_{d_2=1}^D
      \left<
      \partial^4 \vf(\vx_0),
      \left(
      \ve_{d_1} + 3 \ve_{d_2}
      \right)^{\otimes 4}
      \right>
    \\
    &+
      \gamma_{(2, 2), (2, 2)}
      \sum_{d_1=1}^D \sum_{d_2=1}^D
      \left<
      \partial^4 \vf(\vx_0),
      \left( 2 \ve_{d_1} +  2\ve_{d_2}\right)^{\otimes 4}
      \right>
      \Big).
  \end{aligned}
\end{equation}
Now, exploit the symmetry of the coefficients $\gamma_{(2, 2), (4, 0)} = \gamma_{(2, 2), (0, 4)}$ and $\gamma_{(2, 2), (3, 1)} = \gamma_{(2, 2), (1, 3)}$ and the corresponding tensor basis expansion:
\begin{equation} \label{eq:ttc_for_biharm_2}
  \begin{aligned}
    &=\frac{1}{24}
      \Big(
      2D\gamma_{(2, 2),(4, 0)}
      \sum_{d_1=1}^D
      \left<
      \partial^4 \vf(\vx_0),
      \left(4 \ve_{d_1}\right)^{\otimes 4}
      \right>
    \\
    &+
      2 \gamma_{(2, 2), (3, 1)}
      \sum_{d_1=1}^D \sum_{d_2=1}^D
      \left<
      \partial^4 \vf(\vx_0),
      \left(
      3 \ve_{d_1} + \ve_{d_2}
      \right)^{\otimes 4}
      \right>
    \\
    &+
      \gamma_{(2, 2), (2, 2)}
      \sum_{d_1=1}^D \sum_{d_2=1}^D
      \left<
      \partial^4 \vf(\vx_0),
      \left( 2 \ve_{d_1} +  2\ve_{d_2}\right)^{\otimes 4}
      \right>
      \Big)\,.
  \end{aligned}
\end{equation}
Since the first sum captures all diagonal directions $\ve_{d_1} = \ve_{d_2}$, we extract this from the second and third sums to further reduce the computational effort:
\begin{equation} \label{eq:ttc_for_biharm3}
  \begin{aligned}
    &=\frac{1}{24}
      \Big(
      \left(
      2D\gamma_{(2, 2), (4, 0)} + 2 \gamma_{(2, 2), (3, 1)} + \gamma_{(2, 2),(2, 2)}
      \right)
      \sum_{d_1=1}^D
      \left<
      \partial^4 \vf(\vx_0),
      \left( 4 \ve_{d_1} \right)^{\otimes 4}
      \right>
    \\
    &+
      2 \gamma_{(2, 2),(3, 1)}
      \sum_{d_1=1}^D \sum_{\underset{d_2 \neq d_1}{d_2=1}}^D
      \left<
      \partial^4 \vf(\vx_0),
      \left(
      3 \ve_{d_1} + \ve_{d_2}
      \right)^{\otimes 4}
      \right>
    \\
    &+
      \gamma_{(2, 2), (2, 2)}
      \sum_{d_1=1}^D \sum_{\underset{d_2 \neq d_1}{d_2 = 1}}^D
      \left<
      \partial^4 \vf(\vx_0),
      \left( 2 \ve_{d_1} +  2\ve_{d_2}\right)^{\otimes 4}
      \right>
      \Big).
  \end{aligned}
\end{equation}
Exploiting further symmetries in the last term's summation, we obtain
\begin{equation} \label{eq:ttc_for_biharm_final}
  \begin{aligned}
    \Delta^2 \vf(\vx_0) &=
                          \frac{1}{24}
                          \Big(
                          \left(
                          2D\gamma_{(2, 2), (4, 0)} + 2 \gamma_{(2, 2), (3, 1)} + \gamma_{(2, 2),(2, 2)}
                          \right)
                          \sum_{d_1=1}^D
                          \left<
                          \partial^4 \vf(\vx_0),
                          \left(4 \ve_{d_1}\right)^{\otimes 4}
                          \right>
    \\
                        &+
                          2 \gamma_{(2, 2), (3, 1)}
                          \sum_{d_1=1}^D\sum_{\underset{d_2 \neq d_1}{d_2=1}}^D\!\!\!
                          \left<
                          \partial^4 \vf(\vx_0),
                          \left(
                          3 \ve_{d_1}+\ve_{d_2}
                          \right)^{\otimes 4}
                          \right>
    \\
                        & +
                          2 \gamma_{(2, 2), (2, 2)}
                          \sum_{d_1=1}^{D - 1} \sum_{d_2 = d_1 + 1}^D
                          \left<
                          \partial^4 \vf(\vx_0),
                          \left( 2 \ve_{d_1}+2\ve_{d_2}\right)^{ \otimes 4 }
                          \right>
                          \Big)\,.
  \end{aligned}
\end{equation}

\subsection{Pedagogical Approach for the Biharmonic Operator with 6-jets}\label{sec:appendix_ttc_other_methods}

A different approach to compute arbitrary-mixed derivatives was proposed in \cite{shi2024stochastic}. This approach relies, for the biharmonic operator, on the hand-selection of certain $6$-jets to extract the required derivatives. The degree and directions for the jets are obtained by considering the Faà di Bruno formula for the 6-th coefficient $\vf_6$ (see \cref{sec:faa-di-bruno-cheatsheet}). Selecting coefficients of the input $6$-jet to $\vx_1 = \ve_{d_1}, \vx_2 = \ve_{d_2}$ and  $\vx_3 = \vx_4 = \vx_5 = \vx_6 = \vzero$ leads us to
\begin{align}\label{eq:felix-biharmonic-jet1}
  \begin{split}
    \vf_6
    &=
      \left<
      \partial^{6} \vf(\vx_0),
      \ve_{d_1}^{\otimes 6}
      \right>
      +
      15
      \left<
      \partial^5 \vf(\vx_0),
      \ve_{d_1}^{\otimes 4} \otimes \ve_{d_2}
      \right>
    \\
    &  \quad{}  +
      {\color{blue}
      45
      \left<
      \partial^4 \vf(\vx_0),
      \ve_{d_1}^{\otimes 2} \otimes \ve_{d_2}^{\otimes 2}
      \right>
      }
      +
      15
      \left<
      \partial^3 \vf(\vx_0),
      \ve_{d_2}^{\otimes 3}
      \right>.
  \end{split}
\end{align}
Notice the \textcolor{blue}{blue term}, which has the same structure as the summands we want to compute for the biharmonic operator. Therefore, a first $6$-jet is computed as explained above. To cancel out the unwanted terms, we evaluate another $6$-jet with the same input except $\vx_2 = -\ve_{d_2}$ and adding the $6$-th coefficient of this jet to \cref{eq:felix-biharmonic-jet1} gives
\begin{align}\label{eq:felix-biharmonic-jet2}
  2 \left<
  \partial^{6} \vf(\vx_0),
  \ve_{d_1}^{\otimes 6}
  \right>
  +
  {\color{blue}
  90
  \left<
  \partial^4 \vf(\vx_0),
  \ve_{d_1}^{\otimes 2} \otimes \ve_{d_2}^{\otimes 2}
  \right>.
  }
\end{align}
Finally, a third $6$-jet is computed with $\vx_2 = \vzero$. The $6$-th coefficient of this jet contains only
\begin{align}\label{eq:felix-biharmonic-jet3}
  \left<
  \partial^{6} \vf(\vx_0),
  \vx_1^{\otimes 6}
  \right>.
\end{align}
We obtain
\begin{equation}
  {\color{blue}
    90
    \left<
      \partial^4 \vf(\vx_0),
      \vx_1^{\otimes 2} \otimes \vx_2^{\otimes 2}
    \right>
  }
\end{equation}
by subtracting twice of the $6$-th coefficient of the third jet from \cref{eq:felix-biharmonic-jet2}.

To summarize the procedure, we evaluate the 6-jet three times.
The first jet has the input $\vx_1 = \ve_{d_1}, \vx_2 = \ve_{d_2}$ and $\vx_3 = \vx_4 = \vx_5 = \vx_6 = \vzero$, the second jet has the same input jet apart from $\vx_2 = - \ve_{d_2}$, and the third 6-jet takes $\vx_2 = \vzero$.
Then we add the $6$-th coefficient of the first and the second and subtract twice of the $6$-th coefficient of the third jet.
Dividing by $90$ provides the derivative corresponding to the $d_1, d_2$ term of the biharmonic operator.

Standard Taylor mode would propagate $1 + 18D^2$ vectors through every node, if we already exploit that all jets share $\vx_0$.
our collapsed Taylor mode would pass $1 + 3 + 15D^2$ vectors through every node of the compute graph.
This is more costly compared to our approach described before.
In addition, until now, the selection of the jet degree and the input coefficients requires substantial human effort.

\subsection{Another Example: Mixed Third-order Derivatives}
As an additional example, consider computing $\sum_{i=1}^D\sum_{j=1}^D\frac{\partial^3}{\partial x_i^2 x_j}\vf(\vx)$.
This example is from \citep[][\S{}F.2]{shi2024stochastic}, which describes how to compute these 3rd-order derivatives using 7-jets.
The interpolation formula allows using multiple 3-jets instead.
We expect it to be favorable as Taylor mode scales polynomially in the derivative order.

\paragraph{Procedure.}
The goal is to compute $\sum_{i=1}^D\sum_{j=1}^D \frac{\partial^3}{\partial x_i^2 \partial x_j} \vf(\vx)$.
We proceed as follows:

\begin{enumerate}
\item Formulate the operator in our notation:
  \begin{equation*}
    \sum_{i=1}^D\sum_{j=1}^D \langle\partial^3 \vf(\vx), \ve_i^{\otimes 2} \otimes \ve_j    \rangle
  \end{equation*}

\item Compute the interpolation coefficients $\gamma_{\vp,\vq}$ for  $\vq \in \{(3, 0), (2, 1), (1, 2), (0, 3)\}$ and $\vp=(2, 1)$:
  $\gamma_{(2, 1)(0, 3)} =  -\nicefrac{8}{81}, \gamma_{(2, 1)(1, 2)} = \nicefrac{16}{27}, \gamma_{(2, 1)(2, 1)} = -\nicefrac{16}{9}, \gamma_{(2, 1)(3, 0)} = \nicefrac{32}{81}$.

\item Apply the interpolation equation (\cref{eq:ttc_general}):
  \begin{equation*}
    = \sum_{i=1}^D\sum_{j=1}^D \sum_{\vq \in \sN^2, \; \left\lVert \vq \right\rVert_1 = 3} \langle\partial^3 \vf(\vx), \left([\vq]_1 \ve_i + [\vq]_2 \ve_j   \right)^{\otimes 3} \rangle
  \end{equation*}
  Collapsed Taylor mode can directly applied to these $4D^2$ 3-jets.
  However, to exploit the full potential some further steps that leverage the structure are required.

\item Expand and manually simplify, using symmetries.
  The sums for $\gamma_{(2, 1)(3, 0)}$ and $\gamma_{(2, 1)(0, 3)}$ are similar; same for $\gamma_{(2, 1)(2, 1)}$ and $\gamma_{(2, 1)(1, 2)}$.
  We only have $2D^2$-- 3 jets:
  \begin{align*}
    &= (\gamma_{(2, 1)(3, 0)} + \gamma_{(2, 1)(0, 3)}) \sum_{i=1}^D\sum_{j=1}^D \langle\partial^3 \vf(\vx), \left(3 \ve_i \right)^{\otimes 3} \rangle
    \\
    &\phantom{=} +
      (\gamma_{(2, 1)(2, 1)} + \gamma_{(2, 1)(1, 2)}) \sum_{i=1}^D\sum_{j=1}^D \langle\partial^3 \vf(\vx), \left(2 \ve_i + \ve_j \right)^{\otimes 3} \rangle\,.
  \end{align*}

  We further observe that the first summation is independent of $j$:
  \begin{align*}
    &= (\gamma_{(2, 1)(3, 0)} + \gamma_{(2, 1)(0, 3)})D
      \sum_{i=1}^D \langle\partial^3 \vf(\vx), \left(3 \ve_i   \right)^{\otimes 3} \rangle
    \\
    &\phantom{=} +
      (\gamma_{(2, 1)(2, 1)} + \gamma_{(2, 1)(1, 2)})
      \sum_{i=1}^D\sum_{j=1}^D  \langle\partial^3 \vf(\vx), \left(2 \ve_i + \ve_j \right)^{\otimes 3} \rangle\,.
  \end{align*}
  Extracting the case $i=j$ from the last term gives our final form
  \begin{align*}
    &= ((\gamma_{(2, 1)(3, 0)}D + \gamma_{(2, 1)(0, 3)}D + \gamma_{(2, 1)(2, 1)} + \gamma_{(2, 1)(1, 2)})
      \sum_{i=1}^D \langle\partial^3 \vf(\vx), \left(3 \ve_i \right)^{\otimes 3} \rangle
    \\
    &\phantom{=}+
      (\gamma_{(2, 1)(2, 1)} + \gamma_{(2, 1)(1, 2)})
      \sum_{i=1}^D\sum_{j=1, j \neq i}^D  \langle\partial^3 \vf(\vx), \left(2 \ve_i + \ve_j \right)^{\otimes 3} \rangle\,.
  \end{align*}
  This optimized version required $D^2$ 3-jets that can be collapsed.
\end{enumerate}

\section{PyTorch Benchmark}
\label{sec:pytorch-benchmark}
\subsection{Additional Analysis and Impact of \texttt{torch.compile}}

Here, we compare the theoretically estimated performance improvements based on counting the number of forward-propagated vectors with the empirically measured performance.

\paragraph{The number of propagated vectors is a good empirical performance estimate.}
To estimate the performance ratio between standard and collapsed Taylor mode, we can use the number of additional vectors both modes propagate forward as we increase either the batch size or the number of Monte-Carlo samples.
This is a relatively simplistic proxy; \eg, it assumes that each vector adds the same computational load, which is inaccurate as vectors corresponding to higher coefficients require more work and memory (as the Faà di Bruno formula contains more terms in general).
Conversely, while incrementing the MC samples does add additional vectors that are propagated, it does not introduce additional cost to compute or store the derivatives, as they are already computed with just a single sample.
\Cref{tab:benchmark-ratios} summarizes the theoretical and empirical ratios.
We find them to align quite well, despite the overly simplistic assumptions.

\paragraph{Concrete example.}
Consider the exact Laplacian.
Adding one datum introduces {\color{tab-green}$2 + D$} versus {\color{tab-orange}$1 + 2D$} new vectors.
For $D=50$, their ratio is $\nicefrac{\color{tab-green}(2 + D)}{\color{tab-orange}(1 + 2D)} \approx 0.51$.
Empirically, we measure that adding one datum adds {\color{tab-orange}\inputMetricOnly{jet/exp/exp01_benchmark_laplacian/performance/architecture_tanh_mlp_768_768_512_512_1_device_cuda_dim_50_name_laplacian_vary_batch_size/jet_naive_best.txt}\,ms} to standard, and {\color{tab-green}\inputMetricOnly{jet/exp/exp01_benchmark_laplacian/performance/architecture_tanh_mlp_768_768_512_512_1_device_cuda_dim_50_name_laplacian_vary_batch_size/jet_simplified_best.txt}\,ms} to collapsed, Taylor mode (\cref{tab:benchmark}); the ratio of $\approx \!\!\!\inputMetricRatio{jet/exp/exp01_benchmark_laplacian/performance/architecture_tanh_mlp_768_768_512_512_1_device_cuda_dim_50_name_laplacian_vary_batch_size/jet_simplified_best.txt}{jet/exp/exp01_benchmark_laplacian/performance/architecture_tanh_mlp_768_768_512_512_1_device_cuda_dim_50_name_laplacian_vary_batch_size/jet_naive_best.txt}$ is close.

\begin{table}[!b]
  \centering
  \caption{\textbf{Comparison of theoretical and empirical performance ratios between standard and collapsed Taylor mode.}
    We list the number of additional vectors that are used when adding another data point (exact) or another Monte-Carlo sample (stochastic).
    The ratio of vectors offers a good estimate of the empirically measured performance ratio.
  }\label{tab:benchmark-ratios}
  \vspace{0.5ex}
  \begin{small}
    \def\pathToExactLaplacian{jet/exp/exp01_benchmark_laplacian/performance/architecture_tanh_mlp_768_768_512_512_1_device_cuda_dim_50_name_laplacian_vary_batch_size/}
    \def\pathToStochasticLaplacian{jet/exp/exp01_benchmark_laplacian/performance/architecture_tanh_mlp_768_768_512_512_1_batch_size_2048_device_cuda_dim_50_distribution_normal_name_laplacian_vary_num_samples}
    \def\pathToExactWeightedLaplacian{jet/exp/exp01_benchmark_laplacian/performance/architecture_tanh_mlp_768_768_512_512_1_device_cuda_dim_50_name_weighted_laplacian_vary_batch_size_rank_ratio_1_0}
    \def\pathToStochasticWeightedLaplacian{jet/exp/exp01_benchmark_laplacian/performance/architecture_tanh_mlp_768_768_512_512_1_device_cuda_dim_50_name_weighted_laplacian_vary_batch_size_rank_ratio_1_0}
    \def\pathToExactBilaplacian{jet/exp/exp01_benchmark_laplacian/performance/architecture_tanh_mlp_768_768_512_512_1_device_cuda_dim_5_name_bilaplacian_vary_batch_size}
    \def\pathToStochasticBilaplacian{jet/exp/exp01_benchmark_laplacian/performance/architecture_tanh_mlp_768_768_512_512_1_batch_size_256_device_cuda_dim_5_distribution_normal_name_bilaplacian_vary_num_samples}
    \begin{tabular}{cc|ccc}
      \toprule
      \textbf{Mode}
      & \makecell{\textbf{Add one datum} \\ \textbf{or MC sample}}
      & \makecell{\textbf{Laplacian} \\ ($D = 50$)}
      & \makecell{\textbf{Weighted Laplacian} \\ ($D=R=50$)}
      & \makecell{\textbf{Biharmonic} \\ ($D=5$)}
      \\
      \midrule
      \multirow{4}{*}{\textbf{Exact}}
      & \textcolor{tab-orange}{\# vectors (standard)}
      & $1 + 2D$
      & $1 + 2 R$
      & $6D^2 - 2D + 1$
      \\
      & \textcolor{tab-green}{\# vectors (collapsed)}
      & $2 + D$
      & $2 + R$
      & $\nicefrac{9}{2} D^2 - \nicefrac{3}{2} D + 4$
      \\
      & Theoretical ratio $\nicefrac{\color{tab-green}\#}{\color{tab-orange}\#}$
      & \num{0.51}
      & \num{0.51}
      & \num{0.77}
      \\
      & Empirical time ratio
      & \!\inputMetricRatio{\pathToExactLaplacian/jet_simplified_best.txt}{\pathToExactLaplacian/jet_naive_best.txt}
      & \!\inputMetricRatio{\pathToExactWeightedLaplacian/jet_simplified_best.txt}{\pathToExactWeightedLaplacian/jet_naive_best.txt}
      & \!\inputMetricRatio{\pathToExactBilaplacian/jet_simplified_best.txt}{\pathToExactBilaplacian/jet_naive_best.txt}
      \\
      & Empirical mem.\,ratio
      & \!\inputMetricRatio{\pathToExactLaplacian/jet_simplified_peakmem.txt}{\pathToExactLaplacian/jet_naive_peakmem.txt}
      & \!\inputMetricRatio{\pathToExactWeightedLaplacian/jet_simplified_peakmem.txt}{\pathToExactWeightedLaplacian/jet_naive_peakmem.txt}
      & \!\inputMetricRatio{\pathToExactBilaplacian/jet_simplified_peakmem.txt}{\pathToExactBilaplacian/jet_naive_peakmem.txt}
      \\
      \midrule
      \multirow{4}{*}{\textbf{Stochastic}}
      & \textcolor{tab-orange}{\# vectors (standard)}
      & $2$
      & $2$
      & $4$
      \\
      & \textcolor{tab-green}{\# vectors (collapsed)}
      & $1$
      & $1$
      & $3$
      \\
      & Theoretical ratio $\nicefrac{\color{tab-green}\#}{\color{tab-orange}\#}$
      & \num{0.5}
      & \num{0.5}
      & \num{0.75}
      \\
      & Empirical time ratio
      & \!\inputMetricRatio{\pathToStochasticLaplacian/jet_simplified_best.txt}{\pathToStochasticLaplacian/jet_naive_best.txt}
      & \!\inputMetricRatio{\pathToStochasticWeightedLaplacian/jet_simplified_best.txt}{\pathToStochasticWeightedLaplacian/jet_naive_best.txt}
      & \!\inputMetricRatio{\pathToStochasticBilaplacian/jet_simplified_best.txt}{\pathToStochasticBilaplacian/jet_naive_best.txt}
      \\
      & Empirical mem.\,ratio
      & \!\inputMetricRatio{\pathToStochasticLaplacian/jet_simplified_peakmem.txt}{\pathToStochasticLaplacian/jet_naive_peakmem.txt}
      & \!\inputMetricRatio{\pathToStochasticWeightedLaplacian/jet_simplified_peakmem.txt}{\pathToStochasticWeightedLaplacian/jet_naive_peakmem.txt}
      & \!\inputMetricRatio{\pathToStochasticBilaplacian/jet_simplified_peakmem.txt}{\pathToStochasticBilaplacian/jet_naive_peakmem.txt}
      \\
      \bottomrule
    \end{tabular}
  \end{small}
\end{table}

\paragraph{Compilation reduces memory, but not runtime.}
In \cref{tab:benchmark-compiled}, we repeat the benchmark from \cref{tab:benchmark} using \texttt{torch.compile}.
We observe that compiling can further reduce the memory footprint of all approaches for computing the Laplacian and weighted Laplacian, while the runtime remains roughly identical.
For the biharmonic operator, we observe that compilation leaves runtime and memory footprint unchanged.

\begin{table}[!b]
  \centering
  \caption{\textbf{Same as \cref{tab:benchmark}, but using \texttt{torch.compile} (\ie \cref{fig:benchmark-compiled} in numbers).}}
  \label{tab:benchmark-compiled}
  \vspace{1.5ex}
  \def\datapathLaplacianExact{jet/exp/exp01_benchmark_laplacian/performance/architecture_tanh_mlp_768_768_512_512_1_compiled_device_cuda_dim_50_name_laplacian_vary_batch_size}
  \def\datapathLaplacianStochastic{jet/exp/exp01_benchmark_laplacian/performance/architecture_tanh_mlp_768_768_512_512_1_batch_size_2048_compiled_device_cuda_dim_50_distribution_normal_name_laplacian_vary_num_samples}
  \def\datapathWeightedLaplacianExact{jet/exp/exp01_benchmark_laplacian/performance/architecture_tanh_mlp_768_768_512_512_1_compiled_device_cuda_dim_50_name_weighted_laplacian_vary_batch_size_rank_ratio_1_0}
  \def\datapathWeightedLaplacianStochastic{jet/exp/exp01_benchmark_laplacian/performance/architecture_tanh_mlp_768_768_512_512_1_batch_size_2048_compiled_device_cuda_dim_50_distribution_normal_name_weighted_laplacian_vary_num_samples_rank_ratio_1_0}
  \def\datapathBilaplacianExact{jet/exp/exp01_benchmark_laplacian/performance/architecture_tanh_mlp_768_768_512_512_1_compiled_device_cuda_dim_5_name_bilaplacian_vary_batch_size}
  \def\datapathBilaplacianStochastic{jet/exp/exp01_benchmark_laplacian/performance/architecture_tanh_mlp_768_768_512_512_1_batch_size_256_compiled_device_cuda_dim_5_distribution_normal_name_bilaplacian_vary_num_samples}
  \sisetup{%
    round-mode=figures,%
    round-precision=2,%
    detect-weight, %
    tight-spacing=true, %
  }
  \begin{small}
    \begin{tabular}{ccc|cccc}
      \toprule
      \textbf{Mode}
      & \makecell{\textbf{Per-datum or } \\ \textbf{-sample cost}}
      & \textbf{Implementation}
      & \textbf{Laplacian}
      & \makecell{\textbf{Weighted} \\ \textbf{Laplacian}}
      & \!\!\makecell{\textbf{Biharmonic via } \\ \textbf{interpolation \eqref{eq:ttc_for_biharm_final}}}\!\!
      \\
      \midrule
      \multirow{9}{*}{\textbf{Exact}}
      & \multirow{3}{*}{Time [ms]}
      & \textcolor{tab-blue}{Nested 1\textsuperscript{st}-order}
      & 
      & 
      & 
      \\
      &
      & \textcolor{tab-orange}{Standard Taylor}
      & 
      & 
      & 
      \\
      &
      & \textcolor{tab-green}{Collapsed (ours)}
      & \textbf{}
      & \textbf{}
      & \textbf{}
      \\ \cmidrule{2-6}
      & \multirow{3}{*}{\makecell{Mem.\,[MiB] \\ (differentiable)}}
      & \textcolor{tab-blue}{Nested 1\textsuperscript{st}-order}
      & 
      & 
      & 
      \\
      &
      & \textcolor{tab-orange}{Standard Taylor}
      & 
      & 
      & 
      \\
      &
      & \textcolor{tab-green}{Collapsed (ours)}
      & \textbf{}
      & \textbf{}
      & \textbf{}
      \\ \cmidrule{2-6}
      & \multirow{3}{*}{\makecell{Mem.\,[MiB] \\ (non-diff.)}}
      & \textcolor{tab-blue}{Nested 1\textsuperscript{st}-order}
      & 
      & 
      & 
      \\
      &
      & \textcolor{tab-orange}{Standard Taylor}
      & 
      & 
      & 
      \\
      &
      & \textcolor{tab-green}{Collapsed (ours)}
      & \textbf{}
      & \textbf{}
      & \textbf{}
      \\
      \midrule
      \multirow{9}{*}{\textbf{Stochastic}}
      & \multirow{3}{*}{Time [ms]}
      & \textcolor{tab-blue}{Nested 1\textsuperscript{st}-order}
      & 
      & 
      & 
      \\
      &
      & \textcolor{tab-orange}{Standard Taylor}
      & 
      & 
      & 
      \\
      &
      & \textcolor{tab-green}{Collapsed (ours)}
      & \textbf{}
      & \textbf{}
      & \textbf{}
      \\ \cmidrule{2-6}
      & \multirow{3}{*}{\makecell{Mem.\,[MiB]\\(differentiable)}}
      & \textcolor{tab-blue}{Nested 1\textsuperscript{st}-order}
      & 
      & 
      & 
      \\
      &
      & \textcolor{tab-orange}{Standard Taylor}
      & 
      & 
      & 
      \\
      &
      & \textcolor{tab-green}{Collapsed (ours)}
      & \textbf{}
      & \textbf{}
      & \textbf{}
      \\ \cmidrule{2-6}
      & \multirow{3}{*}{\makecell{Mem.\,[MiB]\\(non-diff.)}}
      & \textcolor{tab-blue}{Nested 1\textsuperscript{st}-order}
      & 
      & 
      & 
      \\
      &
      & \textcolor{tab-orange}{Standard Taylor}
      & 
      & 
      & \textbf{}
      \\
      &
      & \textcolor{tab-green}{Collapsed (ours)}
      & \textbf{}
      & \textbf{}
      & \textbf{}
      \\
      \bottomrule
    \end{tabular}
  \end{small}
\end{table}

\begin{figure*}[!t]
  \small
  \centering
  \savebox{\benchmarkLegend}{
    \begin{tikzpicture}[font=\small]
      \matrix [%
      matrix of nodes,%
      ampersand replacement=\&,%
      nodes={anchor=west, align=left, inner sep=1pt},%
      column sep=1ex,%
      row sep=0ex,%
      ] (legend)
      {
        \draw[tab-blue] plot[mark=*] coordinates {(0,0)};
        \& Nested 1\textsuperscript{st}-order\phantom{y}\!\!\!\!
        \\
        \draw[tab-orange] plot[mark=triangle*, rotate=270] coordinates {(0,0)};
        \& Standard Taylor
        \\
        \draw[tab-green] plot[mark=triangle*, rotate=90] coordinates {(0,0)};
        \& Collapsed (ours)\phantom{y}\!\!\!\!
        \\[2ex]
        \node[anchor=center]{\tikz\draw[thick] (0, 0) to ++(2.5ex, 0);};
        \& Differentiable\phantom{y}
        \\
        \node[anchor=center, opacity=0.5]{\tikz\draw[thick, dashed] (0, 0) to ++(2.5ex, 0);};
        \& Non-diff.\phantom{y}
        \\
      };

      \draw[gray, rounded corners] (current bounding box.north west) rectangle (current bounding box.south east);
    \end{tikzpicture}
  }
  \newcolumntype{C}{ >{\centering\arraybackslash} m{0.18\textwidth} }
  \newcolumntype{D}{ >{\centering\arraybackslash} m{0.27\textwidth} }
  \newcolumntype{E}{ >{\centering\arraybackslash} m{0.22\textwidth} }
  \begin{tabular}{CDEE}
    & \makecell{\textbf{Laplacian} \\ $(D=50)$}
    & \makecell{\textbf{Weighted Laplacian} \\ $(D=50)$}
    & \makecell{\textbf{Biharmonic} \\ $(D=5)$}
    \\[1ex]
    \makecell{\textbf{Exact} \\ \\ \\ \\ \\ \\ }
    & \includegraphics{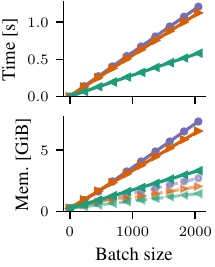}
    &  \includegraphics[trim={0.45cm 0 0 0},clip]{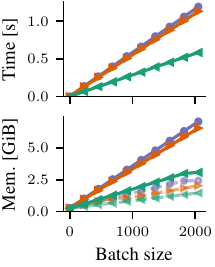}
    & \includegraphics[trim={0.45cm 0 0 0},clip]{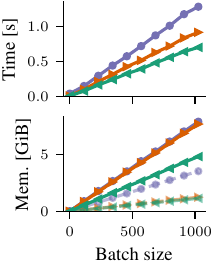}
    \\[-6ex]
  \scalebox{0.85}{
    \begin{tikzpicture}
      \node {\usebox{\benchmarkLegend}};
    \end{tikzpicture}
  }
    & ($N=2048$)
    & ($N=2048$)
    & ($N=256$)
    \\[-6ex]
    \makecell{\\ \textbf{Stochastic}}
    & \includegraphics{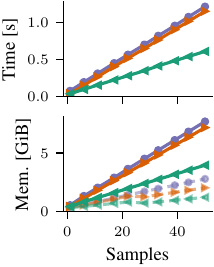}
    & \includegraphics[trim={0.45cm 0 0 0},clip]{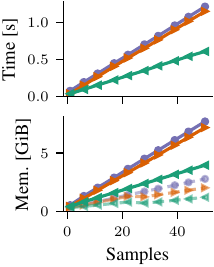}
    & \includegraphics[trim={0.45cm 0 0 0},clip]{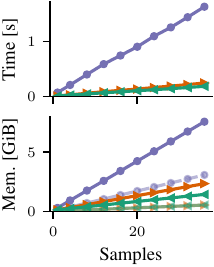}
  \end{tabular}

  \caption{\textbf{Same as \cref{fig:benchmark}, but using \texttt{torch.compile}.}}
  \label{fig:benchmark-compiled}
\end{figure*}

\subsection{Rank-deficient Weighted Laplacian}\label{sec:rank-deficient-weighted-laplacian}

In the main text we use a weighted Laplacian with a full-rank weight matrix (\ie, $R \coloneqq \rank(\mD) = D$).
Since the weight matrix has full rank, the weighted Laplacian is as expensive as the unweighted Laplacian, and this is confirmed by our experiments.
To show that the weight matrix's rank indeed affects the cost, we experiment with a rank-deficient weight matrix in this section and also consider the ranks $R \in \{ \nicefrac{D}{2}, \nicefrac{D}{10} \}$.
\Cref{tab:benchmark-rank} contains the results of this analysis.
We observe that going from full to half-full rank roughly halves both the runtime and memory consumption for all implementations.
For small ranks, this linear relationship weakens because the fraction of computations that do not scale with $R$ grows.

\begin{table}[!t]
  \centering
  \caption{\textbf{Exact weighted Laplacian for different ranks of the weightings.}
    The lower the rank, the lower the memory and time consumption.
    Both scale approximately linear with the rank---with stronger deviations for small ranks---as predicted by our theoretical analysis based on the number of forward-propagated coefficients.
    The `Full-rank' column is identical to the `Weighted Laplacian' columns in \Cref{tab:benchmark,tab:benchmark-compiled}.
  }
  \label{tab:benchmark-rank}
  \vspace{1.5ex}
  \def\datapathExactFullRankCompiled{jet/exp/exp01_benchmark_laplacian/performance/architecture_tanh_mlp_768_768_512_512_1_compiled_device_cuda_dim_50_name_weighted_laplacian_vary_batch_size_rank_ratio_1_0}
  \def\datapathExactHalfRankCompiled{jet/exp/exp01_benchmark_laplacian/performance/architecture_tanh_mlp_768_768_512_512_1_compiled_device_cuda_dim_50_name_weighted_laplacian_vary_batch_size_rank_ratio_0_5}
  \def\datapathExactLowRankCompiled{jet/exp/exp01_benchmark_laplacian/performance/architecture_tanh_mlp_768_768_512_512_1_compiled_device_cuda_dim_50_name_weighted_laplacian_vary_batch_size_rank_ratio_0_1}
  \def\datapathExactFullRank{jet/exp/exp01_benchmark_laplacian/performance/architecture_tanh_mlp_768_768_512_512_1_device_cuda_dim_50_name_weighted_laplacian_vary_batch_size_rank_ratio_1_0}
  \def\datapathExactHalfRank{jet/exp/exp01_benchmark_laplacian/performance/architecture_tanh_mlp_768_768_512_512_1_device_cuda_dim_50_name_weighted_laplacian_vary_batch_size_rank_ratio_0_5}
  \def\datapathExactLowRank{jet/exp/exp01_benchmark_laplacian/performance/architecture_tanh_mlp_768_768_512_512_1_device_cuda_dim_50_name_weighted_laplacian_vary_batch_size_rank_ratio_0_1}
  \sisetup{%
    round-mode=figures,%
    round-precision=2,%
    detect-weight, %
    tight-spacing=true, %
  }
  \begin{small}
    \begin{tabular}{ccc|ccc}
      \toprule
      \multirow{2}{*}{\!\!\!\textbf{\texttt{compile}}\!\!\!}
      & \multirow{2}{*}{\makecell{\textbf{Per-datum or } \\ \textbf{-sample cost}}}
      & \multirow{2}{*}{\textbf{Implementation}}
      & \multicolumn{3}{|c}{\textbf{Weighted Laplacian ($D=50$)}}
      \\
      &
      &
      & \makecell{Full-rank ($D$)}
      & \makecell{Half-full rank ($\nicefrac{D}{2}$)}
      & \makecell{Low-rank ($\nicefrac{D}{10}$)}
      \\
      \midrule
      \multirow{9}{*}{\textbf{\cmark}}
      & \multirow{3}{*}{Time [ms]}
      & \textcolor{tab-blue}{Nested 1\textsuperscript{st}-order}
      & 
      & \num{0.30307596995209785} (\num{1.0}x)
      & \num{0.08222602532972156} (\num{1.0}x)
      \\
      &
      & \textcolor{tab-orange}{Standard Taylor}
      & 
      & \num{0.27849909380375576} (\num{0.9189085292633838}x)
      & \num{0.062330573566910054} (\num{0.7580394810155069}x)
      \\
      &
      & \textcolor{tab-green}{Collapsed (ours)}
      & \textbf{}
      & \textbf{\num{0.1481911226455441} (\num{0.4889570184959441}x)}
      & \textbf{\num{0.04037795763590066} (\num{0.4910605550248526}x)}
      \\ \cmidrule{2-6}
      & \multirow{3}{*}{\makecell{Mem.\,[MiB] \\ (differentiable)}}
      & \textcolor{tab-blue}{Nested 1\textsuperscript{st}-order}
      & 
      & \num{1.7522452866264155} (\num{1.0}x)
      & \num{0.42790848919119395} (\num{1.0}x)
      \\
      &
      & \textcolor{tab-orange}{Standard Taylor}
      & 
      & \num{1.5835423908037376} (\num{0.9037218720973247}x)
      & \num{0.3326379456157352} (\num{0.7773576687961175}x)
      \\
      &
      & \textcolor{tab-green}{Collapsed (ours)}
      & \textbf{}
      & \textbf{\num{0.799173059415732} (\num{0.4560851528694215}x)}
      & \textbf{\num{0.21313957352320645} (\num{0.4980961558534855}x)}
      \\ \cmidrule{2-6}
      & \multirow{3}{*}{\makecell{Mem.\,[MiB] \\ (non-diff.)}}
      & \textcolor{tab-blue}{Nested 1\textsuperscript{st}-order}
      & 
      & \num{0.6203362115563233} (\num{1.0}x)
      & \num{0.15115199598919196} (\num{1.0}x)
      \\
      &
      & \textcolor{tab-orange}{Standard Taylor}
      & 
      & \num{0.47049477606376894} (\num{0.7584512515936698}x)
      & \num{0.12362921034648854} (\num{0.8179131842581064}x)
      \\
      &
      & \textcolor{tab-green}{Collapsed (ours)}
      & \textbf{}
      & \textbf{\num{0.30703197935161164} (\num{0.49494447306456285}x)}
      & \textbf{\num{0.072566394690888} (\num{0.48008889473134597}x)}
      \\
      \midrule
      \multirow{9}{*}{\textbf{\xmark}}
      & \multirow{3}{*}{Time [ms]}
      & \textcolor{tab-blue}{Nested 1\textsuperscript{st}-order}
      & \num{0.6009454650797135} (\num{1.0}x)
      & \num{0.31306521653479463} (\num{1.0}x)
      & \num{0.0837648203736978} (\num{1.0}x)
      \\
      &
      & \textcolor{tab-orange}{Standard Taylor}
      & \num{0.565736619095231} (\num{0.941410913251817}x)
      & \num{0.28676374521606524} (\num{0.9159872450543984}x)
      & \num{0.06503474930222813} (\num{0.7763969290698686}x)
      \\
      &
      & \textcolor{tab-green}{Collapsed (ours)}
      & \textbf{\num{0.290681297499431} (\num{0.48370661630814216}x)}
      & \textbf{\num{0.1523886139497662} (\num{0.4867631595630473}x)}
      & \textbf{\num{0.04248979408697831} (\num{0.50725106193053}x)}
      \\ \cmidrule{2-6}
      & \multirow{3}{*}{\makecell{Mem.\,[MiB]\\(differentiable)}}
      & \textcolor{tab-blue}{Nested 1\textsuperscript{st}-order}
      & \num{4.395973857024179} (\num{1.0}x)
      & \num{2.2457992684383665} (\num{1.0}x)
      & \num{0.5273687659419067} (\num{1.0}x)
      \\
      &
      & \textcolor{tab-orange}{Standard Taylor}
      & \num{4.553236470252666} (\num{1.035774237596341}x)
      & \num{2.355389629241003} (\num{1.0487979323632255}x)
      & \num{0.5979235756607456} (\num{1.1337864778412208}x)
      \\
      &
      & \textcolor{tab-green}{Collapsed (ours)}
      & \textbf{\num{2.0595845966509767} (\num{0.46851611579992347}x)}
      & \textbf{\num{1.1314784750543085} (\num{0.503819949964224}x)}
      & \textbf{\num{0.38993347418144353} (\num{0.739394327771769}x)}
      \\ \cmidrule{2-6}
      & \multirow{3}{*}{\makecell{Mem.\,[MiB]\\(non-diff.)}}
      & \textcolor{tab-blue}{Nested 1\textsuperscript{st}-order}
      & \num{2.1918429240450417} (\num{1.0}x)
      & \num{1.1165443704355433} (\num{1.0}x)
      & \num{0.25748244461888053} (\num{1.0}x)
      \\
      &
      & \textcolor{tab-orange}{Standard Taylor}
      & \num{1.1901513698932633} (\num{0.5429911773498994}x)
      & \num{0.6037358013938453} (\num{0.5407181455389356}x)
      & \num{0.13537026623498113} (\num{0.525745615144182}x)
      \\
      &
      & \textcolor{tab-green}{Collapsed (ours)}
      & \textbf{\num{0.8971034062805763} (\num{0.4092918322016314}x)}
      & \textbf{\num{0.4572917299060127} (\num{0.40955983659443074}x)}
      & \textbf{\num{0.10601757853686448} (\num{0.4117468229486061}x)}
      \\
      \bottomrule
    \end{tabular}
  \end{small}
\end{table}

\section{JAX Benchmark}
\label{sec:jax-benchmark}
This section presents experiments which show that the graph simplifications we propose to collapse standard Taylor mode are currently not applied by the \texttt{jit} compiler in JAX.

\begin{figure*}[!t]
  \centering

  \newcolumntype{C}{ >{\centering\arraybackslash} m{0.11\textwidth} }
  \newcolumntype{D}{ >{\centering\arraybackslash} m{0.4\textwidth} }
  \begin{tabular}{CDD}
    & \textbf{Laplacian $(D=50)$}
    & \makecell{\textbf{Biharmonic $(D=5)$} \\ \textbf{(via nested Laplacians)}}
    \\
    \textbf{Exact}
    & \includegraphics{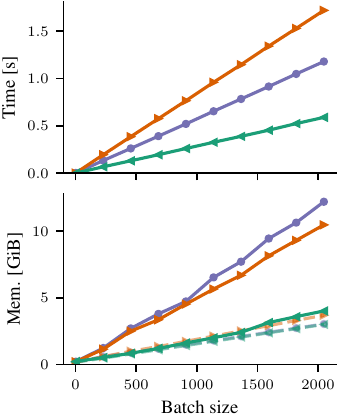}
    & \includegraphics{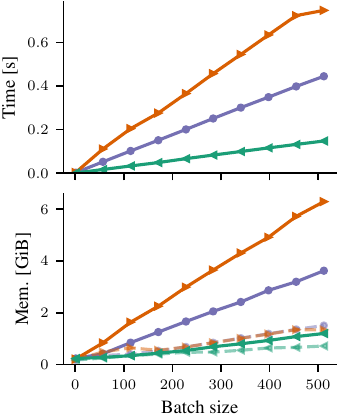}
    \\
    & $(N=2048)$
    & $(N=256)$
    \\
    \textbf{Stochastic}
    & \includegraphics{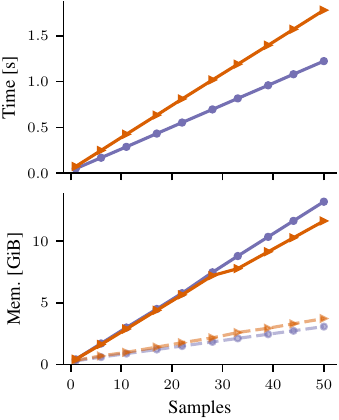}
    & \includegraphics{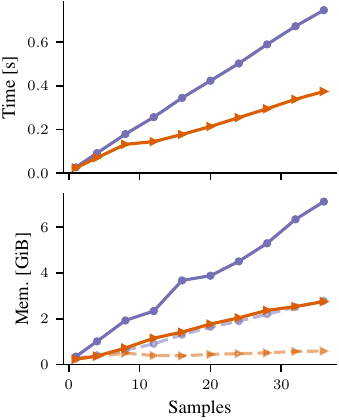}
    \\
  \end{tabular}
  \captionof{figure}{\textbf{JAX's \texttt{jit} compiler does not apply our graph simplifications to standard Taylor mode.} Colors: \textcolor{tab-green}{Collapsed Taylor mode}, \textcolor{tab-orange}{standard Taylor mode}, and \textcolor{tab-blue}{nested first-order automatic differentiaion}, \textcolor{black!50!white}{opaque} memory consumptions are for non-differentiable computations.
    Results are on GPU and we use a $D \to 768 \to 768 \to 512 \to 512 \to 1$ MLP with tanh activations, varying the batch size.
    For each approach, we fit a line to the data and report the slope in \cref{tab:jax-benchmark} to quantify the relative speedup and memory reduction.
  }
  \label{fig:jax-benchmark}
\end{figure*}

\begin{figure*}[!t]
  \centering

  \captionof{table}{\textbf{JAX Benchmark from \cref{fig:jax-benchmark} in numbers.}
    We fit linear functions and report their slopes, \ie, how much runtime and memory increase when incrementing the batch size.
    All numbers are shown with two significant digits and bold values are best according to parenthesized values.}
  \label{tab:jax-benchmark}
  \vspace{1.5ex}
  \def\datapathJAXLaplacianExact{jet/exp/exp04_jax_benchmark/performance/architecture_tanh_mlp_768_768_512_512_1_device_cuda_dim_50_name_jax_laplacian_vary_batch_size}
  \def\datapathJAXBilaplacianExact{jet/exp/exp04_jax_benchmark/performance/architecture_tanh_mlp_768_768_512_512_1_device_cuda_dim_5_name_jax_bilaplacian_vary_batch_size}
  \def\datapathJAXLaplacianStochastic{jet/exp/exp04_jax_benchmark/performance/architecture_tanh_mlp_768_768_512_512_1_batch_size_2048_device_cuda_dim_50_distribution_normal_name_jax_laplacian_vary_num_samples}
  \def\datapathJAXBilaplacianStochastic{jet/exp/exp04_jax_benchmark/performance/architecture_tanh_mlp_768_768_512_512_1_batch_size_256_device_cuda_dim_5_distribution_normal_name_jax_bilaplacian_vary_num_samples}
  \sisetup{%
    round-mode=figures,%
    round-precision=2,%
    detect-weight, %
    tight-spacing=true, %
  }
  \let\origsiunitxnum\num
  \renewcommand{\num}[1]{\IfStrEq{#1}{nan}{\text{n/a} }{\origsiunitxnum{#1}}}
  \begin{tabular}{ccc|ccc}
    \toprule
    \textbf{Mode}
    & \makecell{\textbf{Per-datum or} \\ \textbf{sample cost}}
    & \textbf{Implementation}
    & \textbf{Laplacian}
    & \makecell{\textbf{Biharmonic} \\ \!\!\textbf{(via nested Laplacians)}\!\!}
    \\
    \midrule
    \multirow{9}{*}{\textbf{Exact}}
    & \multirow{3}{*}{Time [ms]}
    & \textcolor{tab-blue}{Nested 1\textsuperscript{st}-order}
    & \input{\datapathJAXLaplacianExact/hessian_trace_best.txt}
    & \input{\datapathJAXBilaplacianExact/hessian_trace_best.txt}
    \\
    &
    & \textcolor{tab-orange}{Standard Taylor}
    & \input{\datapathJAXLaplacianExact/jet_naive_best.txt}
    & \input{\datapathJAXBilaplacianExact/jet_naive_best.txt}
    \\
    &
    & \textcolor{tab-green}{Collapsed (ours)}
    & \textbf{\input{\datapathJAXLaplacianExact/jet_simplified_best.txt}}
    & \textbf{\input{\datapathJAXBilaplacianExact/jet_simplified_best.txt}}
    \\ \cmidrule{2-5}
    & \multirow{3}{*}{\makecell{Mem.\,[MiB] \\ (differentiable)}}
    & \textcolor{tab-blue}{Nested 1\textsuperscript{st}-order}
    & \input{\datapathJAXLaplacianExact/hessian_trace_peakmem.txt}
    & \input{\datapathJAXBilaplacianExact/hessian_trace_peakmem.txt}
    \\
    &
    & \textcolor{tab-orange}{Standard Taylor}
    & \input{\datapathJAXLaplacianExact/jet_naive_peakmem.txt}
    & \input{\datapathJAXBilaplacianExact/jet_naive_peakmem.txt}
    \\
    &
    & \textcolor{tab-green}{Collapsed (ours)}
    & \textbf{\input{\datapathJAXLaplacianExact/jet_simplified_peakmem.txt}}
    & \textbf{\input{\datapathJAXBilaplacianExact/jet_simplified_peakmem.txt}}
    \\ \cmidrule{2-5}
    & \multirow{3}{*}{\makecell{Mem.\,[MiB] \\ (non-diff.)}}
    & \textcolor{tab-blue}{Nested 1\textsuperscript{st}-order}
    & \input{\datapathJAXLaplacianExact/hessian_trace_peakmem_nondifferentiable.txt}
    & \input{\datapathJAXBilaplacianExact/hessian_trace_peakmem_nondifferentiable.txt}
    \\
    &
    & \textcolor{tab-orange}{Standard Taylor}
    & \input{\datapathJAXLaplacianExact/jet_naive_peakmem_nondifferentiable.txt}
    & \input{\datapathJAXBilaplacianExact/jet_naive_peakmem_nondifferentiable.txt}
    \\
    &
    & \textcolor{tab-green}{Collapsed (ours)}
    & \textbf{\input{\datapathJAXLaplacianExact/jet_simplified_peakmem_nondifferentiable.txt}}
    & \textbf{\input{\datapathJAXBilaplacianExact/jet_simplified_peakmem_nondifferentiable.txt}}
    \\
    \midrule
    \multirow{9}{*}{\textbf{Stochastic}}
    & \multirow{3}{*}{Time [ms]}
    & \textcolor{tab-blue}{Nested 1\textsuperscript{st}-order}
    & \textbf{\input{\datapathJAXLaplacianStochastic/hessian_trace_best.txt}}
    & \input{\datapathJAXBilaplacianStochastic/hessian_trace_best.txt}
    \\
    &
    & \textcolor{tab-orange}{Standard Taylor}
    & \input{\datapathJAXLaplacianStochastic/jet_naive_best.txt}
    & \textbf{\input{\datapathJAXBilaplacianStochastic/jet_naive_best.txt}}
    \\
    &
    & \textcolor{tab-green}{Collapsed (ours)}
    & Not implemented %
    & Not implemented %
    \\ \cmidrule{2-5}
    & \multirow{3}{*}{\makecell{Mem.\,[MiB] \\ (differentiable)}}
    & \textcolor{tab-blue}{Nested 1\textsuperscript{st}-order}
    & \input{\datapathJAXLaplacianStochastic/hessian_trace_peakmem.txt}
    & \input{\datapathJAXBilaplacianStochastic/hessian_trace_peakmem.txt}
    \\
    &
    & \textcolor{tab-orange}{Standard Taylor}
    & \textbf{\input{\datapathJAXLaplacianStochastic/jet_naive_peakmem.txt}}
    & \textbf{\input{\datapathJAXBilaplacianStochastic/jet_naive_peakmem.txt}}
    \\
    &
    & \textcolor{tab-green}{Collapsed (ours)}
    & Not implemented %
    & Not implemented %
    \\ \cmidrule{2-5}
    & \multirow{3}{*}{\makecell{Mem.\,[MiB] \\ (non-diff.)}}
    & \textcolor{tab-blue}{Nested 1\textsuperscript{st}-order}
    & \textbf{\input{\datapathJAXLaplacianStochastic/hessian_trace_peakmem_nondifferentiable.txt}}
    & \input{\datapathJAXBilaplacianStochastic/hessian_trace_peakmem_nondifferentiable.txt}
    \\
    &
    & \textcolor{tab-orange}{Standard Taylor}
    & \input{\datapathJAXLaplacianStochastic/jet_naive_peakmem_nondifferentiable.txt}
    & \textbf{\input{\datapathJAXBilaplacianStochastic/jet_naive_peakmem_nondifferentiable.txt}}
    \\
    &
    & \textcolor{tab-green}{Collapsed (ours)}
    & Not implemented %
    & Not implemented %
    \\
    \bottomrule
  \end{tabular}
  \let\num\origsiunitxnum
\end{figure*}

\paragraph{Comparing JAX implementations.} Similar to our PyTorch experiment in \Cref{sec:experiments}, we compare three implementations of the Laplacian in JAX (all compiled with \texttt{jax.jit}):

\begin{enumerate}
\item \textbf{\textcolor{tab-blue}{Nested 1\textsuperscript{st}-order AD}} computes the Hessian using \texttt{jax.hessian}, which relies on forward-over-reverse mode, then traces it.

\item \textbf{\textcolor{tab-orange}{Standard Taylor mode}} propagates multiple univariate Taylor polynomials, each of which computes one element of the Hessian diagonal, then sums them to obtain the Laplacian.
  This is implemented with \texttt{jax.experimental.jet.jet} and \texttt{jax.vmap}.

\item \textbf{\textcolor{tab-green}{Collapsed Taylor mode}} relies on the forward Laplacian implementation in JAX provided by the \texttt{folx} library \cite{gao2023folx} and implements our proposed collapsed Taylor mode for the specific case of the Laplacian.
  \texttt{folx} also enables leveraging sparsity in the tensors, which is beneficial for architectures in VMC.
  To disentangle runtime improvements from sparsity detection versus collapsing Taylor coefficient, we disable \texttt{folx}'s sparsity detection.
\end{enumerate}
For the biharmonic operator, we simply nest the Laplacian implementations.

We only investigate computing the exact Laplacian, as the forward Laplacian in \texttt{folx} currently does not support stochastic computation.
We use the same neural network architecture as for our PyTorch experiments, fix the input dimension to $D=50$ and vary the batch size, recording the runtime and peak memory with the same protocol as described in the main text.
JAX is purely functional and therefore does not have a mechanism to build up a differentiable computational graph similar to evaluating a function in PyTorch where some leafs have \texttt{requires\_grad=True}.
To approximate the peak memory of computing a differentiable Laplacian in JAX, we measure the peak memory of first computing the Laplacian, then evaluating the gradient \wrt the neural network's parameters which backpropagates through the same computation graph built by PyTorch.

\paragraph{Results (Laplacian).} The left column of \cref{fig:jax-benchmark} visualizes the performance of the three implementations.
We fit linear functions to each of them and report the cost incurred by adding one more datum to the batch in \cref{tab:jax-benchmark}.
From them, we draw the following conclusions:

\begin{enumerate}
\item \textbf{Performance is consistent between PyTorch and JAX.} Although our PyTorch implementation does not leverage compilation, the values reported in \cref{tab:benchmark,tab:jax-benchmark} are consistent and only in rare cases differ by a factor of more than two.
  This confirms that our PyTorch-based implementation of Taylor mode is reasonably efficient, and that the presented performance results in the main text are transferable to other frameworks like JAX.

\item \textbf{Our implementation of collapsed Taylor mode based on graph rewrites in PyTorch achieves consistent speed-up with the Laplacian-specific implementation in JAX.}
  Specifically, we observe that \textcolor{tab-green}{collapsed Taylor mode/forward Laplacian} use roughly half the runtime of \textcolor{tab-blue}{nested 1\textsuperscript{st}-order AD} (compare \cref{tab:benchmark,tab:jax-benchmark}).
  This supports our argument that our collapsed Taylor is indeed a generalization of the forward Laplacian, \ie, the latter does not employ additional tricks (leveraging sparsity could also be applied to our approach but we are not aware of a drop-in implementation).
  It also illustrates that the savings we report in PyTorch carry over to other frameworks like JAX.

\item \textbf{JAX's \texttt{jit} compiler is unable to apply the graph rewrites we propose in this work.}
  If the JAX compiler was able to perform our proposed graph rewrites, then the \texttt{jit}-compiled \textcolor{tab-orange}{standard Taylor mode} should yield similar performance than the \textcolor{tab-green}{forward Laplacian}.
  However, we observe a clear performance gap in runtime and memory, from which we conclude that the compilation did not collapse the Taylor coefficients.
  Our contribution is to point out that such rewrites could easily be added to the compiler's ability to unlock these performance gains at zero user overhead.
\end{enumerate}

\paragraph{Results (biharmonic operator).}
For the biharmonic operator (right column of \cref{fig:jax-benchmark} and \cref{tab:jax-benchmark}), we conclude that (i) the most efficient way to compute biharmonics is by nesting Laplacians (compare with \cref{tab:benchmark} where Taylor mode uses the approach for general linear differential operators) and (ii) that nesting Taylor mode Laplacians is more efficient than nesting 1\textsuperscript{st}-order AD Laplacians, while also allowing to apply our collapsing technique.

\section{Numerical Complexity and Error Analysis}
\label{sec:numerical-analysis}
\paragraph{Setup.}
To illustrate the numerical properties of our proposed collapsed Taylor mode, we consider a two-layered MLP with element-wise $\mathrm{tanh}$ activation $\vphi: \sR^I \to \sR^I$.
The MLP is denoted by $\vf := \vg \circ \vphi \circ \vh$.
The two linear layers are given as $\vh: \sR^D \to \sR^I, \vh(\vx_0) = \mW_1 \vx_0 + \vb_1$ and $\vg: \sR^I \to \sR^C$, $\vg(\vphi_0) = \mW_2 \vphi_0 + \vb_2$, with weights $\mW_1 \in \sR^{I \times D}, \mW_2 \in \sR^{C \times I}$ and bias $\vb_1 \in \sR^I, \vb_2 \in \sR^C$.
Below we compare the computational and storage complexity, as well as stability for evaluating the sum of the second coefficients $\smash{\sum_{r=1}^R \langle \partial^2 \vf(\vx_0), \vv_i^{\otimes 2} \rangle = \sum_{r=1}^R \vg_{2,r}}$ (see \cref{eq:sum-k-directional}) between collapsed and standard Taylor mode. For this toy example we show (i) collapsing uses less operations and (ii) both methods are similarly stable based on our simplified error analysis.

\paragraph{Computational \& storage complexity.}
Both vanilla and collapsed Taylor mode evaluate the function values ($\vh_0, \vphi_0, \vg_0$) and the first derivatives ($\{\vh_{1,r}, \vphi_{1, r}, \vg_{1, r}\}$) by propagating $1 + R$ coefficients at each layer
\begin{equation*}
  \begin{pmatrix*}[l]
    \vh_0
    =
    \mW_1 \vx_0 + \vb_1
    \\
    \left\{
    \vh_{1,r}
    \right\}
    =
    \left\{
    \left\langle \mW_1, \vx_{1,r}\right\rangle
    \right\}
  \end{pmatrix*}
  \overset{\text{(\ref{eq:faa-di-bruno})}}{\to}
  \begin{pmatrix*}[l]
    \vphi_0
    =
    \vphi(\vh_0)
    \\
    \left\{
    \vphi_{1,r}
    \right\}
    =
    \left\{
    \left\langle \partial \vphi(\vh_0), \vh_{1,r} \right\rangle
    \right\}
  \end{pmatrix*}
  \overset{\text{(\ref{eq:faa-di-bruno})}}{\to}
  \begin{pmatrix*}[l]
    \vg_0
    =
    \mW_2 \vphi_0 + \vb_2
    \\
    \left\{
    \vg_{1,r}
    \right\}
    =
    \left\{
    \left\langle\mW_2, \vphi_{1,r}\right\rangle
    \right\}
  \end{pmatrix*}
\end{equation*}
with $\partial \vphi(\vh_0) = \partial \mathrm{tanh}(\vh_0) = \diag(\vone - \vphi_0^{\odot 2}) \in \sR^{I \times I}$ the $\mathrm{tanh}$-activation layer's Jacobian.
The propagation costs $1 + R$ matrix-vector multiplications with $\mW_1$, $1 + R$ matrix-vector multiplications with $\mW_2$, $R$ Hadamard products with the derivative of $\vphi$ (since $\left\langle \diag(\va), \vh_{1,r} \right\rangle = \va \odot \vh_{1,r}$), and one Hadamard product to compute $\partial \vphi(\vh_0)$.
Additionally, there is one vector addition with the bias $\vb_1$, one vector addition with $\vb_2$, one vector subtraction in $\partial \vphi(\vh_0)$ (counted as vector addition), as well as the evaluation of $\vphi(\vh_0)$.
$3+3R$ vectors are stored.

\begin{table}[!b]
  \caption{\textbf{Comparison of theoretical computational and storage complexity} between standard Taylor mode and collapsed Taylor mode for a two-layer MLP computing the sum $\smash{\sum_{r=1}^R \langle \partial^2 \vf(\vx_0), \vv_r^{\otimes 2} \rangle}$.}
  \label{tab:run-time-storage-comparison}
  \vspace{1ex}
  \centering
  \renewcommand{\arraystretch}{1.2}
  \begin{tabular}{l|cc}
    \toprule
    \multicolumn{3}{c}{\textbf{Computational Complexity}} \\
    \midrule
    \textbf{Operation}
    & \textcolor{tab-orange}{Standard Taylor}
    & \textcolor{tab-green}{Collapsed (ours)} \\
    \midrule
    \# Matrix-vector products
    & $4R + 2$
    & $2R + 4$ \\

    \# Hadamard products
    & $4R + 2$
    & $3R + 3$ \\

    \# Vector additions
    & $2R + 2$
    & $2R + 2$ \\

    \# Scalar multiplications
    & $1$
    & $1$ \\

    \# Activation evaluations
    & $I$
    & $I$ \\
    \midrule
    \multicolumn{3}{c}{\textbf{Storage Complexity}} \\
    \midrule
    \# Vectors stored
    & $6R + 3$
    & $3R + 6$ \\
    \bottomrule
  \end{tabular}
\end{table}

For the second derivatives, vanilla Taylor mode propagates $R$ vectors
\begin{equation*}
  \begin{aligned}
    \begin{pmatrix*}[l]
      \left\{\vh_{2,r}\right\}
      =
      \left\{\left\langle \mW_1, \vx_{2,r} \right\rangle\right\}
    \end{pmatrix*}
    &\overset{\text{(\ref{eq:faa-di-bruno})}}{\to}
      \begin{pmatrix*}[l]
        \left\{
        \vphi_{2,r}
        \right\}
        =
        \left\{
        \left\langle \partial^2 \vphi(\vh_0), \vh_{1, r}^{\otimes 2} \right\rangle + \left\langle \partial \vphi(\vh_0), \vh_{2,r} \right \rangle
        \right\}
      \end{pmatrix*}
    \\
    &\overset{\text{(\ref{eq:faa-di-bruno})}}{\to}
      \begin{pmatrix*}[l]
        \left\{
        \vg_{2,r}
        \right\}
        =
        \left\{
        \left\langle\mW_2 , \vphi_{2,r}\right\rangle
        \right\}
      \end{pmatrix*}
  \end{aligned}
\end{equation*}
with activation Hessian $\partial^2 \vphi(\vh_0) \in \sR^{I \times I \times I}$ of entries $[\partial^2 \vphi(\vh_0)]_{i,j,k} = [-2 \vphi_0 \odot (\vone - \vphi_0^{\odot 2})]_i \delta_{i,j,k}$ and contraction $\smash{\left\langle \partial^2 \vphi(\vh_0), \vh_{1, r}^{\otimes 2} \right\rangle = -2 \vphi_0 \odot (\vone - \vphi_0^{\odot 2}) \odot \vh_{1,r}^{\odot 2}}$.
These vectors are summed up to get the result $\smash{\sum_{r=1}^R \langle \partial^2 \vf(\vx_0), \vv_i^{\otimes 2} \rangle = \sum_{r=1}^R \vg_{2,r}}$.
This costs $2R$ matrix-vector products with the weights, $1 + 3R$ Hadamard products, $2R - 1$ vector additions, and a single scalar multiplication.

In contrast, collapsed Taylor mode propagates only a single summed vector
\begin{equation*}
  \begin{aligned}
    \begin{pmatrix*}[l]
      \displaystyle\sum_{r=1}^R \vh_{2,r} = \left\langle \mW_1,  \sum_{r=1}^R\vx_{2,r} \right\rangle
    \end{pmatrix*}
    \overset{\text{(\ref{eq:faa-di-bruno})}}{\to}
    &\begin{pmatrix*}[l]
      \displaystyle\sum_{r=1}^R\vphi_{2,r} = \sum_{r=1}^R \left\langle \partial^2 \vphi(\vh_0),  \vh_{1, r}^{\otimes 2} \right\rangle + \left\langle \partial \vphi(\vh_0), \sum_{r=1}^R \vh_{2,r} \right\rangle
    \end{pmatrix*}
    \\
    \overset{\text{(\ref{eq:faa-di-bruno})}}{\to}
    &\begin{pmatrix*}[l]
      \displaystyle\sum_{r=1}^R\vg_{2,r} = \left\langle\mW_2, \sum_{r=1}^R\vphi_{2,r}\right\rangle
    \end{pmatrix*}.
  \end{aligned}
\end{equation*}
This costs two matrix-vector products, $2 + 2R$ Hadamard products, $2R-1$ vector additions, and a single scalar multiplication. 
\Cref{tab:run-time-storage-comparison} summarizes the accumulated costs.

\paragraph{Error analysis.} For our numerical experiments, the result of all implementations (nested 1\textsuperscript{st}-order and standard/collapsed Taylor mode) was always checked to be close.
To supplement this experimental error analysis, we sketch a simplified error analysis below.
We assume that there are error-prone first- and second-order inputs $\{\vx_{1, r} + \vvarepsilon_{1, r}\}_r$ and $\{\vx_{2, r} + \vvarepsilon_{2, r}\}_r$ with errors $\{\vvarepsilon_{1, r}, \vvarepsilon_{2, r}\}_{r=1}^R$ that can be seen as the error of previous propagation steps.
An error-prone $\vx_0$ would complicate our brief discussion too much and is ignored here.
We consider again $\vf = \vg \circ \vphi \circ \vh$.
The error-influenced coefficients are denoted $\vg_{2, r}^\varepsilon$.

Using vanilla Taylor mode, the erroneous result is
\begin{align*}
  \sum_{r=1}^R \vg^\varepsilon_{2,r}
  &=
    \sum_{r=1}^R \Big( \Big\langle\mW_2,
    \left\langle \partial^2 \vphi(\vh_0),
    \left\langle \mW_1,
    \vx_{1,r} +  \vvarepsilon_{1, r} \right\rangle^{\otimes 2}
    \right\rangle
  +
    \left\langle
    \partial \vphi(\vh_0),
    \left\langle
    \mW_1,
    \vx_{2,r}
    +
    \vvarepsilon_{2,r}
    \right\rangle
    \right\rangle
    \Big\rangle
    \Big)
  \\
  &=
    \sum_{r=1}^R \vg_{2, r}
  \\
  &\phantom{=}+
    \sum_{r=1}^R \Big(\left\langle\mW_2,
    \left\langle \partial^2 \vphi(\vh_0),
    \left\langle \mW_1,
    \vx_{1,r}
    \right\rangle
    \otimes
    \left \langle
    \mW_1,
    \vvarepsilon_{1, r}
    \right \rangle
    \right\rangle
    \right \rangle
    \\
    &\phantom{===}+
    \left\langle\mW_2,
    \left\langle \partial^2 \vphi(\vh_0),
    \left \langle
    \mW_1,
    \vvarepsilon_{1, r}
    \right \rangle
    \otimes
    \left\langle \mW_1,
    \vx_{1,r}
    \right\rangle
    \right\rangle
    \right\rangle
  \\
  &\phantom{===}+
    \left\langle\mW_2,
    \left\langle \partial^2 \vphi(\vh_0),
    \langle \mW_1,
    \vvarepsilon_{1,r}
    \rangle^{\otimes 2}
    \right \rangle
    \right\rangle
    +
    \left\langle\mW_2,
    \left\langle
    \partial \vphi(\vh_0),
    \left\langle
    \mW_1,
    \vvarepsilon_{2,r}
    \right\rangle
    \right\rangle
    \right\rangle
    \Big)
  \\
  &=
    \sum_{r=1}^R \vg_{2, r}
    +
    \Delta \vg_{2,R}^S
    +
    \sum_{r= 1}^R
    \left \langle \mW_2
    \left\langle
    \partial \vphi(\vh_0),
    \left\langle
    \mW_1,
    \vvarepsilon_{2,r}
    \right\rangle
    \right\rangle
    \right\rangle.
\end{align*}
All errors related to the first-order coefficients are summarized in
\begin{align*}
  \Delta \vg_{2,R}^S &:= \sum_{r=1}^R \Big(\left\langle\mW_2,
                       \left\langle \partial^2 \vphi(\vh_0),
                       \left\langle \mW_1,
                       \vx_{1,r}
                       \right\rangle
                       \otimes
                       \left \langle
                       \mW_1,
                       \vvarepsilon_{1, r}
                       \right \rangle
                       \right\rangle
                       \right \rangle
                       \\
                       &\phantom{:==}+
                       \left\langle\mW_2,
                       \left\langle \partial^2 \vphi(\vh_0),
                       \left \langle
                       \mW_1,
                       \vvarepsilon_{1, r}
                       \right \rangle
                       \otimes
                       \left\langle \mW_1,
                       \vx_{1,r}
                       \right\rangle
                       \right\rangle
                       \right\rangle
  \\
                     &\phantom{:==}+
                       \left\langle\mW_2,
                       \left\langle \partial^2 \vphi(\vh_0),
                       \langle \mW_1,
                       \vvarepsilon_{1,r}
                       \rangle^{\otimes 2}
                       \right\rangle
                       \right\rangle
                       \Big)\,.
\end{align*}
The collapsed Taylor mode results in
\begin{align*}
  \sum_{r=1}^R \vg^\varepsilon_{2,r}
  &=
    \left\langle\mW_2,  \sum_{r=1}^R \left(
    \left\langle \partial^2 \vphi(\vh_0),
    \left\langle \mW_1,
    \vx_{1,r} + \vvarepsilon_{1, r} \right\rangle^{\otimes 2}
    \right\rangle
    \right)
    \right\rangle
  \\
  &+
    \left\langle\mW_2,
    \left\langle
    \partial \vphi(\vh_0),
    \left\langle
    \mW_1,
    \sum_{r=1}^R \left(
    \vx_{2,r}
    +
    \vvarepsilon_{2,r}
    \right)
    \right\rangle
    \right\rangle
    \right\rangle
  \end{align*}
  \begin{align*}
  &=
    \sum_{r=1}^R \vg_{2,r}
  \\
  &\phantom{=}+
    \Big\langle
    \mW_2,
    \sum_{r=1}^R
    \Big(
    \langle
    \partial^2 \vphi(\vh_0),
    \langle
    \mW_1, \vx_{1, r}
    \rangle
    \otimes
    \langle
    \mW_1, \vvarepsilon_{1, r}
    \rangle
    \rangle
    +
    \langle
    \partial^2 \vphi(\vh_0),
    \langle
    \mW_1, \vvarepsilon_{1, r}
    \rangle
    \otimes
    \langle
    \mW_1, \vx_{1, r}
    \rangle
    \rangle
  \\
  &\phantom{======}+
    \langle
    \partial^2 \vphi(\vh_0),
    \langle
    \mW_1, \vvarepsilon_{1, r}
    \rangle^{\otimes 2}
    \rangle
    \Big)
    \Big\rangle
  \\
  &\phantom{=}+
    \left \langle
    \mW_2,
    \left \langle \partial \vphi(\vh_0),
    \left\langle \mW_1,
    \sum_{r=1}^R
    \vvarepsilon_{2, r}
    \right\rangle
    \right \rangle
    \right \rangle
  \\
  &=
    \sum_{r=1}^R \vg_{2,r}
    +
    \Delta \vg_{2,R}^C
    +
    \left \langle
    \mW_2,
    \left \langle \partial \vphi(\vh_0),
    \left\langle \mW_1,
    \sum_{r=1}^R
    \vvarepsilon_{2, r}
    \right\rangle
    \right \rangle
    \right \rangle,
\end{align*}
where the first-order errors are collected in
\begin{align*}
  \Delta \vg_{2, R}^C &:=
                        \Big\langle
                        \mW_2,
                        \sum_{r=1}^R
                        \Big(
                        \langle
                        \partial^2 \vphi(\vh_0),
                        \langle
                        \mW_1, \vx_{1, r}
                        \rangle
                        \otimes
                        \langle
                        \mW_1, \vvarepsilon_{1, r}
                        \rangle
                        \rangle
                        \\
                        &\phantom{:=====}+
                        \langle
                        \partial^2 \vphi(\vh_0),
                        \langle
                        \mW_1, \vvarepsilon_{1, r}
                        \rangle
                        \otimes
                        \langle
                        \mW_1, \vx_{1, r}
                        \rangle
                        \rangle
                        +
                        \langle
                        \partial^2 \vphi(\vh_0),
                        \langle
                        \mW_1, \vvarepsilon_{1, r}
                        \rangle^{\otimes 2}
                        \rangle
                        \Big)
                        \Big\rangle\,.
\end{align*}
\paragraph{Error analysis (summary and discussion).} Without considering floating-point operations, the errors are equivalent.
This is not surprising, since our collapsing method is mathematically equivalent to the standard Taylor mode on the same input coefficients.

Incorporating floating-point operations for the function evaluations, inner product, tensor product, and summations would greatly complicate the discussion, which is not part of the paper.
Still, the error could be split into the same three parts for both vanilla and collapsed Taylor mode.
For the first-order errors $\smash{\Delta \vg_{2, R}^S}$ and $\smash{\Delta \vg_{2, R}^C}$, however, even with floating-point operations, the errors are structurally similar, since apart from the most outer inner product (with $\mW_2$) and the summation, all operations are done in the same order.
In practice, we would expect smaller errors for the collapsing method due to the reduced number of operations.
The second error term, which collects the error of the second-order coefficients, could also reduce the accumulation of error terms.
Of course, the actual condition and input, and output dimensions of the matrices are crucial.
Theoretically, this could even lead to a similar error asymptotically.
If inputs are small, one could argue that catastrophic cancellations are more likely to happen in our case, since we sum first.
But note that those cancellations are then also likely to happen in the standard Taylor mode, because weight matrices are often normalized, and the outputs of the activation functions are small if the input is small.

We plan to investigate this more rigorously in the future.

\section{Connections of Collapsed Taylor Mode to Existing Methods}
\label{sec:connections}
Here, we make the connection of collapsed Taylor mode to the forward Laplacian \cite{li2023forward} and the randomized estimation of the Laplacian via Hutchinson's trace estimator \cite{hutchinson1989stochastic} from \cite{shi2024stochastic} explicit.

\subsection{Connection to Randomized Laplacian via Hutchinson's Trace Estimator}

For simplicity, we consider a vector-to-scalar function $f: \sR^{D} \to \sR, \vx \mapsto f(\vx)$ (the general vector-to-vector case is straightforward but requires more notation) whose Laplacian is
\begin{align*}
  \Delta f(\vx) =
  \Tr( \nabla^2 f(\vx) )
  =
  \sum_{d=1}^{D} [\nabla^2 f(\vx)]_{d,d}
  =
  \sum_{d=1}^{D} \ve_d^{\top} \nabla^2f(\vx) \ve_d\,,
\end{align*}
with $\nabla^2 f(\vx) \in \sR^{D \times D}$ the Hessian of $f$ evaluated at $\vx$.
Because the Laplacian can be expressed as trace of the Hessian, we can use Hutchinson's trace estimator \cite{hutchinson1989stochastic} to estimate it via Hessian-vector products with random vectors.
Specifically, for any matrix $\mA \in \sR^{D \times D}$ and a distribution $p(\rvv)$ over a vector $\rvv$ with unit covariance ($\E[\rvv \rvv^{\top}] = \mI_D$) we can use the cyclic property of the trace and linearity of the expectation to write
\begin{align*}
  \Tr(\mA)
  =
  \Tr(\mA \mI_D)
  =
  \Tr( \mA \E[\rvv \rvv^{\top}])
  =
  \E[\Tr( \mA \rvv \rvv^{\top})]
  =
  \E[\Tr( \rvv^{\top} \mA \rvv)]\,.
\end{align*}
Then, we can compute an unbiased estimate of the right hand side by drawing $S$ vectors $\vv_1, \vv_2, \dots, \vv_S \sim p(\rvv)$ and evaluating the Monte-Carlo estimator
\begin{align*}
  \Tr(\mA)
  \approx
  \frac{1}{S} \sum_{s=1}^{S} \vv_s^{\top} \mA \vv_s\,.
\end{align*}
Applied to the Hessian, we can estimate the Laplacian of $f$ as
\begin{align*}
  \Delta f(\vx)
  \approx&
           \frac{1}{S} \sum_{s=1}^{S} \vv_s^{\top} \nabla^2 f(\vx) \vv_s\,.
           \shortintertext{Using our tensor notation, we can rewrite this into a sum of terms involving $\left\langle \partial^2f(\vx), \vv_s^{\otimes 2} \right\rangle$, which can be computed with \textcolor{tab-orange}{\textbf{vanilla Taylor mode}} using $S$ 2-jets (see \cref{eq:taylor-mode-composition}):}
           =&
              \frac{1}{S} \sum_{s=1}^{S} \sum_{i,j=1}^D [\partial^2f(\vx)]_{i,j} [\vv_s]_i [\vv_s]_j
              =
              \frac{1}{S} \sum_{s=1}^{S} \left\langle \partial^2f(\vx), \vv_s^{\otimes 2} \right\rangle\,.
              \shortintertext{Instead of propagating then summing the 2-jets, we can also sum the vectors and then propagate the sum (assuming we have a composition, see \cref{eq:laplacian-efficient}), which is our proposed \textcolor{tab-green}{\textbf{collapsed Taylor mode}}:}
              =& \frac{1}{S} \left\langle \partial^2f(\vx), \sum_{s=1}^{S} \vv_s^{\otimes 2} \right\rangle\,.
\end{align*}

\subsection{Connection to the Forward Laplacian}
We start by writing out the propagation rules of the forward Laplacian (eqs.~(5-7) in \citet{li2023forward}) for a function $f = g \circ \vh$ with $g: \sR^C \to \sR$ whose input we denote by $\vh_0 \in \sR^C$.
The forward propagation consumes $\vh_0 = \vh(\vx_0)$, the Jacobian $\nabla_{\vx_0} \vh_0 = \nabla_{\vx_0} \vh(\vx_0) \in \sR^{D \times C}$, and the Laplacian $\Delta_{\vx_0} \vh_0 = \Delta_{\vx_0} \vh(\vx_0) \in \sR^C$:
\begin{align*}
  \begin{pmatrix}
    \vh_0
    &\!\! \in \sR^C
    \\
    \nabla_{\vx_0} \vh_0
    &\!\! \in \sR^{D \times C}
    \\
    \Delta_{\vx_0} \vh_0
    &\!\! \in \sR^C
  \end{pmatrix}
  \!
  \to
  \!
  \begin{pmatrix}
    g_0 = g(\vh_0)
    &\!\! \in \sR
    \\
    \nabla_{\vx_0} g_0 = (\nabla_{\vx_0}\vh_0) (\nabla_{\vh_0} g_0)
    &\!\! \in \sR^D
    \\
    \Delta_{\vx_0} g_0 = (\nabla_{\vh_0}g_0)^{\top} \Delta \vh_0
    +
    \Tr \left(
    (\nabla_{\vx_0} \vh_0)^{\top} (\nabla_{\vx_0} \vh_0) \nabla_{\vh_0} g_0
    \right)
    &\!\! \in \sR
  \end{pmatrix}\,.
\end{align*}
Let us rewrite this in terms of rows of the Jacobian $[\nabla_{\vx_0} \vh_0]_{d,:} \in \sR^{C}$ (where the colon subscript denotes a slice):
\begin{align*}
  \begin{pmatrix}
    \vh_0
    \\
    \left\{ [\nabla_{\vx_0} \vh_0]_{d,:} \right\}_{d=1}^D
    \\
    \Delta_{\vx_0} \vh_0
  \end{pmatrix}
  \to
  \begin{pmatrix}
    g_0 = g(\vh_0)
    \\
    \left\{
    [\nabla_{\vx_0} g_0]_{d,:} = [\nabla_{\vx_0}\vh_0]_{d,:} (\nabla_{\vh_0} g_0)
    \right\}_{d=1}^{D}
    \\
    \Delta_{\vx_0} g_0
    =
    (\nabla_{\vh_0}g_0)^{\top} \Delta \vh_0
    +
    \sum_{d=1}^D
    [\nabla_{\vx_0} \vh_0]_{d,:} \nabla^2_{\vh_0} g_0 [\nabla_{\vx_0} \vh_0]_{d,:}^{\top}
  \end{pmatrix}\,.
\end{align*}
In our tensor notation, this translates to
\begin{align*}
  \begin{pmatrix}
    \vh_0
    \\
    \left\{ [\nabla_{\vx_0} \vh_0]_{d,:} \right\}_{d=1}^D
    \\
    \Delta_{\vx_0} \vh_0
  \end{pmatrix}
  \to
  \begin{pmatrix}
    g_0 = g(\vh_0)
    \\
    \left\{
    [\nabla_{\vx_0} g_0]_{d,:} = \left\langle \partial g(\vh_0), [\nabla_{\vx_0}\vh_0]_{d,:} \right\rangle
    \right\}_{d=1}^{D}
    \\
    \Delta_{\vx_0} g_0
    =
    \left\langle \partial g(\vh_0),
    \Delta \vh_0
    \right\rangle
    +
    \sum_{d=1}^D
    \left\langle
    \partial g(\vh_0), [\nabla_{\vx_0} \vh_0]_{d,:}^{\otimes 2}
    \right\rangle
  \end{pmatrix}\,.
\end{align*}
To obtain the connection to Taylor mode, we define $[\nabla_{\vx_0} \vh_0]_{d,:} = \vh_{1, d}$ and $\Delta_{\vx_0} \vh_0 = \sum_d \vh_{2,d}$ and $[\nabla_{\vx_0} g_0]_{d,:} = g_{1,d}$ and $\Delta_{\vx_0} g_0 = \sum_d g_{2,d}$, which allows us to rewrite the forward Laplacian as
\begin{align*}
  \begin{pmatrix}
    \vh_0
    \\
    \left\{ \vh_{1,d} \right\}_{d=1}^D
    \\
    \sum_d \vh_{2,d}
  \end{pmatrix}
  \to
  \begin{pmatrix}
    g_0 = g(\vh_0)
    \\
    \left\{
    g_{1,d} = \left\langle \partial g(\vh_0), \vh_{1,d} \right\rangle
    \right\}_{d=1}^{D}
    \\
    \sum_d g_{2,d}
    =
    \left\langle \partial g(\vh_0),
    \sum_d \vh_{2,d}
    \right\rangle
    +
    \sum_{d=1}^D
    \left\langle
    \partial g(\vh_0), \vh_{1,d}^{\otimes 2}
    \right\rangle
  \end{pmatrix}\,.
\end{align*}
This yields our collapsed Taylor mode propagation: the first equation is simply the forward pass, the second equation propagates the first-order derivatives along $D$ directions, and the last equation propagates the collapsed second-order derivatives, as described by setting $K=2$ in \Cref{eq:faa-di-bruno-expanded}.

\end{document}